\documentclass{article}

\PassOptionsToPackage{numbers, compress}{natbib}
\usepackage[final]{neurips_2022} 
\usepackage[utf8]{inputenc} % allow utf-8 input
\usepackage[T1]{fontenc}    % use 8-bit T1 fonts
\usepackage{hyperref}       % hyperlinks
\usepackage{url}            % simple URL typesetting
\usepackage{booktabs}       % professional-quality tables
\usepackage{amsfonts}       % blackboard math symbols
\usepackage{nicefrac}       % compact symbols for 1/2, etc.
\usepackage{microtype}      % microtypography
\usepackage{xcolor}         % colors
\usepackage{amsthm}
\usepackage{amssymb}
\usepackage{amsmath}
\usepackage{amstext}
\usepackage{mathtools}
\usepackage{upgreek}
\usepackage{graphicx}
\usepackage{subcaption}
\usepackage{enumitem}
\usepackage{changes}
\usepackage[ruled,vlined]{algorithm2e}
\usepackage{bbm}
\usepackage{breqn}
\usepackage{subcaption}
\usepackage{tabu}
\usepackage{hyperref}
\SetKwInOut{Parameter}{Parameters}

\newtheorem{theorem}{Theorem}[section]
\newtheorem{lemma}{Lemma}[section]
\newtheorem{assumption}{Assumption}[section]

\newtheorem{remark}{Remark}[section]

\makeatletter
\def\BState{\State\hskip-\ALG@thistlm}
\makeatother
\def \vx {\mathbf{x}}   % vector x, single image
\def \vy {\mathbf{y}}   % vector y, single image
\usepackage{amsmath,amsfonts,bm}

\def\eqref#1{equation~\ref{#1}}
\def\1{\bm{1}}

\def\vx{{\bm{x}}}
\def\vy{{\bm{y}}}
\def\vz{{\bm{z}}}

\DeclareMathAlphabet{\mathsfit}{\encodingdefault}{\sfdefault}{m}{sl}
\SetMathAlphabet{\mathsfit}{bold}{\encodingdefault}{\sfdefault}{bx}{n}

\usepackage{comment}
\usepackage{hyperref}
\usepackage{url}
\usepackage{xcolor}
\usepackage{wrapfig}
\usepackage{caption}
\usepackage{subcaption}

\setcitestyle{square}
\title{Recursive Reasoning in Minimax Games: A Level $k$ Gradient Play Method}

\author{%
  Zichu Liu\\
  University of Toronto\\
  \texttt{jieben.liu@mail.utoronto.ca} \\\And
  Lacra Pavel\\
  University of Toronto\\
  \texttt{pavel@ece.utoronto.ca}
}

\begin{document}

\maketitle

\begin{abstract}
Despite the success of generative adversarial networks (GANs) in generating visually appealing images, they are notoriously challenging to train. In order to stabilize the learning dynamics in minimax games, we propose a novel recursive reasoning algorithm: Level $k$ Gradient Play (Lv.$k$ GP) algorithm. In contrast to many existing algorithms, our algorithm does not require sophisticated heuristics or curvature information. We show that as $k$ increases, Lv.$k$ GP converges asymptotically towards an accurate estimation of players' future strategy.
Moreover, we justify that Lv.$\infty$ GP naturally generalizes a line of provably convergent game dynamics which rely on predictive updates. Furthermore, we provide its local convergence property in nonconvex-nonconcave zero-sum games and global convergence in bilinear and quadratic games. By combining Lv.$k$ GP with Adam optimizer, our algorithm shows a clear advantage in terms of performance and computational overhead compared to other methods. Using a single Nvidia RTX3090 GPU and 30 times fewer parameters than BigGAN on CIFAR-10, we achieve an FID of 10.17 for unconditional image generation within 30 hours, allowing GAN training on common computational resources to reach state-of-the-art performance.
\end{abstract}
\vspace{-0.3cm}
\section{Introduction}
\vspace{-0.3cm}
% In traditional machine learning, an ordinary training paradigm is to optimize the parameters of a model with respect to a single objective function given a set of training samples. A large amount of the recent advances in deep learning can be attributed to the success of gradient-based algorithms. A plethora of recent work on gradient-based algorithms has proven the convergence to the local minimum of the objective under a broad range of conditions. 
In recent years, there has been a surge of powerful models that require simultaneous optimization of several objectives. This increasing interest in multi-objective optimization arises in various domains - such as generative adversarial networks \cite{goodfellow2014generative,karras2019style}, adversarial attacks and robust optimization \cite{madry2017towards,carlini2017towards}, and multi-agent reinforcement learning \cite{lowe2017multi,vinyals2019grandmaster} - where several agents aim at minimizing their objectives simultaneously. Games generalize this optimization framework by introducing different objectives for different learning agents, known as players. 
\begin{figure}
\centering
\makebox[\textwidth][c]{\includegraphics[width=1.1\textwidth]{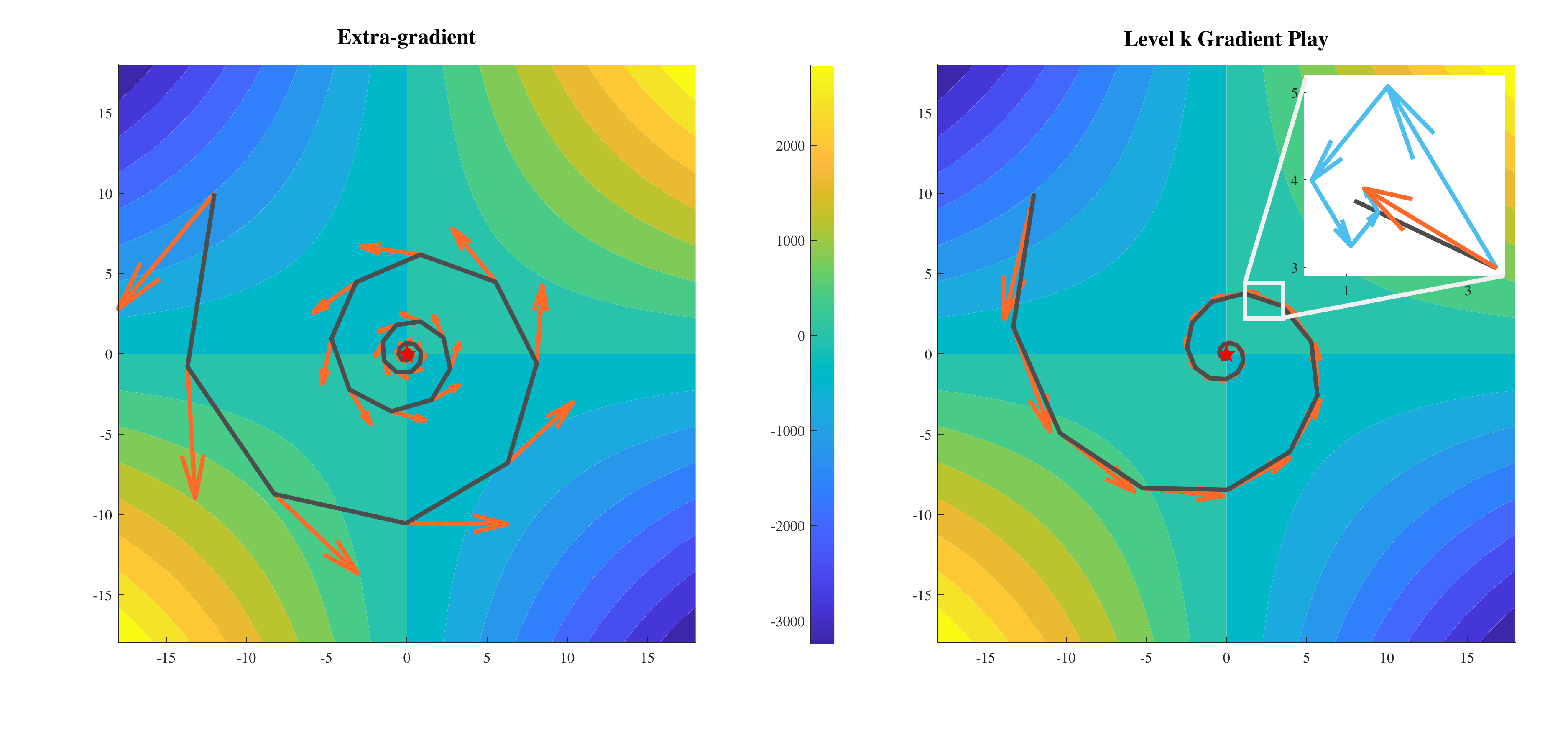}}
\caption{Illustration of predictive algorithms on: $\min_{x}\max_{y}10xy$. Left: Extra-gradient algorithm. Right: Level $k$ gradient play algorithm $(k=6)$. The solution, trajectory $\{x_{t},y_{t}\}_{t=1}^{T}$ and anticipated future state are shown with red star, black line and orange arrow, resp. The subplot in the right figure depicts how Lv.$k$ GP predicts future states by showing its reasoning procedure with blue arrows. More steps in the reasoning process leading to better anticipations and faster convergence.} 
\label{fig:convergence of LVk}
\vspace{-0.3cm}
\end{figure}
The generative adversarial network is a widely-used method of this type, which has demonstrated state-of-the-art performance in a variety of applications, including image generation \cite{karras2020analyzing,sauer2022stylegan}, image super-resolution \cite{ledig2017photo}, and image-to-image translation \cite{choi2018stargan}. Despite their success at generating visually appealing images, GANs are notoriously challenging to train \cite{mescheder2017numerics,mertikopoulos2018cycles}. Naive application of the gradient-based algorithm in GANs often leads to poor image diversity (sometimes manifesting as "mode collapse") \cite{mescheder2017numerics}, Poincare recurrence \cite{mertikopoulos2018cycles}, and subtle dependency on hyperparameters \cite{gemp2019unreasonable}. An immense corpus of work is devoted to exploring and enhancing the stability of GANs, including techniques as diverse as the use of optimal transport distance  \cite{arjovsky2017wasserstein,gulrajani2017improved}, critic gradient penalties \cite{thanh2019improving}, different neural network architectures \cite{karras2017progressive,brock2018large}, feature matching \cite{salimans2016improved}, pre-trained feature space \cite{sauer2021projected}, and minibatch discrimination \cite{salimans2016improved}. Nevertheless, architectural modifications (e.g., StyleGANs \cite{karras2019style}) require extensive computational resources, and many theoretically appealing methods (Follow-the-ridge \cite{wang2019solving}, CGD\cite{schafer2019competitive}) require Hessian inverse operations, which is infeasible for most GAN applications.

To stabilize the learning dynamics in GANs, many recent efforts rely on sophisticated heuristics that allow the agents to anticipate each other's next move \cite{foerster2017learning,schafer2019competitive,hemmat2020lead}. This anticipation is an example of a recursive reasoning procedure in cognitive science \cite{corballis2007uniqueness}. Similar to how humans think, recursive reasoning represents the belief reasoning process where each agent considers the reasoning process of other agents, based on which it expects to make better decisions. Importantly, it enables the use of opponents that reason about the learning agent, rather than assuming fixed opponents; the process can therefore be nested in a form as 'I believe that you believe that I believe...'. 
Based on this intuition, we introduce a novel recursive reasoning algorithm that utilizes only gradient information to optimize GANs. Our contributions can be summarized as follows: 

    (i) We propose a novel algorithm: level $k$ gradient play (Lv.$k$ GP), which is capable of reasoning about players' future strategy. In a game, agents at Lv.$k$ adjust their strategies in accordance with the strategies of Lv.$k-1$ agents. We justify that, while typical GANs optimizers, such as Learning with Opponent Learning Awareness (LOLA) and Symplectic Gradient Adjustment (SGA), approximate Lv.$2$ and Lv.$3$ GP, our algorithm permits higher levels of strategic reasoning. In addition, the proposed algorithm is amenable to neural network optimizers like Adam \cite{kingma2014adam}. 

    (ii) We show that, in smooth games, Lv.$k$ GP converges asymptotically towards an accurate prediction of agents' next move. Under mutual opponent shaping, two Lv.$\infty$ agents will naturally have a consistent view of one another if the Lv.$k$ GP converges as $k$ increases. Based on this idea, we provide a closed-form solution for Lv.$\infty$ GP: the Semi-Proximal Point Method (SPPM).

    (iii) We prove the local convergence property of Lv.$\infty$ GP in nonconvex - nonconcave zero-sum games and its global convergence in bilinear and quadratic games. The theoretical analysis we present indicates that strong interactions between competing agents can increase the convergence rate of Lv.$k$ GP agents in a zero-sum game.
    
    (iv) By combining Lv.$k$ GP with Adam optimizer, our algorithm shows a clear advantage in terms of performance and computational overhead compared to other methods. Using a single 3090 GPU with 30 times fewer parameters and 16 times smaller mini-batches than BigGAN \cite{brock2018large} on CIFAR-10, we achieve an FID score \cite{heusel2017gans} of 10.17 for unconditional image generation within 30 hours, allowing GAN training on common computing resources to reach state-of-the-art performance.  
\section{Related Works}
\vspace{-0.3cm}
In recent years, minimax problems have attracted considerable interest in machine learning in light of their connection with GANs. Gradient descent ascent (GDA), a generalization of gradient descent for minimax games, is the principal approach for training GANs in applications. GDA alternates between a gradient descent step for the min-player and a gradient ascent step for the max-player. The convergence of GDA in games is far from as well understood as gradient descent in single-objective problems. Despite the impressive image quality generated by GANs, GDA fails to converge even in bilinear zero-sum games. Recent research on GDA has established a unified picture of its behavior in bilinear games in continuous and discrete-time \cite{mertikopoulos2018cycles,papadimitriou2018nash,daskalakis2017training,mescheder2018training,gidel2019negative}. First, \cite{mertikopoulos2018cycles} revealed that continuous-time GDA dynamics in zero-sum games result in Poincare recurrence, where agents return arbitrarily close to their initial state infinitely many times. Second, \cite{bailey2020finite,zhang2021near} examined the discrete-time GDA dynamics, showing that simultaneous update of two players results in divergence while the agents' strategies remain bounded and cycle when agents take turns to update their strategies.

The majority of existing approaches to stabilizing GDA follow one of three lines of research. 
The essence of the first method is that the discriminator is trained until convergence while the generator parameters are frozen. As long as the generator changes slowly enough, the discriminator still converges in the presence of small generator perturbations. The two-timescale update rule proposed by \cite{goodfellow2014generative,heusel2017gans,metz2016unrolled} aims to keep the discriminator's optimality while updating the generator at an appropriate step size. \cite{jin2020local,fiez2021local} proved that this two-timescale GDA with finite timescale separation converges towards the strict local minimax/Stackelberg equilibrium in differentiable games. \cite{fiez2019convergence,wang2019solving} explicitly find the local minimax equilibrium in games with secon-order optimization algorithms.  

The second line of research overcomes the failure of GDA in games with predictive updates. Extra-gradient method (EG) \cite{korpelevich1976extragradient} and optimistic gradient descent (OGD) \cite{daskalakis2017training} use the predictability of the agents' strategy to achieve better convergence property. Their variants are developed to improve the training performance of GANs \cite{chavdarova2019reducing,gidel2018variational,mishchenko2020revisiting}. \cite{mokhtari2020unified} provided a unified analysis of EG and OGD, showing that they approximate the classical proximal point methods. Competitive gradient descent \cite{schafer2019competitive} models the agents' next move by solving a regularized bilinear approximation of the underlying game. Learning with opponent learning awareness (LOLA) and consistent opponent learning awareness (COLA) \cite{foerster2017learning,willi2022cola} introduced opponent shaping to this problem by explicitly modeling the learning strategy of other agents in the game. LOLA models opponents as naive learners rather than LOLA agents, while COLA utilizes neural networks to predict opponents' next move. Lookahead-minimax \cite{chavdarova2020taming} stabilizes GAN training by `looking ahead' at the sequence of future states generated by an inner optimizer. In the game theory literature, recent work has proposed Clairvoyant Multiplicative Weights Update (CMWU) for regret minimization in general games \cite{piliouras2021optimal}. Although CMWU is proposed to solve finite normal form games, which are different from unconstrained continuous games that Lv.$k$ GP aims to solve, both CMWU and Lv.$k$ GP share the same motivation of enabling the learning agent to update their strategy based on the opponent's future strategy.  From this aspect, Lv.$k$ GP can be viewed as a specialized variant of CMWU that is specific to the problem of two-player zero-sum games, but adapted for unconstrained continuous kernel games. 

Other methods directly modify the GDA algorithm with ad-hoc modifications of game dynamics and introduction of additional regularizers. Consensus optimization (CO) \cite{mescheder2017numerics} and gradient penalty \cite{gulrajani2017improved,thanh2019improving} improve convergence by directly minimizing the magnitude of players' gradients. Symplectic gradient adjustment (SGA) \cite{letcher2019differentiable,balduzzi2018mechanics,gemp2018global} improves convergence by disentangling convergent potential components from rotational Hamiltonian components of the vector field.
\vspace{-0.3cm}
\section{Preliminaries}
\vspace{-0.3cm}
\subsection{Notation}
In this paper, vectors are lower-case bold letters (e.g. $\bm{\theta}$), matrices are upper-case bold letters (e.g. $\bm{A}$). For a function $f:\mathbb{R}^{d}\to\mathbb{R}$, we denote its gradient by $\nabla f$. For functions of two vector arguments $f(\vx,\vy):\mathbb{R}^{d_1}\times\mathbb{R}^{d_2}\to\mathbb{R}$ we use $\nabla_x f, \nabla_y f$ to denote its partial gradients. We use $\nabla_{xx}f, \nabla_{yy}f, \nabla_{xy}f$ to denote its Hessian. A stationary point of $f$ denotes the point where $\nabla_x f = \nabla_y f = \bm{0}$. We use $\lVert \bm{v}\rVert$ to denote the Euclidean norm of vector $\bm{v}$. We refer to the largest and smallest eigenvalues of a matrix $\bm{A}$ by $\lambda_{\max}(\bm{A})$ and $\lambda_{\min}(\bm{A})$, respectively. Moreover, we denote the spectral radius of matrix $\bm{A}$ by $\rho(\bm{A}) = \max\{\lvert \lambda_1\rvert,\dots,\lvert \lambda_n\rvert\}$, i.e., the eigenvalue with largest absolute value.
\subsection{Problem Definition}
In order to justify the effectiveness of recursive reasoning procedure, in this paper, we consider the problem of training Generative Adversarial Networks (GANs)\cite{goodfellow2014generative}. The GANs training strategy defines a two-player game between a generative neural network $G_{\bm{\theta}}(\cdot)$ and a discriminative neural network $D_{\bm{\phi}}(\cdot)$. The generator takes as input random noise $\vz$ sampled from a known distribution $\mathbf{P}_{\vz}$, e.g., $\vz\sim \mathbf{P}_{\vz}$, and outputs a sample $G_{\bm{\theta}}(\vz)$. A discriminator takes as input a sample $\vx$ (either sampled from the true distribution $\mathbf{P}_{\vx}$ or from the generator) and attempts to classify it as real or fake. The goal of the generator is to fool the discriminator. The optimization of GAN is formulated as a two-player differentiable game where the generator $G_{\bm{\theta}}$ with parameter $\bm{\theta}$, and the discriminator $D_{\bm{\phi}}$ with parameters $\bm{\phi}$, aim at minimizing their own cost function $f(\bm{\theta},\bm{\phi})$ and $g(\bm{\theta},\bm{\phi})$ respectively, as follows:
\begin{equation}
    \min_{\bm{\theta}\in\mathbb{R}^{m}}f(\bm{\theta},\bm{\phi}) \text{   and   }\min_{\bm{\phi}\in\mathbb{R}^{n}}g(\bm{\theta},\bm{\phi}),
\end{equation}
where the two function $f \text{ and } g:\mathbb{R}^{m}\times\mathbb{R}^{n}\to\mathbb{R}$. When $f = -g$ the corresponding optimization problem is called a two-player zero-sum game and it becomes a minimax problem:
\begin{equation}\label{eq: minimax problem}
    \min_{\bm{\theta}\in\mathbb{R}^{m}}\max_{\bm{\phi}\in\mathbb{R}^{n}} f(\bm{\theta},\bm{\phi}).\tag{Minimax}
\end{equation}
In this work, we assume the cost functions have Lipschitz continuous gradients with respect to all model parameters $(\bm{\theta},\bm{\phi})$:

\begin{assumption}\label{assumption: Lipschitz gradient assumption}
The gradient $\nabla_{\bm{\theta}}f(\bm{\theta},\bm{\phi})$, is $L_{\bm{\theta}\bm{\theta}}-$Lipschitz with respect to $\bm{\theta}$ and $L_{\bm{\theta}\bm{\phi}}-$Lipschitz with respect to $\bm{\phi}$ and the gradient $\nabla_{\bm{\phi}}g(\bm{\theta},\bm{\phi})$, is $L_{\bm{\phi}\bm{\phi}}-$Lipschitz with respect to $\bm{\phi}$ and $L_{\bm{\phi}\bm{\theta}}-$Lipschitz with respect to $\bm{\theta}$, i.e.,
\begin{align}
    \lVert \nabla_{\bm{\theta}}f(\bm{\theta}_{1},\bm{\phi})-\nabla_{\bm{\theta}}f(\bm{\theta}_{2},\bm{\phi})\rVert &\leq L_{\bm{\theta}\bm{\theta}}\lVert\bm{\theta}_1 - \bm{\theta}_2 \rVert \text{   for all $\bm{\phi}$,}\nonumber\\
    \lVert \nabla_{\bm{\theta}}f(\bm{\theta},\bm{\phi}_{1})-\nabla_{\bm{\theta}}f(\bm{\theta},\bm{\phi}_{2})\rVert &\leq L_{\bm{\theta}\bm{\phi}}\lVert\bm{\phi}_1 - \bm{\phi}_2 \rVert \text{   for all $\bm{\theta}$,}\nonumber\\
    \lVert \nabla_{\bm{\phi}}g(\bm{\theta}_{1},\bm{\phi})-\nabla_{\bm{\phi}}g(\bm{\theta}_{2},\bm{\phi})\rVert &\leq L_{\bm{\phi}\bm{\theta}}\lVert\bm{\theta}_1 - \bm{\theta}_2 \rVert \text{   for all $\bm{\phi}$,}\nonumber\\
    \lVert \nabla_{\bm{\phi}}g(\bm{\theta},\bm{\phi}_{1})-\nabla_{\bm{\phi}}g(\bm{\theta},\bm{\phi}_{2})\rVert &\leq L_{\bm{\phi}\bm{\phi}}\lVert\bm{\phi}_1 - \bm{\phi}_2 \rVert \text{   for all $\bm{\theta}$.}\nonumber
\end{align}
We define $L:= \max\{L_{\bm{\theta}\bm{\theta}},L_{\bm{\theta}\bm{\phi}},L_{\bm{\phi}\bm{\theta}},L_{\bm{\phi}\bm{\phi}}\}$.
\end{assumption}

\section{Level $k$ Gradient Play}
\vspace{-0.3cm}
In this section, we propose a novel recursive reasoning algorithm, Level $k$ Gradient Play (Lv.$k$ GP), that allows the agents to discover self-interested strategies while taking into account other agents' reasoning processes. In Lv.$k$ GP, $k$ steps of recursive reasoning is applied to obtain an anticipated future state $(\bm{\theta}^{(k)}_{t},\bm{\phi}^{(k)}_{t})$, and the current states $(\bm{\theta}_{t},\bm{\phi}_{t})$ are then updated as follows:

\begin{equation}
\mathmakebox[0.8\textwidth]{
\text{Reasoning:}
\begin{cases}
        \bm{\theta}^{(n)}_{t} = \bm{\theta}_{t} - \eta\nabla_{\bm{\theta}}f(\bm{\theta}_{t},\bm{\phi}_{t}^{(n-1)})\nonumber\\
        \bm{\phi}^{(n)}_{t} = \bm{\phi}_{t} - \eta\nabla_{\bm{\phi}}g(\bm{\theta}_{t}^{(n-1)},\bm{\phi}_{t})
\end{cases}
\text{Update:}
\begin{cases}
    \bm{\theta}_{t+1} = \bm{\theta}_{t}^{(k)}\nonumber\\
    \bm{\phi}_{t+1} = \bm{\phi}_{t}^{(k)}
\end{cases}}
\label{eq: level k gradient play}\tag{Lv.k GP}
\end{equation}

% {\SetAlgoNoLine
% \begin{algorithm}
% \caption{Level-$k$ Gradient Play}\label{eq: level k gradient play}
% \KwIn{Stopping time $T$, reasoning depth $K$, learning rate $\eta$, initial weight $(\bm{\theta}_{0},\bm{\phi}_{0})$.}
% \For{t=0,\dots,T-1}{
%     \bm{\theta}_{t}^{(0)} \xleftarrow{} \bm{\theta}_{t},\bm{\phi}_{t}^{(0)} \xleftarrow{} \bm{\phi}_{t},\\
%     \For{k=1,\dots,K}{
%         \bm{\theta}^{(k)}_{t} = \bm{\theta}_{t} - \eta\nabla_{\bm{\theta}}f(\bm{\theta}_{t},\bm{\phi}_{t}^{(k-1)})\\
%         \bm{\phi}^{(k)}_{t} = \bm{\phi}_{t} - \eta\nabla_{\bm{\phi}}g(\bm{\theta}_{t}^{(k-1)},\bm{\phi}_{t})\\
%     }
%     \bm{\theta}_{t+1} \xleftarrow{} \bm{\theta}_{t}^{(k)},\bm{\phi}_{t+1} \xleftarrow{} \bm{\phi}_{t}^{(k)};\\
% }\\
% \end{algorithm}}
We define the current state $(\bm{\theta}_{t},\bm{\phi}_{t})$, to be the starting point $(\bm{\theta}_{t}^{(0)},\bm{\phi}_{t}^{(0)})$, of the reasoning process. Learning agents that adopt Lv.$k$ GP strategy are then called Lv.$k$ agents. Lv.$1$ agents act naively in response to the current state using GDA dynamics and Lv.$2$ agents act in response to Lv.$1$ agents by assuming its opponent as a naive learner. Therefore, Lv.$k$ GP allows for higher levels of strategic reasoning. The inspiration comes from how humans collaborate: humans are great at anticipating how their actions will affect others, so they frequently find out how to collaborate with other people to reach a "win-win" solution. 
The key to human collaboration is their ability to understand how other humans think which helps them develop strategies that benefit their collaborators. One of our main theoretical results is the following theorem, which demonstrates that agents adopting Lv.$k$ GP can precisely predict other players' next move and reach a consensus on their future strategies:
\begin{theorem}\label{thm: convergence of lvk gp}
Suppose Assumption \ref{assumption: Lipschitz gradient assumption} holds. Let us define $\bm{\omega}_t = [\bm{\theta}_t,\bm{\phi}_t]^T\in\mathbb{R}^{m+n}$, $\bm{\omega}_{t}^{(k)} = [\bm{\theta}_t^{(k)},\bm{\phi}_t^{(k)}]^T\in\mathbb{R}^{m+n}$ and $\Delta_{\max} = 2\times \max(\lVert \nabla_{\bm{\theta}}f(\bm{\theta}_t,\bm{\phi}_t)\rVert,\lVert \nabla_{\bm{\phi}}g(\bm{\theta}_t,\bm{\phi}_t)\rVert)$. Assume $\bm{\omega}_{t}^{(k)}$ lie in a complete subspace of $\mathbb{R}^{m+n}$. Then for Lv.k GP we have:
\begin{equation}
    \lVert \bm{\omega}_{t}^{(k)}-\bm{\omega}_{t}^{(k-1)}\rVert \leq \eta\cdot (\eta L)^{(k-1)}\Delta_{\max},
\end{equation}
Suppose the learning rate satisfies: $\eta < (2L)^{-1}$, then the sequence $\{\bm{\omega}_{t}^{(k)}\}_{k=0}^{\infty}$ is a Cauchy sequence. That is, given $\epsilon>0$, there exists $N$ such that, if $a>b>N$ then:
\begin{equation}
    \lVert \bm{\omega}_{t}^{(a)} - \bm{\omega}_{t}^{(b)}\rVert<\mathcal{O}(\eta^{b})<\epsilon\label{eq: Cauchy sequence}
\end{equation}
Moreover, the sequence $\{\bm{\omega}_{t}^{(k)}\}_{k=0}^{\infty}$ converges to a limit $\bm{\omega}_{t}^{*}$: $\lim_{k\to\infty}\bm{\omega}_{t}^{(k)}=\bm{\omega}_{t}^*$.
\end{theorem}
In accordance with Theorem \ref{thm: convergence of lvk gp}, if we define $\bm{\omega}_{t+1} = \bm{\omega}_{t}^{*}$, then Lv.$\infty$ GP is equivalent to the following implicit algorithm where we call it Semi-Proximal Point Method:
\begin{equation}
\begin{cases}
    \bm{\theta}_{t+1} = \bm{\theta}_{t} - \eta\nabla_{\bm{\theta}}f(\bm{\theta}_t,\bm{\phi}_{t+1})\nonumber\\
    \bm{\phi}_{t+1} = \bm{\phi}_{t} - \eta\nabla_{\bm{\phi}}g(\bm{\theta}_{t+1},\bm{\phi}_{t})\label{eq: SPPM}\tag{SPPM}
    \end{cases}
\end{equation}

\subsection{Algorithms as an Approximation of SPPM}
SPPM players arrive at a consensus by knowing precisely what their opponents' future strategies will be. Existing algorithms are not able to offer this kind of agreement. For instance, consensus optimization\cite{mescheder2017numerics} forces the learning agents to cooperate regardless of their own benefits. Agents employ extra-gradient method\cite{korpelevich1976extragradient}, SGA\cite{balduzzi2018mechanics}, and LOLA\cite{foerster2017learning} consider their opponents as naive learners, ignoring their strategic reasoning ability. 
CGD\cite{schafer2019competitive} takes into account the reasoning process of learning agents; however, it leads to an inaccurate prediction of agents in games that have cost functions with non-zero higher order derivatives ($n\geq 3$) \cite{willi2022cola}. In this section, we consider a subset of provably convergent variants of GDA in the \ref{eq: minimax problem} setting, showing that, for specific choice of hyperparameters, the mentioned algorithms either approximate SPPM or approximate the approximations of SPPM:
\begin{table}[h]
    \centering
    \caption{The update rules for the first player of SGA, LOLA, Lv.$2$ GP, LEAD, CGD, Lv.$3$ GP and SPPM in a \ref{eq: minimax problem} problem and their precision as an approximation of SPPM. $^\dagger$The usage of zero or negative momentum has been suggested in recent works \cite{gidel2019negative,gemp2019unreasonable}. For sake of comparison, we assume no momentum factor in LEAD's update, which corresponds to $\beta = 0$ in Equation 10 of \cite{hemmat2020lead}.\\}
    \begin{tabular}{llc}
    \toprule\midrule
        Algorithm & Update Rule & Precision\\\midrule
        SGA \cite{balduzzi2018mechanics} & $\bm{\theta}_{t+1} = \bm{\theta}_{t} - \eta\nabla_{\theta}f(\bm{\theta}_{t},\bm{\phi}_{t}) - \eta\gamma\nabla_{\theta\phi}f(\bm{\theta}_{t},\bm{\phi}_{t})\nabla_{\phi}f(\bm{\theta}_{t},\bm{\phi}_{t})$&------\\\midrule
        LOLA \cite{foerster2017learning}& $\bm{\theta}_{t+1} = \bm{\theta}_{t} - \eta\nabla_{\theta}f(\bm{\theta}_{t},\bm{\phi}_{t}) - \eta\delta\nabla_{\theta\phi}f(\bm{\theta}_{t},\bm{\phi}_{t})\nabla_{\phi}f(\bm{\theta}_{t},\bm{\phi}_{t})$&------\\\midrule
        Lv.2 GP & $\bm{\theta}_{t+1} = \bm{\theta}_{t} - \eta\nabla_{\theta}f(\bm{\theta}_{t},\bm{\phi}_{t} + \eta\nabla_{\phi}f(\bm{\theta}_{t},\bm{\phi}_{t}))$&$\mathcal{O}(\eta^2)$\\\midrule
        LEAD$^\dagger$ \cite{hemmat2020lead}& $\bm{\theta}_{t+1} = \bm{\theta}_{t} - \eta\nabla_{\theta}f(\bm{\theta}_{t},\bm{\phi}_{t}) - \alpha\nabla_{\theta\phi}f(\bm{\theta}_{t},\bm{\phi}_{t})(\bm{\phi}_{t}-\bm{\phi}_{t-1})$&------\\\midrule
        CGD \cite{schafer2019competitive}& $\bm{\theta}_{t+1} = \bm{\theta}_{t} - \eta\nabla_{\theta}f(\bm{\theta}_{t},\bm{\phi}_{t}) - \eta\nabla_{\theta\phi}f(\bm{\theta}_{t},\bm{\phi}_{t})(\bm{\phi}_{t+1}-\bm{\phi}_{t})$&$\mathcal{O}(\eta^3)$\\\midrule
        Lv.3 GP & $\bm{\theta}_{t+1} = \bm{\theta}_{t} - \eta\nabla_{\theta}f(\bm{\theta}_{t},\bm{\phi}_{t}+\eta\nabla_{\phi}f(\bm{\theta}_{t}-\eta\nabla_{\theta}f(\bm{\theta}_t,\bm{\phi}_t),\bm{\phi}_t))$&$\mathcal{O}(\eta^3)$\\\midrule
        SPPM (Lv.$\infty$ GP) & $\bm{\theta}_{t+1} = \bm{\theta}_{t} - \eta\nabla_{\theta}f(\bm{\theta}_{t},\bm{\phi}_{t+1})$&$0$\\\midrule\bottomrule
    \end{tabular}
    \label{tab:precisions of algorithms}
\end{table}

In Table \ref{tab:precisions of algorithms}, we compare the orders of precision of different algorithms as an approximation of SPPM in \ref{eq: minimax problem} games with infinitely differentiable objective functions. In accordance with Equation (\ref{eq: Cauchy sequence}) of Theorem \ref{thm: convergence of lvk gp}, Lv.k GP is an $\mathcal{O}(\eta^{k})$ approximation of SPPM. In order to analyze how well existing algorithms approximate SPPM, we consider the first-order Taylor approximation to SPPM:\vspace{-.15in}
\begin{equation}
\hspace{-0.1cm}
\begin{cases}
    \bm{\theta}_{t\!+\!1}\! =\!\bm{\theta}_{t}\! - \!\eta \nabla_{\theta}f(\bm{\theta}_{t},\bm{\phi}_{t})\! - \!\eta^2 \nabla_{\theta\phi}f(\bm{\theta}_{t},\bm{\phi}_{t}) \nabla_{\phi}f(\bm{\theta}_{t},\bm{\phi}_{t})\! - \!\eta^2 \nabla_{\theta\phi}f(\bm{\theta}_{t},\bm{\phi}_{t})\nabla_{\phi\theta}f(\bm{\theta}_{t},\bm{\phi}_{t})(\bm{\theta}_{t\!+\!1}\! - \!\bm{\theta}_{t})\nonumber\\
    \smash{\underbrace{\bm{\phi}_{t\!+\!1}\! = \!\bm{\phi}_{t}\! + \!\eta \nabla_{\theta}f(\bm{\theta}_{t},\bm{\phi}_{t})\! - \!\eta^2 \nabla_{\phi\theta}f(\bm{\theta}_{t},\bm{\phi}_{t}) \nabla_{\theta}f(\bm{\theta}_{t},\bm{\phi}_{t})}_{\text{$1^{st}$ order approximation of Lv.2 GP}}\! - \!\eta^2 \nabla_{\phi\theta}f(\bm{\theta}_{t},\bm{\phi}_{t})\nabla_{\theta\phi}f(\bm{\theta}_{t},\bm{\phi}_{t})(\bm{\phi}_{t\!+\!1}\! - \!\bm{\phi}_{t})}\nonumber
\end{cases}
\vphantom{\begin{cases}
    \bm{\theta}_{t+1} =\bm{\theta}_{t} - \eta \nabla_{\theta}f(\bm{\theta}_{t},\bm{\phi}_{t}) - \eta^2 \nabla_{\theta\phi}f(\bm{\theta}_{t},\bm{\phi}_{t}) \nabla_{\phi}f(\bm{\theta}_{t},\bm{\phi}_{t}) - \eta^2 \nabla_{\theta\phi}f(\bm{\theta}_{t},\bm{\phi}_{t})\nabla_{\phi\theta}f(\bm{\theta}_{t},\bm{\phi}_{t})(\bm{\theta}_{t+1} - \bm{\theta}_{t})\nonumber\\
    \underbrace{\bm{\phi}_{t+1} = \bm{\phi}_{t} + \eta \nabla_{\theta}f(\bm{\theta}_{t},\bm{\phi}_{t}) - \eta^2 \nabla_{\phi\theta}f(\bm{\theta}_{t},\bm{\phi}_{t}) \nabla_{\theta}f(\bm{\theta}_{t},\bm{\phi}_{t})}_{\text{$1^{st}$ order approximation of Lv.2 GP}} - \eta^2 \nabla_{\phi\theta}f(\bm{\theta}_{t},\bm{\phi}_{t})\nabla_{\theta\phi}f(\bm{\theta}_{t},\bm{\phi}_{t})(\bm{\phi}_{t+1} - \bm{\phi}_{t})\nonumber\vspace{0.12in}
\end{cases}}
\end{equation}
Under-brace terms correspond to the first-order Taylor approximation of Lv.$2$ GP. For an appropriate choice of hyperparameters, SGA ($\gamma = \eta$) and LOLA ($\delta = \eta$) are identical to the first-order Taylor approximation of Lv.2 GP, where each agent models their opponent as a naive learner. Hence, we list them above the Lv.2 GP, which approximates SPPM up to $\mathcal{O}(\eta^2)$. CGD exactly recovers the first-order Taylor approximation of SPPM.\footnote{The CGD update for the max player $\bm{\phi}$ is $\bm{\phi}_{t+1} = \bm{\phi}_{t} +\eta\nabla_{\phi}f(\bm{\theta}_{t},\bm{\phi}_{t}) + \eta\nabla_{\phi\theta}f(\bm{\theta}_{t},\phi_{t})(\bm{\theta}_{t+1}-\bm{\theta}_{t})$. If we substitute $(\bm{\phi}_{t+1}-\bm{\phi}_{t})$ into $\bm{\theta}$'s update (and substitute $(\bm{\theta}_{t+1}-\bm{\theta}_{t})$ into $\bm{\phi}$'s update, respectively), we arrive at the first-order approximation of SPPM. See A.5 for derivation details.} In games with cost functions that have non-negative higher order derivatives ($n\geq3$), the remaining term in SPPM's first-order approximation is an error of magnitude $\mathcal{O}(\eta^3)$, which means that CGD's accuracy is in the same range as that of Lv.$3$ GP. In bilinear and quadratic games where the objective function is at most twice differentiable, CGD is equivalent to SPPM. The distinction between LEAD ($\alpha=\eta$) and CGD can be understood by considering their update rules. LEAD is an explicit method where opponents' potential next strategies are anticipated based on their most recent move $(\bm{\phi}_{t}-\bm{\phi}_{t-1})$. On the contrary, CGD accounts for this anticipation in an implicit manner, $(\bm{\phi}_{t+1}-\bm{\phi}_{t})$, where the future states appear in current states' update rules. Therefore, the computation of CGD updates requires solving an function involving additional Hessian inverse operations. A numerical justification is also provided in Table \ref{tab:Lv.k convergence}, showing that the approximation accuracy of Lv.$k$ GP improves as $k$ increases.

\section{Convergence Property}\label{sec:6}
\vspace{-0.3cm}
In Theorem \ref{thm: convergence of lvk gp}, we have analytically proved that Lv.$k$ GP convergences asymptotically towards SPPM, we will use this result to study the convergence property of Lv.$k$ GP in games based on our analysis of SPPM.
The local convergence of SPPM in a non convex - non concave game can be analyzed via the spectral radius of the game Jacobian around a stationary point:
\begin{theorem}\label{thm: convergence in nonconvex gamae}
Consider the (\ref{eq: minimax problem}) problem under Assumption \ref{assumption: Lipschitz gradient assumption} and Lv.$k$ GP. Let $(\bm{\theta}^*,\bm{\phi}^*)$ be a stationary point. Suppose $\bm{\theta}_{t}-\bm{\theta}^*$ not in kernel of $\nabla_{\phi\theta}f(\bm{\theta}^*,\bm{\phi}^*)$, $\bm{\phi}_{t}-\bm{\phi}^*$ not in kernel of $\nabla_{\theta\phi}f(\bm{\theta}^*,\bm{\phi}^*)$ and $\eta<(L)^{-1}$. There exists a neighborhood $\mathcal{U}$ of $(\bm{\theta}^*,\bm{\phi}^*)$ such that if SPPM started at $(\bm{\theta}_{0},\bm{\phi}_{0})\in\mathcal{U}$, the iterates $\{\bm{\theta}_{t},\bm{\phi}_{t}\}_{t\geq0}$ generated by SPPM satisfy:
\begin{equation}
    \lVert\bm{\theta}_{t+1} - \bm{\theta}^*\rVert^2 + \lVert\bm{\phi}_{t+1} - \bm{\phi}^*\rVert^2\leq \frac{\rho^2(\bm{I}-\eta\nabla_{\theta\theta}f^*)\lVert\bm{\theta}_{t} - \bm{\theta}^*\rVert^2+\rho^2(\bm{I}+\eta\nabla_{\phi\phi}f^*)\lVert\bm{\phi}_{t} - \bm{\phi}^*\rVert^2}{1+\eta^2\lambda_{\min}(\nabla_{\theta\phi}f^*\nabla_{\phi\theta}f^*)}\nonumber
\end{equation}
where $f^* = f(\bm{\theta}^*,\bm{\phi}^*)$. Moreover,for any $\eta$ satisfying:
\begin{equation}
    \frac{\max(\rho^2(\bm{I}-\eta\nabla_{\theta\theta}f^*),\rho^2(\bm{I}+\eta\nabla_{\phi\phi}f^*))}{1+\eta^2\lambda_{\min}(\nabla_{\theta\phi}f^*\nabla_{\phi\theta}f^*)}< 1\label{eq:local convergence},
\end{equation}
SPPM converges asymptotically to $(\bm{\theta}^*,\bm{\phi}^*)$.
\end{theorem}
\begin{remark}\label{remark: convergence of Lvk GP}
Following the same condition as in \ref{thm: convergence in nonconvex gamae}, the iterates generated by Lv.$2k$ GP satisfies:
\begin{align}
    \lVert &\bm{\theta}_{t}^{(2k)}-\bm{\theta}^*\rVert^2 + \lVert \bm{\phi}_{t}^{(2k)}-\bm{\phi}^*\rVert^2 \nonumber\\
    &\leq a \!\left( \frac{\rho^2(\bm{I}\!-\eta\nabla_{\theta\theta}f^*)\lVert\bm{\theta}_{t} \!-\! \bm{\theta}^*\rVert^2\!+\!\rho^2(\bm{I}\!+\!\eta\nabla_{\phi\phi}f^*)\lVert\bm{\phi}_{t} \!-\! \bm{\phi}^*\rVert^2}{1\!+\!\eta^2\lambda_{\min}(\nabla_{\theta\phi}f^*\nabla_{\phi\theta}f^*)}\right) \!+\! b(\lVert \bm{\theta}_{t} \!-\! \bm{\theta}^*\rVert^2\!+\!\lVert \bm{\phi}_{t} \!-\! \bm{\phi}^*\rVert^2)\nonumber
\end{align}
where
% \begin{align}
% \hspace{-0.8cm}
% a \!=\!
% \begin{cases}
%     \frac{(1\!+\!(\eta^2\lambda_{\max}(\nabla_{\theta\phi}f^*\nabla_{\phi\theta}f^*))^{k})^2}{1\!-\!(\eta^2\lambda_{\max}(\nabla_{\theta\phi}f^*\nabla_{\phi\theta}f^*))^{k}} \text{  , odd $k$  }\\
%     \frac{(1\!-\!(\eta^2\lambda_{\min}(\nabla_{\theta\phi}f^*\nabla_{\phi\theta}f^*))^{k})^2}{1\!-\!(\eta^2\lambda_{\max}(\nabla_{\theta\phi}f^*\nabla_{\phi\theta}f^*))^{k}} \text{  , even $k$}\nonumber
% \end{cases}
% \hspace{-0.3cm}\text{and }\nonumber\\b \!=\! \frac{(\eta^2\lambda_{\max}(\nabla_{\theta\phi}f^*\nabla_{\phi\theta}f^*))^k(1\!-\!(\eta^2\lambda_{\min}(\nabla_{\theta\phi}f^*\nabla_{\phi\theta}f^*))^{k})}{1\!-\!(\eta^2\lambda_{\max}(\nabla_{\theta\phi}f^*\nabla_{\phi\theta}f^*))^{k}}\nonumber
% \end{align}
\begin{align}
\hspace{-0.8cm}
a &\!=\!
    \frac{(1\!+\!(\eta^2\lambda_{\max}(\nabla_{\theta\phi}f^*\nabla_{\phi\theta}f^*))^{k})^2}{1\!-\!(\eta^2\lambda_{\max}(\nabla_{\theta\phi}f^*\nabla_{\phi\theta}f^*))^{k}} \text{  for odd $k$, or  }
    \frac{(1\!-\!(\eta^2\lambda_{\min}(\nabla_{\theta\phi}f^*\nabla_{\phi\theta}f^*))^{k})^2}{1\!-\!(\eta^2\lambda_{\max}(\nabla_{\theta\phi}f^*\nabla_{\phi\theta}f^*))^{k}} \text{  for even $k$,}\nonumber\\
b &\!=\! \frac{(\eta^2\lambda_{\max}(\nabla_{\theta\phi}f^*\nabla_{\phi\theta}f^*))^k(1\!-\!(\eta^2\lambda_{\min}(\nabla_{\theta\phi}f^*\nabla_{\phi\theta}f^*))^{k})}{1\!-\!(\eta^2\lambda_{\max}(\nabla_{\theta\phi}f^*\nabla_{\phi\theta}f^*))^{k}}.\nonumber
\end{align}
\end{remark}
\begin{wrapfigure}{r}{0.4\textwidth}
    \vspace{-0.9cm}
    \includegraphics[width=0.4\textwidth]{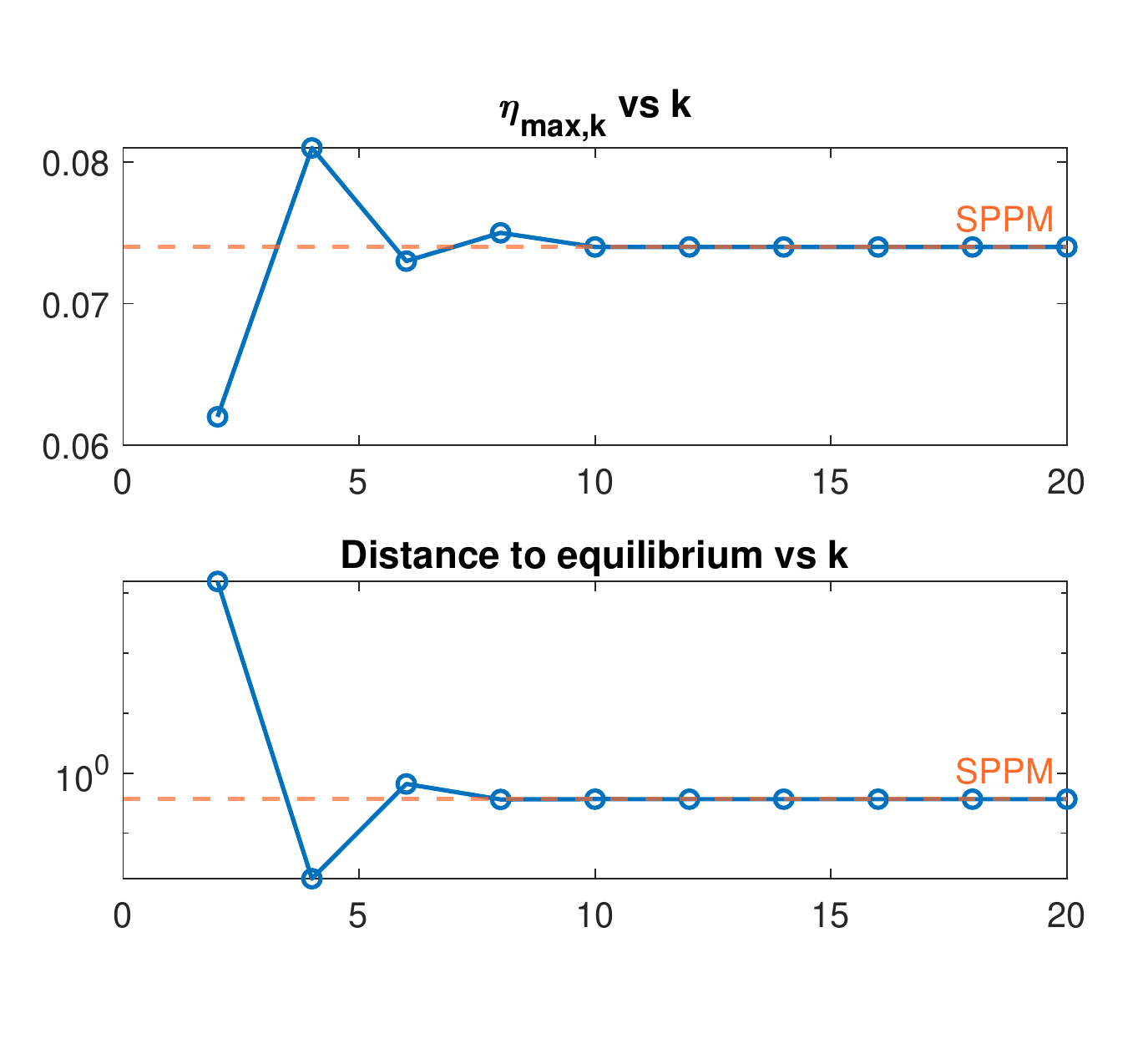}
    \caption{\textbf{Top}: max step size against $k$, \textbf{Right}: distance to equilibrium against $k$. In both figures the red dashed line represents the value for SPPM.}
\label{fig:Lvk vs SPPM}
\end{wrapfigure}
We assume that the difference between the trajectory $\{\bm{\theta}_{t},\bm{\phi}_{t}\}_{t\geq0}$ and the stationary point $(\bm{\theta}^*,\bm{\phi}^*)$ is not in the kernel of $\nabla_{\phi\theta}f(\bm{\theta}^*,\bm{\phi}^*)$ and $\nabla_{\theta\phi}f(\bm{\theta}^*,\bm{\phi}^*)$ for the sake of simplicity. Detailed proofs without this assumption are provided in Appendix \ref{appendix: without kernel assumption}. Following the same setting as Theorem \ref{thm: convergence of lvk gp}, we have $\eta<L^{-1}$, which ensures that $\eta^2\lambda_{\max}(\nabla_{\theta\phi}f^*\nabla_{\phi\theta}f^*)<1$. Therefore, as $k\to\infty$, $a\to1$ and $b\to0$ in Remark \ref{remark: convergence of Lvk GP}, and as such Lv.$k$ GP has similar local convergence properties to SPPM. Figure \ref{fig:Lvk vs SPPM} illustrates this property in a quadratic game. Lv.$k$ GP may behave differently than SPPM at lower values of $k$, but as $k$ increases, both max step size $\eta_{\max,k}$ and distance to equilibrium for a fixed number of iterations under $\eta_{\max,\infty}$ converges to that of SPPM. Thus, we observe that Lv.$k$ GP is empirically similar to SPPM at higher values of $k$.
%SPPM is thus a suitable choice for describing how Lv.$k$ GP converges at higher values of $k$.

We study the global convergence property of SPPM by analyzing its behavior in bilinear and quadratic games. Consider the following bilinear game:
\begin{equation} \min_{\bm{\theta}\in\mathbb{R}^n}\max_{\bm{\phi}\in\mathbb{R}^{n}}\bm{\theta}^{T}\bm{M}\bm{\phi}\tag{Bilinear game}\label{eq: bilinear game}
\end{equation}
where $\bm{M}$ is a full rank matrix. The following theorem summarizes SPPM's convergence property:
\begin{theorem}\label{thm: convergence in bilinear game}
Consider the \ref{eq: bilinear game} and the SPPM method. Further, we define $r_{t} = \lVert\bm{\theta}_{t} - \bm{\theta}^*\rVert^2 + \lVert\bm{\phi}_{t} - \bm{\phi}^*\rVert^2$. Then, for any $\eta>0$, the iterates $\{\bm{\theta}_{t},\bm{\phi}_{t}\}_{t\geq0}$ generated by SPPM satisfy
\begin{equation}
    r_{t+1}\leq \frac{1}{1+\eta^2\lambda_{\min}(\bm{M}^T\bm{M})}r_{t}.
\end{equation}
\end{theorem}
It is worth noting that the convergence property of Lv.$k$ GP in bi-linear game has been studied in \cite{azizian2020tight}.
Furthermore, we study the convergence property of SPPM in the \ref{eq: quadratic game}:
\begin{equation}
\min_{\bm{\theta}\in\mathbb{R}^n}\max_{\bm{\phi}\in\mathbb{R}^{n}}\bm{\theta}^{T}\bm{A}\bm{\theta} + \bm{\phi}^{T}\bm{B}\bm{\phi} + \bm{\theta}^{T}\bm{C}\bm{\phi}\tag{Quadratic game}\label{eq: quadratic game}
\end{equation}
where $\bm{A}\in\mathbb{R}^{n\times n}$ is symmetric and positive definite, $\bm{B}\in\mathbb{R}^{n\times n}$ is symmetric and negative definite and the interaction term $\bm{C}\in\mathbb{R}^{n\times n}$ is full rank. SPPM in quadratic games converges with the following rate:
\begin{theorem}\label{thm: convergence in quadratic game}
Consider the \ref{eq: quadratic game} and the SPPM. Then, for any $\eta>0$, the iterates $\{\bm{\theta}_{t},\bm{\phi}_{t}\}_{t\geq0}$ generated by SPPM satisfy
\begin{equation}
    \lVert\bm{\theta}_{t+1} - \bm{\theta}^*\rVert^2 + \lVert\bm{\phi}_{t+1} - \bm{\phi}^*\rVert^2\leq \frac{\rho^2(1-\eta\bm{A})\lVert\bm{\theta}_{t} - \bm{\theta}^*\rVert^2+\rho^2(1+\eta\bm{B})\lVert\bm{\phi}_{t} - \bm{\phi}^*\rVert^2}{1+\eta^2\lambda_{\min}(\bm{C}^T\bm{C})}
\end{equation}
\end{theorem}
\begin{figure}[h]
\centering
\makebox[\textwidth][c]{\includegraphics[width=1.06\textwidth]{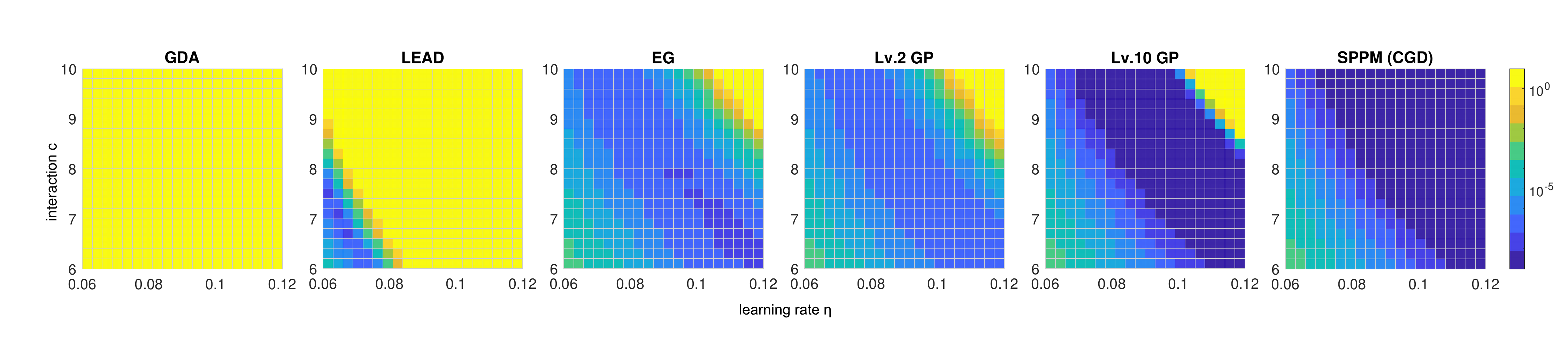}}
\vspace*{-1mm}\caption{A grid of experiments for different algorithms with different values of interaction $c$ and learning rates $\eta$. The color in each cell indicates the distance to the equilibrium after 100 iterations. Note that the CGD update is equivalent to SPPM in quadratic games.} 
\label{fig:quadratic game example}
\end{figure}
Theorem \ref{thm: convergence in nonconvex gamae},\ref{thm: convergence in bilinear game} and \ref{thm: convergence in quadratic game} indicate that stronger interaction $\nabla_{\theta\phi}f(\bm{\theta}^*,\bm{\phi}^*)$ improve the convergence rate towards the stationary points. By contrast, in existing modifications of GDA, the step size is chosen in inversely proportional to the interaction term $\nabla_{\theta\phi}f(\bm{\theta}^*,\bm{\phi}^*)$ \cite{mescheder2017numerics,daskalakis2017training,liang2019interaction}. In Figure \ref{fig:quadratic game example}, we showcase the effect of interaction on different algorithms in the \ref{eq: quadratic game} setup with dimension $n=5$ and the interaction matrix is defined as $\bm{C} = c\bm{I}$. Stronger interaction corresponds to higher values of $c$. A key difference between SPPM and Lv.$k$ GP in the experiments is that, SPPM converges with any step size - and so arbitrarily fast - while it is not the case for Lv.$k$ GP. In order to approximate SPPM, one needs Lv.$k$ GP to be a contraction and so have $\eta<L^{-1}$. This result implies an additional bound on step sizes for Lv.$k$ GP. As long as the constraint on $\eta$ is satisfied, stronger interaction only improves convergence for higher order Lv.$k$ GP while all other algorithms quickly diverge as $\eta$ and $c$ increase.

\section{Experimental Results}\label{sec:exp}\vspace{-2mm}
In this section, we discuss our implementation of Lv.$k$ GP algorithm for training GANs. Its performance is evaluated on 8-Gaussians and two representative datasets CIFAR-10 and STL-10.
% In this section, we propose to combine Lv.$k$ GP with standard neural network optimizers and investigate the empirical performance of our algorithm for GAN training.\vspace{-2mm}
\vspace{-2mm}
\subsection{Level $k$ Adam}\vspace{-2mm}
We propose to combine Lv.$k$ GP with the Adam optimizer \cite{kingma2014adam}. Preliminary experiments find that Lv.$k$ Adam to converge much faster than Lv.$k$ GP, see A.8 for experiment results. A detailed pseudo-code for Level $k$ Adam (Lv.$k$ Adam) on GAN training with loss functions $\mathcal{L}_{G}$ and $\mathcal{L}_{D}$ is given in Algorithm \ref{algo: level k adam}. For the Adam optimizer, there are several possible choices on how to update the moments. This choice can lead to different algorithms in practice. Unlike \cite{gidel2018variational} where the moments are updated on the fly, in Algorithm \ref{algo: level k adam}, we keep the moments fixed in the reasoning steps and update it together with model parameters. In Table \ref{tab:Lv.k convergence}, our experiment result suggests that the proposed Lv.$k$ Adam algorithm converges asymptotically as the number of reasoning steps $k$ increases. 
\begin{algorithm}[h]
\caption{Level $k$ Adam: proposed Adam with recursive reasoning steps}\label{algo: level k adam}
\KwIn{Stopping time $T$, reasoning steps $k$, learning rate $\eta_{\bm{\theta}},\eta_{\bm{\phi}}$, decay rates for momentum estimates $\beta_1,\beta_2$, initial weight $(\bm{\theta}_{0},\bm{\phi}_{0})$, $\bm{P}_{\vx}$ and $\bm{P}_{\vz}$ real and noise-data distributions, losses $\mathcal{L}_{G}(\bm{\theta},\bm{\phi},\vx,\vz)$ and $\mathcal{L}_{D}(\bm{\theta},\bm{\phi},\vx,\vz)$, $\epsilon=1e-8$.}
\Parameter{Initial parameters: $\bm{\theta}_{0},\bm{\phi}_{0}$\\
Initialize first moments:$\bm{m}_{\theta,0} \xleftarrow{} 0,\bm{m}_{\phi,0} \xleftarrow{} 0$\\
Initialize second moments:$\bm{v}_{\theta,0} \xleftarrow{} 0,\bm{v}_{\phi,0} \xleftarrow{} 0$}
\For{t=0,\dots,T-1}{
    \text{\textbf{Sample} new mini-batch:} $\vx,\vz\sim\bm{P}_{\vx},\bm{P}_{\vz}$,\\
    $\bm{\theta}_{t}^{(0)} \xleftarrow{} \bm{\theta}_{t},\bm{\phi}_{t}^{(0)} \xleftarrow{} \bm{\phi}_{t}$,\\
    \For{n=1,\dots,k}{
        Compute stochastic gradient: $\bm{g}_{\bm{\theta},t}^{(n)} = \nabla_{\theta}\mathcal{L}_{G}(\bm{\theta}_{t},\bm{\phi}^{(n-1)}_{t},\vx,\vz); \bm{g}_{\bm{\phi},t}^{(n)} =\nabla_{\phi}\mathcal{L}_{D}(\bm{\theta}^{(n-1)}_{t},\bm{\phi}_{t},\vx,\vz)$\\
        Update estimate of first moment:
        $\bm{m}_{\theta,t}^{(n)}=\beta_1\bm{m}_{\theta,t-1} + (1-\beta_1)\bm{g}^{(n)}_{\theta,t};\bm{m}_{\phi,t}^{(n)}=\beta_1\bm{m}_{\phi,t-1} + (1-\beta_1)\bm{g}^{(n)}_{\phi,t}$\\
        Update estimate of second moment:
        $\bm{v}_{\theta,t}^{(n)}=\beta_2\bm{v}_{\theta,t-1} + (1-\beta_2)(\bm{g}^{(n)}_{\theta,t})^2;\bm{v}_{\phi,t}^{(n)}=\beta_2\bm{v}_{\phi,t-1} + (1-\beta_2)(\bm{g}^{(n)}_{\phi,t})^2$\\
        Correct the bias for the moments: $\bm{\hat{m}}_{\theta,t}^{(n)} = \frac{\bm{m}^{(n)}_{\theta,t}}{(1-\beta_1^{t})},\bm{\hat{m}}_{\phi,t}^{(n)} = \frac{\bm{m}^{(n)}_{\phi,t}}{(1-\beta_1^{t})};\bm{\hat{v}}_{\theta,t}^{(n)} = \frac{\bm{v}^{(n)}_{\theta,t}}{(1-\beta_2^{t})},\bm{\hat{v}}_{\phi,t}^{(n)} = \frac{\bm{v}^{(n)}_{\phi,t}}{(1-\beta_2^{t})}$\\
        Perform Adam update:   $\bm{\theta}^{(n)}_{t} = \bm{\theta}_{t} - \eta_{\theta}\frac{\bm{\hat{m}}_{\theta,t}^{(n)}}{\sqrt{\bm{\hat{v}}_{\theta,t}^{(n)}}+\epsilon};\bm{\phi}^{(n)}_{t} = \bm{\phi}_{t} - \eta_{\phi}\frac{\bm{\hat{m}}_{\phi,t}^{(n)}}{\sqrt{\bm{\hat{v}}_{\phi,t}^{(n)}}+\epsilon}$\\
    }
    $\bm{\theta}_{t+1} \xleftarrow{} \bm{\theta}_{t}^{(k)},\bm{\phi}_{t+1} \xleftarrow{} \bm{\phi}_{t}^{(k)}$;\\
    $\bm{m}_{\theta,t} \xleftarrow{} \bm{m}_{\theta,t}^{(k)},\bm{m}_{\phi,t} \xleftarrow{} \bm{m}_{\phi,t}^{(k)}$;\\
    $\bm{v}_{\theta,t} \xleftarrow{} \bm{v}_{\theta,t}^{(k)},\bm{v}_{\phi,t} \xleftarrow{} \bm{v}_{\phi,t}^{(k)}$\\
}
\end{algorithm}
\subsection{8-Gaussians}
\begin{wrapfigure}{r}{0.4\textwidth}
    \includegraphics[width=0.4\textwidth]{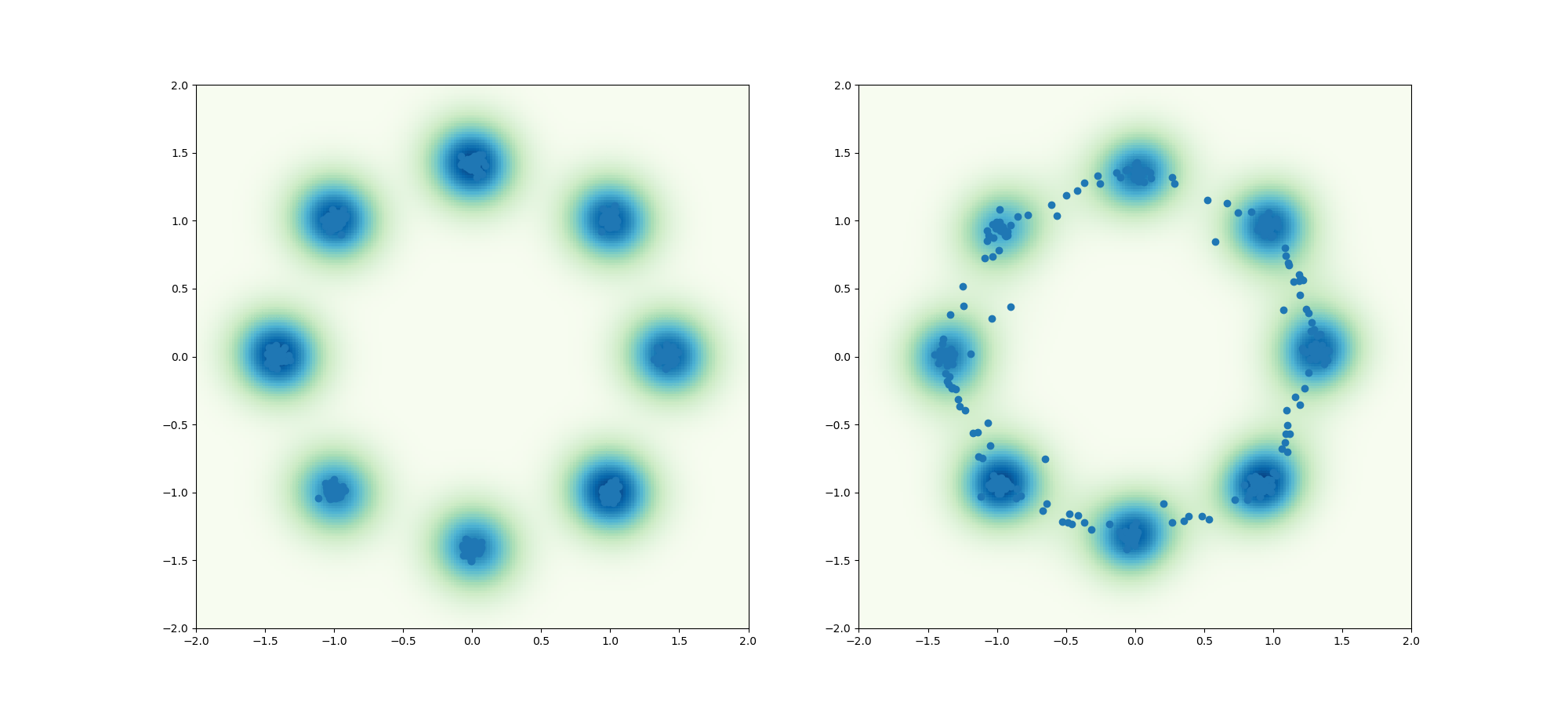}
    \caption{\textbf{Left}: real distribution, \textbf{Right}: generated distribution}
\label{fig: training result of Lvk adam on 8-Gaussians}
\end{wrapfigure}

In our first experiment, we evaluate Lv.$k$ Adam on generating a mixture of 8-Gaussians with standard deviations equal to $0.05$ and modes uniformly distributed around the unit circle. We use a two layer multi-layer perceptron with ReLU activations, latent dimension of~$64$ and batch size of~$128$. The generated distribution is presented in Figure \ref{fig: training result of Lvk adam on 8-Gaussians}.
\begin{table}[h]
    \centering
    \caption{The difference between two states generated by Lv.$k$ Adam and Lv.$k$ GP averaged over 100 steps. $^\dagger$The difference is smaller than machine precision.\\}
    \begin{tabular}{cccccc}
    \toprule\midrule
        $\frac{1}{100}\sum_{t=1}^{100}r_{t}^{(k)}$& $k=2$  & $k=4$   & $k=6$   & $k=8$   & $k=10$\\\midrule
        Lv.$k$ Adam& $1.04\times 10^{-1}$& $1.60\times 10^{-2}$& $2.91\times 10^{-3}$& $6.08\times 10^{-4}$& $1.68\times 10^{-4}$\\
        Lv.$k$ GP& $9.79\times 10^{-9}$& $3.84\times 10^{-15}$& $1.31\times 10^{-17}$& $1.68\times 10^{-19}$& $\approx0 ^\dagger$\\\midrule\bottomrule
    \end{tabular}
    \label{tab:Lv.k convergence}
\end{table}
In addition to presenting the mode coverage of the generated distribution after training, we also study the convergence of the reasoning steps of Lv.$k$ Adam and Lv.$k$ GP. Let us define the difference between the states of Lv.$k$ and Lv.$k-1$ agents at time $t$ as:
\begin{equation}
    r^{(k)}_{t} = \lVert \bm{\theta}^{(k)}_{t} - \bm{\theta}^{(k-1)}_{t}\rVert^2+\lVert \bm{\phi}^{(k)}_{t} - \bm{\phi}^{(k-1)}_{t}\rVert^2.
\end{equation}
We measure the difference averaged over the first 100 iterations, $\frac{1}{100}\sum_{t=1}^{100}r_{t}^{k}$, for Lv.$10$ Adam ($\eta=10^{-4}$) and Lv.$10$ GP ($\eta=10^{-2}$). The result presented in Table\ref{tab:Lv.k convergence} demonstrates that both Lv.$k$ GP and Lv.$k$ Adam are converging as $k$ increases. Moreover, the estimation precision of Lv.$k$ GP improves rapidly and converges to $0$ within finite steps, making it an accurate estimation for SPPM.
\begin{table}[h]
    \centering
    \caption{FID and Inception scores of different algorithms and architectures on CIFAR-10. Results are averaged over 3 runs. $^\dagger$We re-evaluate its performance on the official implementation of FID.\\}
    \begin{tabular}{l l c c}
    \toprule\midrule
    Algorithm & Architecture & FID $\downarrow$ & IS $\uparrow$\\\midrule
    Adam \cite{kingma2014adam} & BigGAN \cite{brock2018large} & 14.73 & $\bm{9.22}$\\\midrule
    Adam \cite{kingma2014adam} & StyleGAN2 \cite{zhao2020differentiable} & 11.07 & 9.18\\\midrule
    Adam \cite{kingma2014adam} & SN-GAN \cite{miyato2018spectral} & $21.70 \pm 0.21$ & $7.60 \pm 0.06$\\\midrule
    Unrolled GAN \cite{metz2016unrolled,chavdarova2020taming}& SN-GAN \cite{miyato2018spectral} & $17.51\pm1.08$&------\\\midrule
    Extra-Adam \cite{gidel2018variational,mertikopoulos2018optimistic,chavdarova2020taming}& SN-GAN \cite{miyato2018spectral}& $15.47\pm1.82$&------\\\midrule
    LEAD$^\dagger$ \cite{hemmat2020lead}& SN-GAN \cite{miyato2018spectral} &$14.45\pm 0.45$&------\\\midrule
    LA-AltGAN \cite{chavdarova2020taming}& SN-GAN \cite{miyato2018spectral} &$12.67\pm0.57$& $8.55 \pm 0.04$\\\midrule
    ODE-GAN(RK4) \cite{qin2020training}& SN-GAN \cite{miyato2018spectral} & $11.85\pm0.21$ & $8.61 \pm 0.06$ \\\midrule
    Lv.$6$ Adam & SN-GAN \cite{miyato2018spectral}& $\bm{10.17\pm 0.16}$ & $\bm{8.78 \pm 0.06}$\\\midrule\bottomrule
    \end{tabular}
    \label{tab:cifar10 fid}
\end{table}
\begin{figure}
\centering
\makebox[\textwidth][c]{\includegraphics[width=1.03\textwidth]{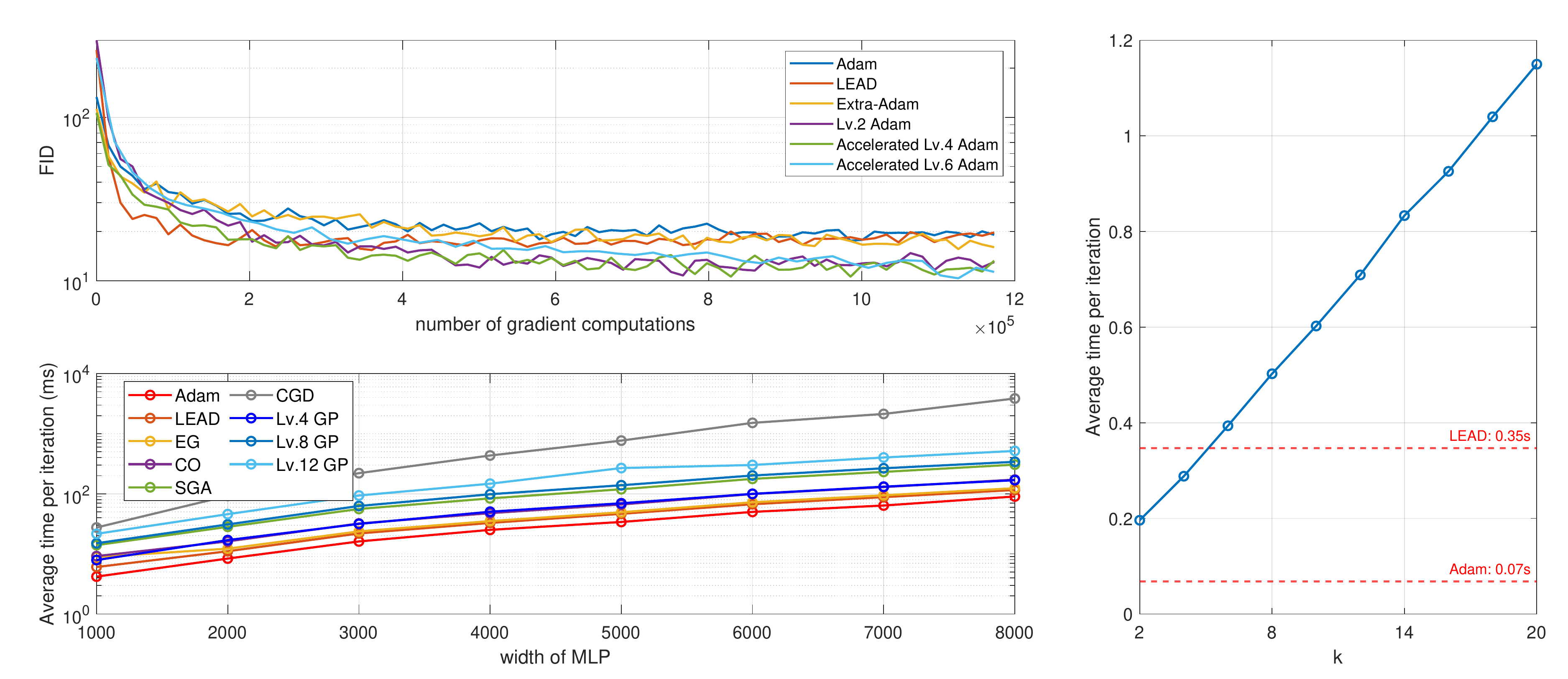}}
\caption{\textbf{Top Left}: Change of FID scores over 1.2 million gradient computations for Adam, LEAD, Extra-Adam, Lv.$2$, Lv.$4$ and Lv.$6$ Adam on CIFAR-10 with SNGAN. Note for Lv.$4$ and Lv.$6$ Adam, we use the accelerated implementation introduced in Appendix \ref{appendix: cifar10 and stl10}. \textbf{Bottom Left}: Average computational cost per iteration on 8-Gaussians experiment for MLPs with different widths. \textbf{Right}: Average computational cost per iteration on CIFAR-10 for Lv.$k$-Adam with different $k$. The values for LEAD (2.88) and Adam (14.65) are highlighted by dashed line.} 
\label{fig:Efficiency of Lv.k GP}
\end{figure}
\vspace{-4mm}
\subsection{Image Generation Experiments}\vspace{-2mm}
% We evaluate the effectiveness of our Lv.$k$ Adam algorithm on the task of CIFAR-10 \cite{krizhevsky2009learning} image generation on the SN-GAN architecture similar to \cite{miyato2018spectral}. We use the Inception score (the higher the better) \cite{salimans2016improved} and the Fréchet Inception distance (the lower the better) \cite{heusel2017gans} as performance metrics for image synthesis. Moreover, we evaluate our method on an exponential moving average of the generator's parameters with averaging factor $\beta = 0.999$.

% For Lv.$k$ Adam, we use $\beta_1 = 0$ and $\beta_2 = 0.9$ for all experiments. We use different learning rates for the generator ($\eta_{\theta} = 4e-5$) and the discriminator ($\eta_{\phi} = 2e-4$). We train the GANs with batch size $128$ for $600$ epochs.
% We present our results in Table \ref{tab:cifar10 fid}. On SN-GAN, we compare the performance of Lv.$6$ Adam to that of other first-order and second-order optimization algorithms. As a further comparison, we compare the SN-GAN trained with Lv.$6$ Adam with state-of-the-art models trained with Adam optimizer. The best FID score and Inception score, $10.17\pm0.16$ and $8.78\pm0.06$, on SN-GAN are obtained with our Lv.$6$ Adam. We also outperform BigGAN and StyleGAN2 in terms of FID score. Notably, our model has 5.1M parameters in total, and we use minibatches of size 128, whereas BigGAN uses 158.3M parameters and 2048 samples per minibatch. 
We evaluate the effectiveness of our Lv.$k$ Adam algorithm on unconditional generation of CIFAR-10 \cite{krizhevsky2009learning}. We use the Inception score (the higher the better) \cite{salimans2016improved} and the Fréchet Inception distance (the lower the better) \cite{heusel2017gans} as performance metrics for image synthesis. 
For architecture, we use the SN-GAN architecture based on \cite{miyato2018spectral}. 
For baselines, we compare the performance of Lv.$6$ Adam to that of other first-order and second-order optimization algorithms with the same SN-GAN architecture, and to state-of-the-art models trained with Adam.
For Lv.$k$ Adam, we use $\beta_1 = 0$ and $\beta_2 = 0.9$ for all experiments. We use different learning rates for the generator ($\eta_{\theta} = 4e-5$) and the discriminator ($\eta_{\phi} = 2e-4$). We train Lv.$k$ Adam with batch size $128$ for $600$ epochs.
For testing, we use an exponential moving average of the generator's parameters with averaging factor $\beta = 0.999$.

In table \ref{tab:cifar10 fid}, we present the performance of our method and baselines. The best FID score and Inception score, $10.17\pm0.16$ and $8.78\pm0.06$, on SN-GAN are obtained with our Lv.$6$ Adam. We also outperform BigGAN and StyleGAN2 in terms of FID score. Notably, our model has 5.1M parameters in total, and is trained with a small batch size of 128, whereas BigGAN uses 158.3M parameters and a batch size of 2048. 

\vspace{-1mm}
\textbf{The effect of $k$ and losses:}
We evaluate values of $k=\{2,4,6\}$ on non-saturated loss \cite{goodfellow2014generative} (non-zero-sum formulation) and hinge loss \cite{lim2017geometric} (zero-sum formulation). The result is presented in Table \ref{tab: increasing k, nsloss and hinge loss}.
\begin{table}[]
\centering
\caption{FID scores for the different loss functions with $k=\{2,4,6\}.$}
\vspace{2mm}
\begin{tabular}{l l l l }
\toprule\midrule
    & Lv.$2$ Adam & Lv.$4$ Adam & Lv.$6$ Adam\\\midrule
Non-saturated loss & $11.33\pm 0.18$& $11.62\pm 0.25$& $10.93\pm 0.24$\\\midrule
Hinge loss & $10.68\pm 0.20$ & $10.33\pm 0.22$ & $10.17\pm 0.16$\\\midrule\bottomrule
\end{tabular}
\label{tab: increasing k, nsloss and hinge loss}
\end{table}
Remarkably, our experiments demonstrate that few steps of recursive reasoning can result in significant performance gain comparing to existing GAN optimizers. The gradual improvements in the FID scores justify the idea that better estimation of opponents' next move improves performance. Moreover, we observe performance gains in both zero-sum and non-zero-sum formulations which supplement our theoretical convergence guarantees in zero-sum games.

\vspace{-1mm}
\textbf{Experiment results on STL-10:} To test whether the proposed Lv.$k$ Adam optimizer works on higher resolution images, we evaluate its performance on the STL-10 dataset \cite{coates2011analysis} with $3\times48\times48$ resolutions. In our experiments, Lv.$6$ Adam obtained an averaged FID of $25.43\pm0.18$ which outperforms that of the Adam optimizer, $30.25\pm 0.26$, using the same SN-GAN architecture. 

\vspace{-1mm}
\textbf{Memory and computation cost:} Compared to SGD, Lv.$k$ Adam requires the same extra memory as the EG method (one additional set of parameters per player). The relative cost of one iteration versus SGD is a factor of $k$ and the computational cost increases linearly as $k$ increases, we illustrate this relationship in Figure \ref{fig:Efficiency of Lv.k GP} (right). We provide an accelerated version of Lv.$k$ Adam in A.8 which reduces the computation cost by half for $k>2$. 
In Figure~\ref{fig:Efficiency of Lv.k GP} (top-left), we compare the FID scores obtained by Lv.$k$ Adam, Adam, and LEAD on CIFAR-10 over the same number of gradient computations. LEAD, Lv4 Adam, and Lv6 Adam all outperform Adam in this experiment. Lv4 Adam outperforms LEAD after $2\times 10^5$ gradient computations. 
%Lv.k GP's computational cost increases almost linearly as k increases, as shown in Figure 5 (right). 
Our method is also compared with different algorithms on the 8-Gaussian problem in terms of its computational cost. On the same architecture with different widths, Figure~\ref{fig:Efficiency of Lv.k GP} (bottom-left) illustrates the wall-clock time per computation for different algorithms. We observe that the computational cost of Lv.$k$ Adam while being much lower than CGD, is similar to LEAD, SGA and CO which involve JVP operations.  
Each run on CIFAR-10 dataset takes $30\sim33$ hours on a Nvidia RTX3090 GPU. Each experiment on STL-10 takes $48\sim60$ hours on a Nvidia RTX3090 GPU.

\vspace{-2mm}
\section{Conclusion and Future Work}\vspace{-3mm}
This paper proposes a novel algorithm: Level $k$ gradient play, capable of reasoning about players' future strategies. We achieve an average FID score of 10.17 for unconditional image generation on CIFAR-10 dataset, allowing GAN training on common computing resources to reach state-of-the-art performance. Our results suggest that Lv.$k$ GP is a flexible add-on that can be easily attached to existing GAN optimizers (e.g., Adam) and provides noticeable gains in performance and stability. In future work, we will examine the effectiveness of our approach on more complicated GAN designs, such as Progressive GANs \cite{karras2017progressive} and StyleGANs \cite{karras2019style}, where optimization plays a more significant role. Additionally, we intend to examine the convergence property of Lv.$k$ GP in games with more than two player in the future.
\vspace{-4mm}
\paragraph{Broader Impact} Our work introduces a novel optimizer that improves the performance of GANs and may reduce the amount of hyperparameter tuning required by practitioners of generative modeling. Generative models have been used to create illegal content(a.k.a. deepfakes~\cite{deepfake}). There is risk of negative social impact resulting from malicious use of the proposed methods.
\vspace{-4mm}
\paragraph{Acknowledgement} This work was supported by a funding from a NSERC Alliance grant and Huawei Technologies Canada. ZL would like to thank Tianshi Cao for insightful discussions on algorithm and experiment design. We are grateful to Bolin Gao and Dian Gadjov for their support.
\bibliographystyle{plainnat}
\bibliography{reference.bib}  

\newpage

\newpage
\appendix
\section{Appendix}

\subsection{Proof of Theorem \ref{thm: convergence of lvk gp}}
\begin{proof}
To begin with, let us consider Lv.$1$ GP:
\begin{align}
    \bm{\theta}_{t}^{(1)} &= \bm{\theta}_{t} - \eta \nabla_{\bm{\theta}}f(\bm{\theta}_{t},\bm{\phi}_{t}^{(0)}) = \bm{\theta}_{t} - \eta \nabla_{\bm{\theta}}f(\bm{\theta}_{t},\bm{\phi}_{t})\nonumber\\
    \bm{\phi}_{t}^{(1)} &= \bm{\phi}_{t} - \eta \nabla_{\bm{\phi}}g(\bm{\theta}_{t}^{(0)},\bm{\phi}_{t}) = \bm{\phi}_{t} - \eta \nabla_{\bm{\phi}}g(\bm{\theta}_{t},\bm{\phi}_{t})\nonumber
\end{align}
The differences between $(\bm{\theta}_{t}^{(1)},\bm{\phi}_{t}^{(1)})$ and $(\bm{\theta}_{t}^{(0)},\bm{\phi}_{t}^{(0)})$ are:
\begin{align}
    \lVert\bm{\theta}_{t}^{(1)}-\bm{\theta}_{t}^{(0)} \rVert &= \eta\lVert \nabla_{\theta}f(\bm{\theta}_t,\bm{\phi}_t)\rVert,\nonumber\\
    \lVert\bm{\phi}_{t}^{(1)}-\bm{\phi}_{t}^{(0)} \rVert &= \eta\lVert \nabla_{\phi}f(\bm{\theta}_t,\bm{\phi}_t)\rVert.\nonumber
\end{align}
Recall our definition of $\Delta_{\max}$ we have:
\begin{equation}
    \lVert\bm{\theta}_{t}^{(1)}-\bm{\theta}_{t}^{(0)} \rVert + \lVert\bm{\phi}_{t}^{(1)}-\bm{\phi}_{t}^{(0)} \rVert \leq \eta\Delta_{\max}\label{eq: base difference}
\end{equation}
Then, with $\bm{\omega}_t = [\bm{\theta}_t,\bm{\phi}_t]^T$ and $\bm{\omega}_{t}^{(k)} = [\bm{\theta}_t^{(k)},\bm{\phi}_t^{(k)}]^T$, the differences between Lv.$2$ agents and Lv.$1$ agents are:
\begin{align}
    \lVert\bm{\theta}_{t}^{(2)}-\bm{\theta}_{t}^{(1)} \rVert &= \eta\lVert \nabla_{\theta}f(\bm{\theta}_t,\bm{\phi}_t^{(1)})-\nabla_{\theta}f(\bm{\theta}_t,\bm{\phi}_t^{(0)})\rVert\nonumber\\
    &\leq \eta L_{\theta\phi}\lVert \bm{\phi}_{t}^{(1)} - \bm{\phi}_{t}^{(0)} \rVert,\nonumber\\
    \lVert\bm{\phi}_{t}^{(2)}-\bm{\phi}_{t}^{(1)} \rVert &= \eta\lVert \nabla_{\phi}f(\bm{\theta}_t^{(1)},\bm{\phi}_t)-\nabla_{\phi}f(\bm{\theta}_t^{(0)},\bm{\phi}_t)\rVert\nonumber\\
    &\leq \eta L_{\phi\theta}\lVert \bm{\theta}_{t}^{(1)} - \bm{\theta}_{t}^{(0)} \rVert.\nonumber
\end{align}
Recall that $L:= \max\{L_{\theta\theta},L_{\theta\phi},L_{\phi\theta},L_{\phi\phi}\}$, using Equation (\ref{eq: base difference}) we have:
\begin{equation}
    \lVert\bm{\theta}_{t}^{(2)}-\bm{\theta}_{t}^{(1)} \rVert + \lVert\bm{\phi}_{t}^{(2)}-\bm{\phi}_{t}^{(1)} \rVert \leq \eta^2 L\Delta_{\max}
\end{equation}
Similarly, we can derive the differences between Lv.$3$ and Lv.$2$ agents:
\begin{align}
    \lVert\bm{\theta}_{t}^{(3)}-\bm{\theta}_{t}^{(2)} \rVert &= \eta\lVert \nabla_{\theta}f(\bm{\theta}_t,\bm{\phi}_t^{(2)})-\nabla_{\theta}f(\bm{\theta}_t,\bm{\phi}_t^{(1)})\rVert\nonumber\\
    &\leq \eta L_{\theta\phi}\lVert \bm{\phi}_{t}^{(2)} - \bm{\phi}_{t}^{(1)} \rVert\nonumber\\
    \lVert\bm{\phi}_{t}^{(3)}-\bm{\phi}_{t}^{(2)} \rVert &= \eta\lVert \nabla_{\phi}f(\bm{\theta}_t^{(2)},\bm{\phi}_t)-\nabla_{\phi}f(\bm{\theta}_t^{(1)},\bm{\phi}_t)\rVert\nonumber\\
    &\leq \eta L_{\phi\theta}\lVert \bm{\theta}_{t}^{(2)} - \bm{\theta}_{t}^{(1)} \rVert\nonumber\\
    \lVert\bm{\theta}_{t}^{(3)}-\bm{\theta}_{t}^{(2)} \rVert &+ \lVert\bm{\phi}_{t}^{(3)}-\bm{\phi}_{t}^{(2)} \rVert \leq \eta^3 L^2\Delta_{\max}
\end{align}
Consequently, the difference between any two consecutive states $k$ and $k-1$ are upper bounded by:
\begin{align}
    \lVert\bm{\theta}_{t}^{(k)}-\bm{\theta}_{t}^{(k-1)} \rVert &= \eta\lVert \nabla_{\theta}f(\bm{\theta}_t,\bm{\phi}_t^{(k-1)})-\nabla_{\theta}f(\bm{\theta}_t,\bm{\phi}_t^{(k-2)})\rVert\nonumber\\
    &\leq \eta L_{\theta\phi}\lVert \bm{\phi}_{t}^{(k-1)} - \bm{\phi}_{t}^{(k-2)} \rVert\nonumber\\
    \lVert\bm{\phi}_{t}^{(k)}-\bm{\phi}_{t}^{(k-1)} \rVert &= \eta\lVert \nabla_{\phi}f(\bm{\theta}_t^{(k-1)},\bm{\phi}_t)-\nabla_{\phi}f(\bm{\theta}_t^{(k-2)},\bm{\phi}_t)\rVert\nonumber\\
    &\leq \eta L_{\phi\theta}\lVert \bm{\theta}_{t}^{(k-1)} - \bm{\theta}_{t}^{(k-2)} \rVert\nonumber\\
    \lVert\bm{\theta}_{t}^{(k)}-\bm{\theta}_{t}^{(k-1)} \rVert &+ \lVert\bm{\phi}_{t}^{(k)}-\bm{\phi}_{t}^{(k-1)} \rVert \leq \eta \cdot (\eta L)^{(k-1)}\Delta_{\max}
\end{align}
Since $\lVert \bm{\omega}_{t}^{(k)} - \bm{\omega}_{t}^{(k-1)}\rVert \leq \lVert\bm{\theta}_{t}^{(k)}-\bm{\theta}_{t}^{(k-1)} \rVert+\lVert\bm{\phi}_{t}^{(k)}-\bm{\phi}_{t}^{(k-1)} \rVert$ we have:
\begin{equation}
    \lVert \bm{\omega}_{t}^{(k)}-\bm{\omega}_{t}^{(k-1)}\rVert \leq \eta\cdot (\eta L)^{(k-1)}\Delta_{\max}
\end{equation}
Suppose $\eta< (2L)^{-1}$, such that the difference between any two consecutive states is a contraction, then we consider the  difference, $\lVert \bm{\omega}_{t}^{(a)} - \bm{\omega}_{t}^{(b)}\rVert$, where $a>b>0$. We can rewrite it as:
\begin{align}
    \lVert \bm{\omega}_{t}^{(a)} - \bm{\omega}_{t}^{(b)}\rVert &= \left\lVert \sum_{i=b+1}^{a}\bm{\omega}_{t}^{(i)} - \bm{\omega}_{t}^{(i-1)}\right\rVert\nonumber\\
    &\leq\sum_{i=b+1}^{a}\left\lVert \bm{\omega}_{t}^{(i)} - \bm{\omega}_{t}^{(i-1)}\right\rVert\nonumber\\
    &\leq\sum_{i=b+1}^{a}\eta\cdot (\eta L)^{(i-1)}\Delta_{\max}\nonumber\\
    &\leq\eta\Delta_{\max}\cdot \left[(\eta L)^{(b)} + \dots + (\eta L)^{(a-1)}\right]\nonumber\\
    &\leq \eta\Delta_{\max}\cdot (\eta L)^{(b-1)}\nonumber\\
    &\leq \eta^{b} L^{(b-1)} \Delta_{\max} = \mathcal{O}(\eta^{b}).
\end{align}
Since $\eta<(2L)^{-1}$, we have that $\eta L<1$ and for any $\epsilon >0$, we can solve for $b$ such that $\eta^{b} L^{(b-1)} \Delta_{\max} < \epsilon$. Therefore the sequence $\{\bm{\omega}_{t}^{k}\}_{k=0}^{\infty}$ is a Cauchy sequence. Moreover, in a complete space, every Cauchy sequence has a limit: $\lim_{k\to\infty}\bm{\omega}_{t}^{(k)} = \bm{\omega}_{t}^{*}$
\end{proof}
\subsection{Proof of Theorem \ref{thm: convergence in nonconvex gamae}}\label{appendix: prove convergence in nonconvex games}
\begin{theorem}
Consider the (\ref{eq: minimax problem}) problem under Assumption \ref{assumption: Lipschitz gradient assumption} and Lv.$k$ GP. Let $(\bm{\theta}^*,\bm{\phi}^*)$ be a stationary point. Suppose $\bm{\theta}_{t}-\bm{\theta}^*$ not in kernel of $\nabla_{\phi\theta}f(\bm{\theta}^*,\bm{\phi}^*)$, $\bm{\phi}_{t}-\bm{\phi}^*$ not in kernel of $\nabla_{\theta\phi}f(\bm{\theta}^*,\bm{\phi}^*)$ and $\eta<(L)^{-1}$. There exists a neighborhood $\mathcal{U}$ of $(\bm{\theta}^*,\bm{\phi}^*)$ such that if SPPM started at $(\bm{\theta}_{0},\bm{\phi}_{0})\in\mathcal{U}$, the iterates $\{\bm{\theta}_{t},\bm{\phi}_{t}\}_{t\geq0}$ generated by SPPM satisfy:
\begin{equation}
    \lVert\bm{\theta}_{t+1} - \bm{\theta}^*\rVert^2 + \lVert\bm{\phi}_{t+1} - \bm{\phi}^*\rVert^2\leq \frac{\rho^2(\bm{I}-\eta\nabla_{\theta\theta}f^*)\lVert\bm{\theta}_{t} - \bm{\theta}^*\rVert^2+\rho^2(\bm{I}+\eta\nabla_{\phi\phi}f^*)\lVert\bm{\phi}_{t} - \bm{\phi}^*\rVert^2}{1+\eta^2\lambda_{\min}(\nabla_{\theta\phi}f^*\nabla_{\phi\theta}f^*)}\nonumber
\end{equation}
where $f^* = f(\bm{\theta}^*,\bm{\phi}^*)$. Moreover, for any $\eta$ satisfying:
\begin{equation}
    \frac{\max(\rho^2(\bm{I}-\eta\nabla_{\theta\theta}f^*),\rho^2(\bm{I}+\eta\nabla_{\phi\phi}f^*))}{1+\eta^2\lambda_{\min}(\nabla_{\theta\phi}f^*\nabla_{\phi\theta}f^*)}< 1\label{eq:local convergence},
\end{equation}
SPPM converges asymptotically to $(\bm{\theta}^*,\bm{\phi}^*)$.
\end{theorem}
\begin{proof}
Consider the learning dynamics:
\begin{align}
    \bm{\theta}_{t+1} &= \bm{\theta}_{t} - \eta\nabla_{\theta}f(\bm{\theta}_{t},\bm{\phi}_{t+1})\nonumber\\
    \bm{\phi}_{t+1} &= \bm{\phi}_{t} + \eta\nabla_{\theta}f(\bm{\theta}_{t+1},\bm{\phi}_{t})\nonumber
\end{align}
Let us define
\begin{align}
    \hat{\bm{\theta}}_{t} = \bm{\theta}_{t} - \bm{\theta}^*\nonumber\\
    \hat{\bm{\phi}}_{t} = \bm{\phi}_{t} - \bm{\phi}^*\nonumber
\end{align}
It follows immediately by linearizing the system about the stationary point $(\bm{\theta}^*,\bm{\phi}^*)$ that
\begin{align}
    \begin{bmatrix}
    \hat{\bm{\theta}}_{t+1}\\
    \hat{\bm{\phi}}_{t+1}
    \end{bmatrix} \simeq&
    \begin{bmatrix}
    \bm{I}-\eta\nabla_{\theta\theta}^2f(\bm{\theta}^*,\bm{\phi}^*) & \bm{0}\\
    \bm{0} & \bm{I} + \eta\nabla_{\phi\phi}^2f(\bm{\theta}^*,\bm{\phi}^*)
    \end{bmatrix}
    \begin{bmatrix}
    \hat{\bm{\theta}}_{t}\\
    \hat{\bm{\phi}}_{t}
    \end{bmatrix} \nonumber\\
    &+ 
    \begin{bmatrix}
    \bm{0} & -\eta\nabla_{\theta\phi}^2f(\bm{\theta}^*,\bm{\phi}^*)\\
    \eta\nabla_{\theta\phi}^2f(\bm{\theta}^*,\bm{\phi}^*) & \bm{0}
    \end{bmatrix}
    \begin{bmatrix}
    \hat{\bm{\theta}}_{t+1}\\
    \hat{\bm{\phi}}_{t+1}
    \end{bmatrix}\nonumber
\end{align}
Let us denote the Jacobian by
\begin{align}
    \begin{bmatrix}
    -\nabla_{\theta\theta}^2f(\bm{\theta}^*,\bm{\phi}^*) & -\nabla_{\theta\phi}^2f(\bm{\theta}^*,\bm{\phi}^*)\\
    \nabla_{\phi\theta}^2f(\bm{\theta}^*,\bm{\phi}^*) & \nabla_{\phi\phi}^2f(\bm{\theta}^*,\bm{\phi}^*)
    \end{bmatrix} = \begin{bmatrix}
    -\bm{A} & -\bm{B}\\
    \bm{B}^T & \bm{C}
    \end{bmatrix}
\end{align}
Then we can rewrite the dynamics around the stationary point as 
\begin{align}
    \hat{\bm{\theta}}_{t+1} &= \hat{\bm{\theta}}_{t} - \eta \bm{A}\hat{\bm{\theta}}_{t} - \eta \bm{B}\hat{\bm{\phi}}_{t+1}\nonumber\\
    \hat{\bm{\theta}}_{t+1} &= \hat{\bm{\theta}}_{t} - \eta \bm{A}\hat{\bm{\theta}}_{t} - \eta \bm{B}(\hat{\bm{\phi}}_{t} + \eta \bm{B}^T\hat{\bm{\theta}}_{t+1} + \eta \bm{C}\hat{\bm{\phi}}_{t})\nonumber\\
    (\bm{I} + \eta^2\bm{BB}^T) \hat{\bm{\theta}}_{t+1} &= (\bm{I} - \eta \bm{A})\hat{\bm{\theta}}_{t} - \eta\bm{B}(\bm{I} + \eta\bm{C})\bm{\phi}_{t}\nonumber\\
    \hat{\bm{\theta}}_{t+1} &= (\bm{I} + \eta^2\bm{BB}^T)^{-1}\left[(\bm{I} - \eta \bm{A})\hat{\bm{\theta}}_{t} - \eta\bm{B}(\bm{I} + \eta\bm{C})\bm{\phi}_{t}\right]\label{eq: linearized xt+1}
\end{align}
Similarly, for the other player we have
\begin{align}
    \hat{\bm{\phi}}_{t+1} &= \hat{\bm{\phi}}_{t} + \eta \bm{B}^T\hat{\bm{\theta}}_{t+1} + \eta \bm{C}\hat{\bm{\phi}}_{t}\nonumber\\
    \hat{\bm{\phi}}_{t+1} &= \hat{\bm{\phi}}_{t} + \eta \bm{B}^T(\hat{\bm{\theta}}_{t} - \eta \bm{A}\hat{\bm{\theta}}_{t} - \eta \bm{B}\hat{\bm{\phi}}_{t+1}) + \eta \bm{C}\hat{\bm{\phi}}_{t}\nonumber\\
    (\bm{I} + \eta^2\bm{B}^T\bm{B}) \hat{\bm{\phi}}_{t+1} &= \eta\bm{B}^T(\bm{I} - \eta \bm{A})\hat{\bm{\theta}}_{t} + (\bm{I} + \eta\bm{C})\bm{\phi}_{t}\nonumber\\
    \hat{\bm{\phi}}_{t+1} &= (\bm{I} + \eta^2\bm{B}^T\bm{B})^{-1}\left[ \eta\bm{B}^T(\bm{I} - \eta \bm{A})\hat{\bm{\theta}}_{t} + (\bm{I} + \eta\bm{C})\bm{\phi}_{t}\right]\label{eq: linearized yt+1}
\end{align}
Let us define the symmetric matrices $\bm{Q}_{\bm{\theta}} = (\bm{I} + \eta^2\bm{BB}^T)^{-1}$, $\bm{Q}_{\bm{\phi}} = (\bm{I} + \eta^2\bm{B}^T\bm{B})^{-1}$ and $\bm{P}_{\bm{\theta}} = (\bm{I} - \eta\bm{A})$, $\bm{P}_{\bm{\phi}} = (\bm{I} + \eta\bm{C})$. Further we define $r_{t} = \lVert \hat{\bm{\theta}}_{t+1}\rVert ^2 + \lVert \hat{\bm{\phi}}_{t+1}\rVert ^2$. Based on these definitions, and the expressions in (\ref{eq: linearized xt+1}) and (\ref{eq: linearized yt+1}) we have
\begin{align}
    \lVert \bm{\theta}_{t+1}\rVert ^2 + \lVert \bm{\phi}_{t+1}\rVert ^2
    = \lVert \bm{Q}_{\bm{\theta}}\bm{P}_{\bm{\theta}}\hat{\bm{\theta}}_{t}\rVert^2 &+ \eta^2\lVert \bm{Q}_{\bm{\theta}}\bm{B}\bm{P}_{\bm{\phi}}\hat{\bm{\phi}}_{t}\rVert^2 + \lVert \bm{Q}_{\bm{\phi}}\bm{B}^T\bm{P}_{\bm{\theta}}\hat{\bm{\theta}}_{t}\rVert^2 + \lVert \bm{Q}_{\bm{\phi}}\bm{P}_{\bm{\phi}}\hat{\bm{\phi}}_{t}\rVert^2\nonumber\\
    &-2\eta\hat{\bm{\theta}}_{t}^T\bm{P}_{\bm{\theta}}^T\bm{Q}_{\bm{\theta}}^T\bm{Q}_{\bm{\theta}}\bm{B}\bm{P}_{\bm{\phi}}\hat{\bm{\phi}}_{t}+2\eta\hat{\bm{\phi}}_{t}^T\bm{P}_{\bm{\phi}}^T\bm{Q}_{\bm{\phi}}^T\bm{Q}_{\bm{\phi}}\bm{B}^T\bm{P}_{\bm{\theta}}\hat{\bm{\theta}}_{t}\label{eq: sum of norm}
\end{align}
To simplify the expression in (\ref{eq: sum of norm}) we use the following lemma:
\begin{lemma}\label{lemma: Q}
The matrices $\bm{Q}_{\bm{\theta}} = (\bm{I} + \eta^2\bm{BB}^T)^{-1}$, $\bm{Q}_{\bm{\phi}} = (\bm{I} + \eta^2\bm{B}^T\bm{B})^{-1}$ satisfy the following properties:
\begin{align}
    \bm{Q}_{\bm{\theta}}\bm{B} &= \bm{B}\bm{Q}_{\bm{\phi}}\label{eq:1}\\
    \bm{Q}_{\bm{\phi}}\bm{B}^T &= \bm{B}^T\bm{Q}_{\bm{\theta}}\label{eq:2}
\end{align}
\end{lemma}
Using this lemma, we can show that
\begin{equation}
    \hat{\bm{\theta}}_{t}^T\bm{P}_{\bm{\theta}}^T\bm{Q}_{\bm{\theta}}^T\bm{Q}_{\bm{\theta}}\bm{B}\bm{P}_{\bm{\phi}}\hat{\bm{\phi}}_{t}=\hat{\bm{\theta}}_{t}^T\bm{P}_{\bm{\theta}}^T\bm{Q}_{\bm{\theta}}^T\bm{B}\bm{Q}_{\bm{\phi}}\bm{P}_{\bm{\phi}}\hat{\bm{\phi}}_{t}=\hat{\bm{\phi}}_{t}^T\bm{P}_{\bm{\phi}}^T\bm{Q}_{\bm{\phi}}^T\bm{B}^T\bm{Q}_{\bm{\theta}}\bm{P}_{\bm{\theta}}\hat{\bm{\theta}}_{t}=\hat{\bm{\phi}}_{t}^T\bm{P}_{\bm{\phi}}^T\bm{Q}_{\bm{\phi}}^T\bm{Q}_{\bm{\phi}}\bm{B}^T\bm{P}_{\bm{\theta}}\hat{\bm{\theta}}_{t}\nonumber
\end{equation}
where the intermediate equality holds as $\bm{a}^T\bm{b} = \bm{b}^T\bm{a}$. Hence, the expression in (\ref{eq: sum of norm}) can be simplified as
\begin{equation}
    \lVert \hat{\bm{\theta}}_{t+1}\rVert ^2 + \lVert \hat{\bm{\phi}}_{t+1}\rVert ^2
    = \lVert \bm{Q}_{\bm{\theta}}\bm{P}_{\bm{\theta}}\hat{\bm{\theta}}_{t}\rVert^2 + \eta^2\lVert \bm{Q}_{\bm{\theta}}\bm{B}\bm{P}_{\bm{\phi}}\hat{\bm{\phi}}_{t}\rVert^2 + \lVert \bm{Q}_{\bm{\phi}}\bm{B}^T\bm{P}_{\bm{\theta}}\hat{\bm{\theta}}_{t}\rVert^2 + \lVert \bm{Q}_{\bm{\phi}}\bm{P}_{\bm{\phi}}\hat{\bm{\phi}}_{t}\rVert^2\label{eq: norm of sum clean}
\end{equation}
We simplify equation (\ref{eq: norm of sum clean}) as follows. Consider the term involving $\hat{\bm{\theta}}_{t}$. We have
\begin{align}
    \lVert \bm{Q}_{\bm{\theta}}\bm{P}_{\bm{\theta}}\hat{\bm{\theta}}_{t}\rVert^2 + \eta^2\lVert \bm{Q}_{\bm{\phi}}\bm{B}^T\bm{P}_{\bm{\theta}}\hat{\bm{\theta}}_{t}\rVert^2 &= \hat{\bm{\theta}}_{t}^T\bm{P}_{\bm{\theta}}^T\bm{Q}_{\bm{\theta}}^2\bm{P}_{\bm{\theta}}\hat{\bm{\theta}}_{t} + \eta^2\hat{\bm{\theta}}_{t}^T\bm{P}_{\bm{\theta}}^T\bm{B}\bm{Q}_{\bm{\phi}}^2\bm{B}^T\bm{P}_{\bm{\theta}}\hat{\bm{\theta}}_{t}\nonumber\\
    &= \hat{\bm{\theta}}_{t}^T\bm{P}_{\bm{\theta}}^T(\bm{Q}_{\bm{\theta}}^2 + \eta^2\bm{B}\bm{Q}_{\bm{\phi}}^2\bm{B}^T)\bm{P}_{\bm{\theta}}\hat{\bm{\theta}}_{t}\nonumber\\
    &= \hat{\bm{\theta}}_{t}^T\bm{P}_{\bm{\theta}}^T(\bm{Q}_{\bm{\theta}}^2 + \eta^2\bm{B}\bm{Q}_{\bm{\phi}}\bm{B}^T\bm{Q}_{\bm{\theta}})\bm{P}_{\bm{\theta}}\hat{\bm{\theta}}_{t}\nonumber\\
    &= \hat{\bm{\theta}}_{t}^T\bm{P}_{\bm{\theta}}^T(\bm{Q}_{\bm{\theta}}^2 + \eta^2\bm{B}\bm{B}^T\bm{Q}_{\bm{\theta}}\bm{Q}_{\bm{\theta}})\bm{P}_{\bm{\theta}}\hat{\bm{\theta}}_{t}\nonumber\\
    &= \hat{\bm{\theta}}_{t}^T\bm{P}_{\bm{\theta}}^T(\bm{I} + \eta^2\bm{B}\bm{B}^T)\bm{Q}_{\bm{\theta}}^2\bm{P}_{\bm{\theta}}\hat{\bm{\theta}}_{t}\nonumber\\
    &= \hat{\bm{\theta}}_{t}^T\bm{P}_{\bm{\theta}}^T(\bm{I} + \eta^2\bm{B}\bm{B}^T)^{-1}\bm{P}_{\bm{\theta}}\hat{\bm{\theta}}_{t}\label{eq: expression for xt}
\end{align}
where the last equality follows by replacing $\bm{Q}_{\bm{\theta}}$ by its definition. The same procedure follows for the term involving $\bm{\phi}_{t}$ which leads to the expression
\begin{equation}
    \lVert \bm{Q}_{\bm{\phi}}\bm{P}_{\bm{\phi}}\hat{\bm{\phi}}_{t}\rVert^2 + \eta^2\lVert \bm{Q}_{\bm{\theta}}\bm{B}\bm{P}_{\bm{\phi}}\hat{\bm{\phi}}_{t}\rVert^2 = \hat{\bm{\phi}}_{t}^T\bm{P}_{\bm{\phi}}^T(\bm{I} + \eta^2\bm{B}^T\bm{B})^{-1}\bm{P}_{\bm{\phi}}\hat{\bm{\phi}}_{t}. \label{eq: expression for yt}
\end{equation}
Substitute $\lVert \bm{Q}_{\bm{\theta}}\bm{P}_{\bm{\theta}}\hat{\bm{\theta}}_{t}\rVert^2 + \eta^2\lVert \bm{Q}_{\bm{\phi}}\bm{B}^T\bm{P}_{\bm{\theta}}\hat{\bm{\theta}}_{t}\rVert^2$ and $\lVert \bm{Q}_{\bm{\phi}}\bm{P}_{\bm{\phi}}\hat{\bm{\phi}}_{t}\rVert^2 + \eta^2\lVert \bm{Q}_{\bm{\theta}}\bm{B}\bm{P}_{\bm{\phi}}\hat{\bm{\phi}}_{t}\rVert^2 $ in (\ref{eq: norm of sum clean}) with the expressions in (\ref{eq: expression for xt}) and (\ref{eq: expression for yt}), respectively, to obtain
\begin{equation}
    \lVert \hat{\bm{\theta}}_{t+1}\rVert^2 + \lVert\hat{\bm{\phi}}_{t+1} \rVert^2 = \hat{\bm{\theta}}_{t}^T\bm{P}_{\bm{\theta}}^T(\bm{I} + \eta^2\bm{B}\bm{B}^T)^{-1}\bm{P}_{\bm{\theta}}\hat{\bm{\theta}}_{t} + \hat{\bm{\phi}}_{t}^T\bm{P}_{\bm{\phi}}^T(\bm{I} + \eta^2\bm{B}^T\bm{B})^{-1}\bm{P}_{\bm{\phi}}\hat{\bm{\phi}}_{t}.\label{eq: sum of norm result}
\end{equation}

Note that, we assume that the trajectory $\{\hat{\bm{\theta}}_{t},\hat{\bm{\phi}}_{t}\}_{t\geq 0}$ is not in the kernel of $\bm{B}\bm{B}^T$ and $\bm{B}^T\bm{B}$, thus $\bm{B}\bm{B}^T\hat{\bm{\theta}}_{t}\neq 0$ and $\bm{B}^T\bm{B}\hat{\bm{\phi}}_{t}\neq 0$.
Now using the expression in (\ref{eq: sum of norm result}) and the fact that $\bm{P}_{\bm{\theta}} = \bm{P}_{\bm{\theta}}^T$, $\bm{P}_{\bm{\phi}} = \bm{P}_{\bm{\phi}}^T$ and $\bm{B}\bm{B}^T$ and $\bm{B}^T\bm{B}$ have the same set of non-zero eigenvalues, if we denote the minimum non-zero eigenvalues by $\lambda_{\min}(\bm{B}\bm{B}^T)$ and $\lambda_{\min}(\bm{B}^T\bm{B})$, we can write
\begin{equation}
    \lVert\bm{\theta}_{t+1} - \bm{\theta}^*\rVert^2 + \lVert\bm{\phi}_{t+1} - \bm{\phi}^*\rVert^2\leq \frac{\rho^2(\bm{I}-\eta\bm{A})\lVert\bm{\theta}_{t} - \bm{\theta}^*\rVert^2+\rho^2(1+\eta\bm{C})\lVert\bm{\phi}_{t} - \bm{\phi}^*\rVert^2}{\bm{I}+\eta^2\lambda_{\min}(\bm{B}^T\bm{B})}.\nonumber
\end{equation}
Replacing $\lVert\bm{\theta}_{t+1} - \bm{\theta}^*\rVert^2 + \lVert\bm{\phi}_{t+1} - \bm{\phi}^*\rVert^2$ and $\lVert\bm{\theta}_{t} - \bm{\theta}^*\rVert^2 + \lVert\bm{\phi}_{t} - \bm{\phi}^*\rVert^2$ with $r_{t+1}$ and $r_{t}$ we have:
\begin{equation}
    r_{t+1}\leq \frac{\max(\rho^2(\bm{I}-\eta\bm{A}),\rho^2(\bm{I}+\eta\bm{C}))}{1+\eta^2\lambda_{\min}(\bm{B}^T\bm{B})} r_{t}.\nonumber
\end{equation}
Recall that $\bm{A} = \nabla_{\theta\theta}f(\bm{\theta}^*,\bm{\phi}^*)$, $\bm{B} = \nabla_{\theta\phi}f(\bm{\theta}^*,\bm{\phi}^*)$ and $\bm{C} = \nabla_{\phi\phi}f(\bm{\theta}^*,\bm{\phi}^*)$, therefore for any $\eta$ satisfying that:
\begin{equation}
    \frac{\max(\rho^2(\bm{I}-\eta\nabla_{\theta\theta}f(\bm{\theta}^*,\bm{\phi}^*)),\rho^2(\bm{I}+\eta\nabla_{\phi\phi}f(\bm{\theta}^*,\bm{\phi}^*)))}{1+\eta^2\lambda_{\min}(\nabla_{\theta\phi}f(\bm{\theta}^*,\bm{\phi}^*)\nabla_{\phi\theta}f(\bm{\theta}^*,\bm{\phi}^*))}< 1,
\end{equation}
we have $r_{t+1}< r_{t}$. Since we linearize the system about the stationary point $(\bm{\theta}^*,\bm{\phi}^*)$, there exists a neighborhood $\mathcal{U}$ around the stationary point, such that, SPPM started at $(\bm{\theta}_{0},\bm{\phi}_{0})\in\mathcal{U}$ converges asymptotically to $(\bm{\theta}^*,\bm{\phi}^*)$.
\end{proof}
\subsection{Proof of Remark \ref{remark: convergence of Lvk GP}}
\begin{proof}
To prove the local convergence of Lv.$k$ GP in non-convex non-concave games, we first consider the update rule of Lv.$k$ GP:
\begin{equation}
\mathmakebox[0.8\textwidth]{
\text{Reasoning:}
\begin{cases}
        \bm{\theta}^{(k)}_{t} = \bm{\theta}_{t} - \eta\nabla_{\bm{\theta}}f(\bm{\theta}_{t},\bm{\phi}_{t}^{(k-1)})\nonumber\\
        \bm{\phi}^{(k)}_{t} = \bm{\phi}_{t} - \eta\nabla_{\bm{\phi}}g(\bm{\theta}_{t}^{(k-1)},\bm{\phi}_{t})
\end{cases}
\text{Update:}
\begin{cases}
    \bm{\theta}_{t+1} = \bm{\theta}_{t}^{(k)}\nonumber\\
    \bm{\phi}_{t+1} = \bm{\phi}_{t}^{(k)}
\end{cases}}
\end{equation}
Similar to Section \ref{appendix: prove convergence in nonconvex games} let us denote
\begin{align}
    \begin{bmatrix}
    -\nabla_{\theta\theta}^2f(\bm{\theta}^*,\bm{\phi}^*) & -\nabla_{\theta\phi}^2f(\bm{\theta}^*,\bm{\phi}^*)\\
    \nabla_{\phi\theta}^2f(\bm{\theta}^*,\bm{\phi}^*) & \nabla_{\phi\phi}^2f(\bm{\theta}^*,\bm{\phi}^*)
    \end{bmatrix} = \begin{bmatrix}
    -\bm{A} & -\bm{B}\\
    \bm{B}^T & \bm{C}
    \end{bmatrix}\nonumber
\end{align}
and we define the difference between states and stationary points as
\begin{align}
\hat{\bm{\theta}}^{(k)}_{t} = \bm{\theta}^{(k)}_{t} - \bm{\theta}^* &\text{ and } \hat{\bm{\theta}}_{t} = \bm{\theta}_{t} - \bm{\theta}^*\nonumber\\
\hat{\bm{\phi}}^{(k)}_{t} = \bm{\phi}^{(k)}_{t} - \bm{\phi}^* &\text{ and } \hat{\bm{\phi}}_{t} = \bm{\phi}_{t} - \bm{\phi}^*\nonumber
\end{align}
Linearizing the dynamical system induced by Lv.$k$ GP about the stationary point $(\bm{\theta}^*,\bm{\phi}^*)$ we get:
\begin{align}
    \begin{cases}
        \hat{\bm{\theta}}_{t+1} &= \hat{\bm{\theta}}^{(k)}_{t} = (\bm{I} - \eta\bm{A})\hat{\bm{\theta}}_{t} - \eta\bm{B}\hat{\bm{\phi}}_{t}^{(k-1)}\nonumber\\
        \hat{\bm{\phi}}_{t+1} &= \hat{\bm{\phi}}^{(k)}_{t} = \eta\bm{B}^T\hat{\bm{\theta}}_{t}^{(k-1)} + (\bm{I} + \eta\bm{C})\hat{\bm{\phi}}_{t}\nonumber
    \end{cases}
\end{align}
Note, in Lv.$k$ GP, we define $\bm{\theta}^{(0)}_{t} = \bm{\theta}_{t}$ and $\bm{\phi}^{(0)}_{t} = \bm{\phi}_{t}$, thus for Lv.$1$ GP, we have:
\begin{align}
    \begin{cases}
         \hat{\bm{\theta}}^{(1)}_{t} = (\bm{I} - \eta\bm{A})\hat{\bm{\theta}}_{t} - \eta\bm{B}\hat{\bm{\phi}}_{t}\nonumber\\
         \hat{\bm{\phi}}^{(1)}_{t} = \eta\bm{B}^T\hat{\bm{\theta}}_{t} + (\bm{I} + \eta\bm{C})\hat{\bm{\phi}}_{t}\nonumber
    \end{cases}    
\end{align}
For Lv.$2$ GP, we have:
\begin{align}
    \begin{cases}
         \hat{\bm{\theta}}^{(2)}_{t} = (\bm{I} - \eta\bm{A})\hat{\bm{\theta}}_{t} - \eta\bm{B}\hat{\bm{\phi}}_{t}^{(1)}\nonumber\\
         \hat{\bm{\phi}}^{(2)}_{t} = \eta\bm{B}^T\hat{\bm{\theta}}_{t}^{(1)} + (\bm{I} + \eta\bm{C})\hat{\bm{\phi}}_{t}\nonumber
    \end{cases}
\end{align}
Substituting $\hat{\bm{\theta}}^{(1)}_{t}$ and $\hat{\bm{\phi}}^{(1)}_{t}$ into the update rule above we get:
\begin{align}
    \begin{cases}
        \hat{\bm{\theta}}^{(2)}_{t} = (\bm{I} - \eta\bm{A})\hat{\bm{\theta}}_{t} - \eta\bm{B}(\bm{I}+\eta\bm{C})\hat{\bm{\phi}}_{t} -\eta^2 \bm{BB}^T \hat{\bm{\theta}}_{t}\nonumber\\
        \hat{\bm{\phi}}^{(2)}_{t} = \eta\bm{B}^T(\bm{I}-\eta\bm{A})\hat{\bm{\theta}}_{t} + (\bm{I} + \eta\bm{C})\hat{\bm{\phi}}_{t} - \eta^2\bm{B}^T\bm{B}\hat{\bm{\phi}}_{t}\nonumber
    \end{cases}
\end{align}
Similarly, for Lv.$3$ and Lv.$4$ GP we have:
\begin{align}
    \begin{cases}
        \hat{\bm{\theta}}^{(3)}_{t} = (\bm{I} - \eta^2\bm{B}\bm{B}^T)(\bm{I} - \eta\bm{A})\hat{\bm{\theta}}_{t} - \eta\bm{B}(\bm{I}+\eta\bm{C})\hat{\bm{\phi}}_{t} +\eta^3 \bm{BB}^T\bm{B} \hat{\bm{\phi}}_{t}\nonumber\\
        \hat{\bm{\phi}}^{(3)}_{t} = \eta\bm{B}^T(\bm{I}-\eta\bm{A})\hat{\bm{\theta}}_{t} + (\bm{I} - \eta^2\bm{B}^T\bm{B})(\bm{I} + \eta\bm{C})\hat{\bm{\phi}}_{t} - \eta^3\bm{B}\bm{B}^T\bm{B}\hat{\bm{\theta}}_{t}\nonumber
    \end{cases}
\end{align}
and
\begin{align}
    \begin{cases}
        \hat{\bm{\theta}}^{(4)}_{t} = (\bm{I} - \eta^2\bm{B}\bm{B}^T)\left[(\bm{I} - \eta\bm{A})\hat{\bm{\theta}}_{t} - \eta\bm{B}(\bm{I}+\eta\bm{C})\hat{\bm{\phi}}_{t}\right] +\eta^4 \bm{BB}^T\bm{BB}^T \hat{\bm{\theta}}_{t}\nonumber\\
        \hat{\bm{\phi}}^{(4)}_{t} =(\bm{I} - \eta^2\bm{B}^T\bm{B})\left[ \eta\bm{B}^T(\bm{I}-\eta\bm{A})\hat{\bm{\theta}}_{t} + (\bm{I} + \eta\bm{C})\hat{\bm{\phi}}_{t}\right] + \eta^4\bm{B}^T\bm{B}\bm{B}^T\bm{B}\hat{\bm{\phi}}_{t}\nonumber
    \end{cases}
\end{align}
Summarizing the equations above we have that for Lv.$2k$ GP, its update can be written as:
\begin{align}
    \begin{cases}
        \hat{\bm{\theta}}^{(2k)}_{t} = (\sum_{i=0}^{k-1}(-\eta^2\bm{B}\bm{B}^T)^k)\left[(\bm{I} - \eta\bm{A})\hat{\bm{\theta}}_{t} - \eta\bm{B}(\bm{I}+\eta\bm{C})\hat{\bm{\phi}}_{t}\right] +(-\eta^2 \bm{BB}^T)^{k} \hat{\bm{\theta}}_{t}\nonumber\\
        \hat{\bm{\phi}}^{(2k)}_{t} =(\sum_{i=0}^{k-1}(-\eta^2\bm{B}^T\bm{B})^k)\left[ \eta\bm{B}^T(\bm{I}-\eta\bm{A})\hat{\bm{\theta}}_{t} + (\bm{I} + \eta\bm{C})\hat{\bm{\phi}}_{t}\right] + (-\eta^2\bm{B}^T\bm{B})^{k}\hat{\bm{\phi}}_{t}\nonumber
    \end{cases}
\end{align}
Similar to Appendix \ref{appendix: prove convergence in nonconvex games}, let us define $\bm{Q}_{\bm{\theta}} = (\bm{I} + \eta^2\bm{BB}^T)^{-1}$, $\bm{Q}_{\bm{\phi}} = (\bm{I} + \eta^2\bm{B}^T\bm{B})^{-1}$ and $\bm{P}_{\bm{\theta}} = (\bm{I} - \eta\bm{A})$, $\bm{P}_{\bm{\phi}} = (\bm{I} + \eta\bm{C})$.
Further, we define $\bm{R}^{(k)}_{\bm{\theta}} = (\sum_{i=0}^{k-1}(-\eta^2\bm{B}\bm{B}^T)^k)$, $\bm{R}^{(k)}_{\bm{\phi}} = (\sum_{i=0}^{k-1}(-\eta^2\bm{B}^T\bm{B})^k)$ and $\bm{E}^{(k)}_{\bm{\theta}} = \bm{R}^{(k)}_{\bm{\theta}} - \bm{Q}_{\bm{\theta}}$, $\bm{E}^{(k)}_{\bm{\phi}} = \bm{R}^{(k)}_{\bm{\phi}} - \bm{Q}_{\bm{\phi}}$.

Since $\eta<L^{-1}$, we have that:
\begin{align}
\bm{Q}_{\bm{\theta}} = \sum_{i=0}^{\infty}(-\eta^2\bm{B}\bm{B}^T)^k\nonumber\\
\bm{Q}_{\bm{\phi}} = \sum_{i=0}^{\infty}(-\eta^2\bm{B}^T\bm{B})^k\nonumber
\end{align}
and
\begin{align}
    \bm{E}^{(k)}_{\bm{\theta}} = \bm{R}^{(k)}_{\bm{\theta}} - \bm{Q}_{\bm{\theta}} = -\sum_{i=k}^{\infty}(-\eta^2\bm{B}\bm{B}^T)^k =-(\bm{I}+\eta^2\bm{B}\bm{B}^T)^{-1}\cdot(-\eta^2\bm{B}\bm{B}^T)^k \label{eq: Ethetak}\\
    \bm{E}^{(k)}_{\bm{\phi}} = \bm{R}^{(k)}_{\bm{\phi}} - \bm{Q}_{\bm{\phi}} = -\sum_{i=k}^{\infty}(-\eta^2\bm{B}^T\bm{B})^k=-(\bm{I}+\eta^2\bm{B}^T\bm{B})^{-1}\cdot(-\eta^2\bm{B}^T\bm{B})^k\label{eq: Ephik}
\end{align}
Also, from Lemma \ref{lemma: Q} and the definition of the error terms, it can be verified that
\begin{align}
    \bm{E}_{\bm{\theta}}^{(k)}\bm{B} &= \bm{B}\bm{E}_{\bm{\phi}}^{(k)}\label{eq: EB}\\
    \bm{E}_{\bm{\phi}}^{(k)}\bm{B}^T &= \bm{B}^T\bm{E}_{\bm{\theta}}^{(k)}\label{eq: EBT}
\end{align}
Then we can rewrite the update rule of Lv.$2k$ GP:
\begin{align}
    \begin{cases}
        \hat{\bm{\theta}}^{(2k)}_{t} = (\bm{Q}_{\bm{\theta}} + \bm{E}_{\bm{\theta}}^{(k)})\left[\bm{P}_{\bm{\theta}}\hat{\bm{\theta}}_{t} - \eta\bm{B}\bm{P}_{\bm{\phi}}\hat{\bm{\phi}}_{t}\right] +(-\eta^2 \bm{BB}^T)^{k} \hat{\bm{\theta}}_{t}\nonumber\\
        \hat{\bm{\phi}}^{(2k)}_{t} =(\bm{Q}_{\bm{\phi}} + \bm{E}_{\bm{\phi}}^{(k)})\left[ \eta\bm{B}^T\bm{P}_{\bm{\theta}}\hat{\bm{\theta}}_{t} + \bm{P}_{\bm{\phi}}\hat{\bm{\phi}}_{t}\right] + (-\eta^2\bm{B}^T\bm{B})^{k}\hat{\bm{\phi}}_{t}\nonumber
    \end{cases}
\end{align}
Let us consider the following sum:
\begin{align}
    \lVert &\hat{\bm{\theta}}^{(2k)}_{t}-(-\eta^2 \bm{BB}^T)^{k} \hat{\bm{\theta}}_{t}\rVert^2 + \lVert \hat{\bm{\phi}}^{(2k)}_{t} - (-\eta^2\bm{B}^T\bm{B})^{k}\hat{\bm{\phi}}_{t}\rVert^2\nonumber\\
    &= \left\lVert(\bm{Q}_{\bm{\theta}} + \bm{E}_{\bm{\theta}}^{(k)})\left[\bm{P}_{\bm{\theta}}\hat{\bm{\theta}}_{t} - \eta\bm{B}\bm{P}_{\bm{\phi}}\hat{\bm{\phi}}_{t}\right]\right\rVert^2 + \left\lVert(\bm{Q}_{\bm{\phi}} + \bm{E}_{\bm{\phi}}^{(k)})\left[ \eta\bm{B}^T\bm{P}_{\bm{\theta}}\hat{\bm{\theta}}_{t} + \bm{P}_{\bm{\phi}}\hat{\bm{\phi}}_{t}\right]\right\rVert^2\label{eq: intermediate result of Lv.2k GP}
\end{align}
The R.H.S. of Eq.(\ref{eq: intermediate result of Lv.2k GP}) can be written as:
\begin{align}
    &\left\lVert(\bm{Q}_{\bm{\theta}} + \bm{E}_{\bm{\theta}}^{(k)})\left[\bm{P}_{\bm{\theta}}\hat{\bm{\theta}}_{t} - \eta\bm{B}\bm{P}_{\bm{\phi}}\hat{\bm{\phi}}_{t}\right]\right\rVert^2 + \left\lVert(\bm{Q}_{\bm{\phi}} + \bm{E}_{\bm{\phi}}^{(k)})\left[ \eta\bm{B}^T\bm{P}_{\bm{\theta}}\hat{\bm{\theta}}_{t} + \bm{P}_{\bm{\phi}}\hat{\bm{\phi}}_{t}\right]\right\rVert^2\nonumber\\
    =&\hat{\bm{\theta}}_{t}^T\bm{P}_{\bm{\theta}}^T\bm{Q}_{\bm{\theta}}^2\bm{P}_{\bm{\theta}}\hat{\bm{\theta}}_{t} - 2\eta\hat{\bm{\theta}}_{t}^T\bm{P}_{\bm{\theta}}^T\bm{Q}_{\bm{\theta}}^2\bm{B}\bm{P}_{\bm{\phi}}\hat{\bm{\phi}}_{t} + \eta^2\hat{\bm{\phi}}_{t}^T\bm{P}_{\bm{\phi}}^T\bm{B}^T\bm{Q}_{\bm{\theta}}^2\bm{B}\bm{P}_{\bm{\phi}}\hat{\bm{\phi}}_{t}\nonumber\\
    +&2\hat{\bm{\theta}}_{t}^T\bm{P}_{\bm{\theta}}^T\bm{Q}_{\bm{\theta}}\bm{E}_{\bm{\theta}}^{(k)}\bm{P}_{\bm{\theta}}\hat{\bm{\theta}}_{t} - 4\eta\hat{\bm{\theta}}_{t}^T\bm{P}_{\bm{\theta}}^T\bm{Q}_{\bm{\theta}}\bm{E}_{\bm{\theta}}^{(k)}\bm{B}\bm{P}_{\bm{\phi}}\hat{\bm{\phi}}_{t} + 2\eta^2\hat{\bm{\phi}}_{t}^T\bm{P}_{\bm{\phi}}^T\bm{B}^T\bm{Q}_{\bm{\theta}}\bm{E}_{\bm{\theta}}^{(k)}\bm{B}\bm{P}_{\bm{\phi}}\hat{\bm{\phi}}_{t}\nonumber\\
    +&\hat{\bm{\theta}}_{t}^T\bm{P}_{\bm{\theta}}^T[\bm{E}_{\bm{\theta}}^{(k)}]^2\bm{P}_{\bm{\theta}}\hat{\bm{\theta}}_{t} - 2\eta\hat{\bm{\theta}}_{t}^T\bm{P}_{\bm{\theta}}^T[\bm{E}_{\bm{\theta}}^{(k)}]^2\bm{B}\bm{P}_{\bm{\phi}}\hat{\bm{\phi}}_{t} + \eta^2\hat{\bm{\phi}}_{t}^T\bm{P}_{\bm{\phi}}^T\bm{B}^T[\bm{E}_{\bm{\theta}}^{(k)}]^2\bm{B}\bm{P}_{\bm{\phi}}\hat{\bm{\phi}}_{t}\nonumber\\
    +&\hat{\bm{\phi}}_{t}^T\bm{P}_{\bm{\phi}}^T\bm{Q}_{\bm{\phi}}^2\bm{P}_{\bm{\phi}}\hat{\bm{\phi}}_{t} + 2\eta\hat{\bm{\theta}}_{t}^T\bm{P}_{\bm{\theta}}^T\bm{B}\bm{Q}_{\bm{\phi}}^2\bm{P}_{\bm{\phi}}\hat{\bm{\phi}}_{t} + \eta^2\hat{\bm{\theta}}_{t}^T\bm{P}_{\bm{\theta}}^T\bm{B}\bm{Q}_{\bm{\phi}}^2\bm{B}^T\bm{P}_{\bm{\theta}}\hat{\bm{\theta}}_{t}\nonumber\\
    +&2\hat{\bm{\phi}}_{t}^T\bm{P}_{\bm{\phi}}^T\bm{Q}_{\bm{\phi}}\bm{E}_{\bm{\phi}}^{(k)}\bm{P}_{\bm{\phi}}\hat{\bm{\phi}}_{t} + 4\eta\hat{\bm{\theta}}_{t}^T\bm{P}_{\bm{\theta}}^T\bm{B}\bm{Q}_{\bm{\phi}}\bm{E}_{\bm{\phi}}^{(k)}\bm{P}_{\bm{\phi}}\hat{\bm{\phi}}_{t} + 2\eta^2\hat{\bm{\theta}}_{t}^T\bm{P}_{\bm{\theta}}^T\bm{B}\bm{Q}_{\bm{\phi}}\bm{E}_{\bm{\phi}}^{(k)}\bm{B}^T\bm{P}_{\bm{\theta}}\hat{\bm{\theta}}_{t}\nonumber\\
    +&\hat{\bm{\phi}}_{t}^T\bm{P}_{\bm{\phi}}^T[\bm{E}_{\bm{\phi}}^{(k)}]^2\bm{P}_{\bm{\phi}}\hat{\bm{\phi}}_{t} + 2\eta\hat{\bm{\theta}}_{t}^T\bm{P}_{\bm{\theta}}^T\bm{B}[\bm{E}_{\bm{\phi}}^{(k)}]^2\bm{P}_{\bm{\phi}}\hat{\bm{\phi}}_{t} + \eta^2\hat{\bm{\theta}}_{t}^T\bm{P}_{\bm{\theta}}^T\bm{B}[\bm{E}_{\bm{\phi}}^{(k)}]^2\bm{B}^T\bm{P}_{\bm{\theta}}\hat{\bm{\theta}}_{t}\label{eq: 18 terms}
\end{align}
Now, before adding all terms in Eq.(\ref{eq: 18 terms}), note that all of the cross terms in Eq.(\ref{eq: 18 terms}) cancel out.

For instance, using Lemma \ref{lemma: Q} and Eq.(\ref{eq: EB}), Eq.(\ref{eq: EBT}) we can show that
\begin{align}
    &4\eta\hat{\bm{\theta}}_{t}^T\bm{P}_{\bm{\theta}}^T\bm{B}\bm{Q}_{\bm{\phi}}\bm{E}_{\bm{\phi}}^{(k)}\bm{P}_{\bm{\phi}}\hat{\bm{\phi}}_{t} - 4\eta\hat{\bm{\theta}}_{t}^T\bm{P}_{\bm{\theta}}^T\bm{Q}_{\bm{\theta}}\bm{E}_{\bm{\theta}}^{(k)}\bm{B}\bm{P}_{\bm{\phi}}\hat{\bm{\phi}}_{t}\nonumber\\ 
    &= 4\eta\hat{\bm{\theta}}_{t}^T\bm{P}_{\bm{\theta}}^T\bm{Q}_{\bm{\theta}}\bm{B}\bm{E}_{\bm{\phi}}^{(k)}\bm{P}_{\bm{\phi}}\hat{\bm{\phi}}_{t} - 4\eta\hat{\bm{\theta}}_{t}^T\bm{P}_{\bm{\theta}}^T\bm{Q}_{\bm{\theta}}\bm{E}_{\bm{\theta}}^{(k)}\bm{B}\bm{P}_{\bm{\phi}}\hat{\bm{\phi}}_{t}\nonumber\\
    &= 4\eta\hat{\bm{\theta}}_{t}^T\bm{P}_{\bm{\theta}}^T\bm{Q}_{\bm{\theta}}\bm{E}_{\bm{\theta}}^{(k)}\bm{B}\bm{P}_{\bm{\phi}}\hat{\bm{\phi}}_{t} - 4\eta\hat{\bm{\theta}}_{t}^T\bm{P}_{\bm{\theta}}^T\bm{Q}_{\bm{\theta}}\bm{E}_{\bm{\theta}}^{(k)}\bm{B}\bm{P}_{\bm{\phi}}\hat{\bm{\phi}}_{t}\nonumber\\
    &= 0\nonumber
\end{align}
By using similar arguments it can be shown that terms in Eq.(\ref{eq: 18 terms}) leads to:
\begin{align}
    &\left\lVert(\bm{Q}_{\bm{\theta}} + \bm{E}_{\bm{\theta}}^{(k)})\left[\bm{P}_{\bm{\theta}}\hat{\bm{\theta}}_{t} - \eta\bm{B}\bm{P}_{\bm{\phi}}\hat{\bm{\phi}}_{t}\right]\right\rVert^2 + \left\lVert(\bm{Q}_{\bm{\phi}} + \bm{E}_{\bm{\phi}}^{(k)})\left[ \eta\bm{B}^T\bm{P}_{\bm{\theta}}\hat{\bm{\theta}}_{t} + \bm{P}_{\bm{\phi}}\hat{\bm{\phi}}_{t}\right]\right\rVert^2\nonumber\\
    =&\hat{\bm{\theta}}_{t}^T\bm{P}_{\bm{\theta}}^T\bm{Q}_{\bm{\theta}}^2\bm{P}_{\bm{\theta}}\hat{\bm{\theta}}_{t} + \eta^2\hat{\bm{\phi}}_{t}^T\bm{P}_{\bm{\phi}}^T\bm{B}^T\bm{Q}_{\bm{\theta}}^2\bm{B}\bm{P}_{\bm{\phi}}\hat{\bm{\phi}}_{t}\nonumber\\
    +&2\hat{\bm{\theta}}_{t}^T\bm{P}_{\bm{\theta}}^T\bm{Q}_{\bm{\theta}}\bm{E}_{\bm{\theta}}^{(k)}\bm{P}_{\bm{\theta}}\hat{\bm{\theta}}_{t}+ 2\eta^2\hat{\bm{\phi}}_{t}^T\bm{P}_{\bm{\phi}}^T\bm{B}^T\bm{Q}_{\bm{\theta}}\bm{E}_{\bm{\theta}}^{(k)}\bm{B}\bm{P}_{\bm{\phi}}\hat{\bm{\phi}}_{t}\nonumber\\
    +&\hat{\bm{\theta}}_{t}^T\bm{P}_{\bm{\theta}}^T[\bm{E}_{\bm{\theta}}^{(k)}]^2\bm{P}_{\bm{\theta}}\hat{\bm{\theta}}_{t} + \eta^2\hat{\bm{\phi}}_{t}^T\bm{P}_{\bm{\phi}}^T\bm{B}^T[\bm{E}_{\bm{\theta}}^{(k)}]^2\bm{B}\bm{P}_{\bm{\phi}}\hat{\bm{\phi}}_{t}\nonumber\\
    +&\hat{\bm{\phi}}_{t}^T\bm{P}_{\bm{\phi}}^T\bm{Q}_{\bm{\phi}}^2\bm{P}_{\bm{\phi}}\hat{\bm{\phi}}_{t} + \eta^2\hat{\bm{\theta}}_{t}^T\bm{P}_{\bm{\theta}}^T\bm{B}\bm{Q}_{\bm{\phi}}^2\bm{B}^T\bm{P}_{\bm{\theta}}\hat{\bm{\theta}}_{t}\nonumber\\
    +&2\hat{\bm{\phi}}_{t}^T\bm{P}_{\bm{\phi}}^T\bm{Q}_{\bm{\phi}}\bm{E}_{\bm{\phi}}^{(k)}\bm{P}_{\bm{\phi}}\hat{\bm{\phi}}_{t}  + 2\eta^2\hat{\bm{\theta}}_{t}^T\bm{P}_{\bm{\theta}}^T\bm{B}\bm{Q}_{\bm{\phi}}\bm{E}_{\bm{\phi}}^{(k)}\bm{B}^T\bm{P}_{\bm{\theta}}\hat{\bm{\theta}}_{t}\nonumber\\
    +&\hat{\bm{\phi}}_{t}^T\bm{P}_{\bm{\phi}}^T[\bm{E}_{\bm{\phi}}^{(k)}]^2\bm{P}_{\bm{\phi}}\hat{\bm{\phi}}_{t} + \eta^2\hat{\bm{\theta}}_{t}^T\bm{P}_{\bm{\theta}}^T\bm{B}[\bm{E}_{\bm{\phi}}^{(k)}]^2\bm{B}^T\bm{P}_{\bm{\theta}}\hat{\bm{\theta}}_{t}\label{eq: 12 terms}
\end{align}
Similar to Eq.(\ref{eq: expression for xt}) we have the following simplification:
\begin{align}
    \hat{\bm{\theta}}_{t}^T\bm{P}_{\bm{\theta}}^T\bm{Q}_{\bm{\theta}}^2\bm{P}_{\bm{\theta}}\hat{\bm{\theta}}_{t} + \eta^2\hat{\bm{\theta}}_{t}^T\bm{P}_{\bm{\theta}}^T\bm{B}\bm{Q}_{\bm{\phi}}^2\bm{B}^T\bm{P}_{\bm{\theta}}\hat{\bm{\theta}}_{t} &=     \hat{\bm{\theta}}_{t}^T\bm{P}_{\bm{\theta}}^T\bm{Q}_{\bm{\theta}}\bm{P}_{\bm{\theta}}\hat{\bm{\theta}}_{t}\nonumber\\
    \hat{\bm{\phi}}_{t}^T\bm{P}_{\bm{\phi}}^T\bm{Q}_{\bm{\phi}}^2\bm{P}_{\bm{\phi}}\hat{\bm{\phi}}_{t} + \eta^2\hat{\bm{\phi}}_{t}^T\bm{P}_{\bm{\phi}}^T\bm{B}^T\bm{Q}_{\bm{\theta}}^2\bm{B}\bm{P}_{\bm{\phi}}\hat{\bm{\phi}}_{t} &=     \hat{\bm{\phi}}_{t}^T\bm{P}_{\bm{\phi}}^T\bm{Q}_{\bm{\phi}}\bm{P}_{\bm{\phi}}\hat{\bm{\phi}}_{t}\nonumber\\
    2\hat{\bm{\theta}}_{t}^T\bm{P}_{\bm{\theta}}^T\bm{Q}_{\bm{\theta}}\bm{E}_{\bm{\theta}}^{(k)}\bm{P}_{\bm{\theta}}\hat{\bm{\theta}}_{t} + 2\eta^2\hat{\bm{\theta}}_{t}^T\bm{P}_{\bm{\theta}}^T\bm{B}\bm{Q}_{\bm{\phi}}\bm{E}_{\bm{\phi}}^{(k)}\bm{B}^T\bm{P}_{\bm{\theta}}\hat{\bm{\theta}}_{t} &=     2\hat{\bm{\theta}}_{t}^T\bm{P}_{\bm{\theta}}^T\bm{E}_{\bm{\theta}}^{(k)}\bm{P}_{\bm{\theta}}\hat{\bm{\theta}}_{t}\nonumber\\
    2\hat{\bm{\phi}}_{t}^T\bm{P}_{\bm{\phi}}^T\bm{Q}_{\bm{\phi}}\bm{E}_{\bm{\phi}}^{(k)}\bm{P}_{\bm{\phi}}\hat{\bm{\phi}}_{t} + 2\eta^2\hat{\bm{\phi}}_{t}^T\bm{P}_{\bm{\phi}}^T\bm{B}^T\bm{Q}_{\bm{\theta}}\bm{E}_{\bm{\theta}}^{(k)}\bm{B}\bm{P}_{\bm{\phi}}\hat{\bm{\phi}}_{t} &=     2\hat{\bm{\phi}}_{t}^T\bm{P}_{\bm{\phi}}^T\bm{E}_{\bm{\phi}}^{(k)}\bm{P}_{\bm{\phi}}\hat{\bm{\phi}}_{t}\nonumber
\end{align}
Now we can further simplify Eq.(\ref{eq: 18 terms}) as:
\begin{align}
    &\left\lVert(\bm{Q}_{\bm{\theta}} + \bm{E}_{\bm{\theta}}^{(k)})\left[\bm{P}_{\bm{\theta}}\hat{\bm{\theta}}_{t} - \eta\bm{B}\bm{P}_{\bm{\phi}}\hat{\bm{\phi}}_{t}\right]\right\rVert^2 + \left\lVert(\bm{Q}_{\bm{\phi}} + \bm{E}_{\bm{\phi}}^{(k)})\left[ \eta\bm{B}^T\bm{P}_{\bm{\theta}}\hat{\bm{\theta}}_{t} + \bm{P}_{\bm{\phi}}\hat{\bm{\phi}}_{t}\right]\right\rVert^2\nonumber\\
    =& (\bm{P}_{\bm{\theta}}\hat{\bm{\theta}}_{t})^T \left[ \bm{Q}_{\bm{\theta}} + 2\bm{E}_{\bm{\theta}}^{(k)} + [\bm{E}_{\bm{\theta}}^{(k)}]^2 + \eta^2\bm{B}[\bm{E}_{\bm{\phi}}^{(k)}]^2\bm{B}^T\right](\bm{P}_{\bm{\theta}}\hat{\bm{\theta}}_{t})\nonumber\\
    +& (\bm{P}_{\bm{\phi}}\hat{\bm{\phi}}_{t})^T \left[ \bm{Q}_{\bm{\phi}} + 2\bm{E}_{\bm{\phi}}^{(k)} + [\bm{E}_{\bm{\phi}}^{(k)}]^2 + \eta^2\bm{B}[\bm{E}_{\bm{\theta}}^{(k)}]^2\bm{B}^T\right](\bm{P}_{\bm{\phi}}\hat{\bm{\phi}}_{t})\nonumber
\end{align}
Using Eq.(\ref{eq: Ethetak}) and Eq.(\ref{eq: Ephik}) and definition of $\bm{Q}_{\bm{\theta}}$ and $\bm{Q}_{\bm{\phi}}$ we have:
\begin{align}
    &(\bm{P}_{\bm{\theta}}\hat{\bm{\theta}}_{t})^T \left[ \bm{Q}_{\bm{\theta}} + 2\bm{E}_{\bm{\theta}}^{(k)} + [\bm{E}_{\bm{\theta}}^{(k)}]^2 + \eta^2\bm{B}[\bm{E}_{\bm{\phi}}^{(k)}]^2\bm{B}^T\right](\bm{P}_{\bm{\theta}}\hat{\bm{\theta}}_{t})\nonumber\\
    =&(\bm{P}_{\bm{\theta}}\hat{\bm{\theta}}_{t})^T(\bm{I} + \eta^2\bm{B}\bm{B}^T)^{-1}(\bm{P}_{\bm{\theta}}\hat{\bm{\theta}}_{t}) -2(\bm{P}_{\bm{\theta}}\hat{\bm{\theta}}_{t})^T(\bm{I} + \eta^2\bm{B}\bm{B}^T)^{-1}(-\eta^2\bm{B}\bm{B}^T)^{k}(\bm{P}_{\bm{\theta}}\hat{\bm{\theta}}_{t}) \nonumber\\
    +&(\bm{P}_{\bm{\theta}}\hat{\bm{\theta}}_{t})^T(\bm{I} + \eta^2\bm{B}\bm{B}^T)(\bm{I} + \eta^2\bm{B}\bm{B}^T)^{-2}(-\eta^2\bm{B}\bm{B}^T)^{2k}(\bm{P}_{\bm{\theta}}\hat{\bm{\theta}}_{t})\nonumber\\
    =&(\bm{P}_{\bm{\theta}}\hat{\bm{\theta}}_{t})^T(\bm{I} + \eta^2\bm{B}\bm{B}^T)^{-1}(\bm{I} -2 (-\eta^2\bm{B}\bm{B}^T)^{(k)} + (-\eta^2\bm{B}\bm{B}^T)^{2k})(\bm{P}_{\bm{\theta}}\hat{\bm{\theta}}_{t})\nonumber\\
    =&((\bm{I}-(-\eta^2\bm{B}\bm{B}^T)^{k})\bm{P}_{\bm{\theta}}\hat{\bm{\theta}}_{t})^T(\bm{I} + \eta^2\bm{B}\bm{B}^T)^{-1}((\bm{I}-(-\eta^2\bm{B}\bm{B}^T)^{k})\bm{P}_{\bm{\theta}}\hat{\bm{\theta}}_{t})\nonumber
\end{align}
Similarly, we have that 
\begin{align}
    &(\bm{P}_{\bm{\phi}}\hat{\bm{\phi}}_{t})^T \left[ \bm{Q}_{\bm{\phi}} + 2\bm{E}_{\bm{\phi}}^{(k)} + [\bm{E}_{\bm{\phi}}^{(k)}]^2 + \eta^2\bm{B}^T[\bm{E}_{\bm{\theta}}^{(k)}]^2\bm{B}\right](\bm{P}_{\bm{\phi}}\hat{\bm{\phi}}_{t})\nonumber\\
    =&((\bm{I}-(-\eta^2\bm{B}^T\bm{B})^{k})\bm{P}_{\bm{\phi}}\hat{\bm{\phi}}_{t})^T(\bm{I} + \eta^2\bm{B}^T\bm{B})^{-1}((\bm{I}-(-\eta^2\bm{B}^T\bm{B})^{k})\bm{P}_{\bm{\phi}}\hat{\bm{\phi}}_{t})\nonumber
\end{align}
Thus we simplify the R.H.S. of Eq.(\ref{eq: intermediate result of Lv.2k GP}) as
\begin{align}
    &\left\lVert(\bm{Q}_{\bm{\theta}} + \bm{E}_{\bm{\theta}}^{(k)})\left[\bm{P}_{\bm{\theta}}\hat{\bm{\theta}}_{t} - \eta\bm{B}\bm{P}_{\bm{\phi}}\hat{\bm{\phi}}_{t}\right]\right\rVert^2 + \left\lVert(\bm{Q}_{\bm{\phi}} + \bm{E}_{\bm{\phi}}^{(k)})\left[ \eta\bm{B}^T\bm{P}_{\bm{\theta}}\hat{\bm{\theta}}_{t} + \bm{P}_{\bm{\phi}}\hat{\bm{\phi}}_{t}\right]\right\rVert^2\nonumber\\
    =&((\bm{I}-(-\eta^2\bm{B}\bm{B}^T)^{k})\bm{P}_{\bm{\theta}}\hat{\bm{\theta}}_{t})^T(\bm{I} + \eta^2\bm{B}\bm{B}^T)^{-1}((\bm{I}-(-\eta^2\bm{B}\bm{B}^T)^{k})\bm{P}_{\bm{\theta}}\hat{\bm{\theta}}_{t})\nonumber\\
    +&((\bm{I}-(-\eta^2\bm{B}^T\bm{B})^{k})\bm{P}_{\bm{\phi}}\hat{\bm{\phi}}_{t})^T(\bm{I} + \eta^2\bm{B}^T\bm{B})^{-1}((\bm{I}-(-\eta^2\bm{B}^T\bm{B})^{k})\bm{P}_{\bm{\phi}}\hat{\bm{\phi}}_{t})\label{eq:R.H.S.}
\end{align}
Let us consider the L.H.S. of Eq.(\ref{eq: intermediate result of Lv.2k GP})
\begin{align}
     \lVert &\hat{\bm{\theta}}^{(2k)}_{t}-(-\eta^2 \bm{BB}^T)^{k} \hat{\bm{\theta}}_{t}\rVert^2 + \lVert \hat{\bm{\phi}}^{(2k)}_{t} - (-\eta^2\bm{B}^T\bm{B})^{k}\hat{\bm{\phi}}_{t}\rVert^2\nonumber\\
     =&\lVert \hat{\bm{\theta}}^{(2k)}_{t}\rVert^2 - 2\langle\hat{\bm{\theta}}^{(2k)}_{t},(-\eta^2 \bm{BB}^T)^{k} \hat{\bm{\theta}}_{t} \rangle +\lVert(-\eta^2 \bm{BB}^T)^{k} \hat{\bm{\theta}}_{t}\rVert^2 \nonumber\\
     +&\lVert \hat{\bm{\phi}}^{(2k)}_{t}\rVert^2 - 2\langle\hat{\bm{\phi}}^{(2k)}_{t},(-\eta^2 \bm{B}^T\bm{B})^{k} \hat{\bm{\phi}}_{t} \rangle +\lVert(-\eta^2 \bm{B}^T\bm{B})^{k} \hat{\bm{\phi}}_{t}\rVert^2 \label{eq: L.H.S.}
\end{align}
Substituting Eq.(\ref{eq: L.H.S.}) and Eq.(\ref{eq:R.H.S.}) into L.H.S. and R.H.S. of Eq.(\ref{eq: intermediate result of Lv.2k GP}) respectively we get: 
\begin{align}
    &\lVert \hat{\bm{\theta}}^{(2k)}_{t}\rVert^2 - 2\langle\hat{\bm{\theta}}^{(2k)}_{t},(-\eta^2 \bm{BB}^T)^{k} \hat{\bm{\theta}}_{t} \rangle +\lVert(-\eta^2 \bm{BB}^T)^{k} \hat{\bm{\theta}}_{t}\rVert^2 \nonumber\\
    +&\lVert \hat{\bm{\phi}}^{(2k)}_{t}\rVert^2 - 2\langle\hat{\bm{\phi}}^{(2k)}_{t},(-\eta^2 \bm{B}^T\bm{B})^{k} \hat{\bm{\phi}}_{t} \rangle +\lVert(-\eta^2 \bm{B}^T\bm{B})^{k} \hat{\bm{\phi}}_{t}\rVert^2\nonumber\\
    =&((\bm{I}-(-\eta^2\bm{B}\bm{B}^T)^{k})\bm{P}_{\bm{\theta}}\hat{\bm{\theta}}_{t})^T(\bm{I} + \eta^2\bm{B}\bm{B}^T)^{-1}((\bm{I}-(-\eta^2\bm{B}\bm{B}^T)^{k})\bm{P}_{\bm{\theta}}\hat{\bm{\theta}}_{t})\nonumber\\
    +&((\bm{I}-(-\eta^2\bm{B}^T\bm{B})^{k})\bm{P}_{\bm{\phi}}\hat{\bm{\phi}}_{t})^T(\bm{I} + \eta^2\bm{B}^T\bm{B})^{-1}((\bm{I}-(-\eta^2\bm{B}^T\bm{B})^{k})\bm{P}_{\bm{\phi}}\hat{\bm{\phi}}_{t})\nonumber
\end{align}
Now we have the following equation:
\begin{align}
    &\lVert \hat{\bm{\theta}}^{(2k)}_{t}\rVert^2 + \lVert \hat{\bm{\phi}}^{(2k)}_{t}\rVert^2\nonumber\\ 
    =& ((\bm{I}-(-\eta^2\bm{B}\bm{B}^T)^{k})\bm{P}_{\bm{\theta}}\hat{\bm{\theta}}_{t})^T(\bm{I} + \eta^2\bm{B}\bm{B}^T)^{-1}((\bm{I}-(-\eta^2\bm{B}\bm{B}^T)^{k})\bm{P}_{\bm{\theta}}\hat{\bm{\theta}}_{t})\nonumber\\
    +& 2\langle\hat{\bm{\theta}}^{(2k)}_{t},(-\eta^2 \bm{BB}^T)^{k} \hat{\bm{\theta}}_{t} \rangle -\lVert(-\eta^2 \bm{BB}^T)^{k} \hat{\bm{\theta}}_{t}\rVert^2\nonumber\\
    +&((\bm{I}-(-\eta^2\bm{B}^T\bm{B})^{k})\bm{P}_{\bm{\phi}}\hat{\bm{\phi}}_{t})^T(\bm{I} + \eta^2\bm{B}^T\bm{B})^{-1}((\bm{I}-(-\eta^2\bm{B}^T\bm{B})^{k})\bm{P}_{\bm{\phi}}\hat{\bm{\phi}}_{t})\nonumber\\
    +& 2\langle\hat{\bm{\phi}}^{(2k)}_{t},(-\eta^2 \bm{B}^T\bm{B})^{k} \hat{\bm{\phi}}_{t} \rangle -\lVert(-\eta^2 \bm{B}^T\bm{B})^{k} \hat{\bm{\phi}}_{t}\rVert^2\nonumber
\end{align}
Note that 
\begin{align}
    2\langle\hat{\bm{\theta}}^{(2k)}_{t},(-\eta^2 \bm{BB}^T)^{k} \hat{\bm{\theta}}_{t} \rangle &= 2\langle(-\eta^2\bm{BB}^T)^{\frac{k}{2}}\hat{\bm{\theta}}^{(2k)}_{t},(\eta^2 \bm{BB}^T)^{\frac{k}{2}} \hat{\bm{\theta}}_{t} \rangle\\
    &\leq \eta^{2k}(\hat{\bm{\theta}}^{(2k)}_{t})^T(\bm{BB}^T)^{k}\hat{\bm{\theta}}^{(2k)}_{t} + \eta^{2k}\hat{\bm{\theta}}_{t}^T(\bm{BB}^T)^{k}\hat{\bm{\theta}}_{t}
\end{align}
Similarly
\begin{align}
    2\langle\hat{\bm{\phi}}^{(2k)}_{t},(-\eta^2 \bm{BB}^T)^{k} \hat{\bm{\phi}}_{t} \rangle &= 2\langle(-\eta^2\bm{B}^T\bm{B})^{\frac{k}{2}}\hat{\bm{\phi}}^{(2k)}_{t},(\eta^2 \bm{B}^T\bm{B})^{\frac{k}{2}} \hat{\bm{\phi}}_{t} \rangle\\
    &\leq \eta^{2k}(\hat{\bm{\phi}}^{(2k)}_{t})^T(\bm{B}^T\bm{B})^{k}\hat{\bm{\phi}}^{(2k)}_{t} + \eta^{2k}\hat{\bm{\phi}}_{t}^T(\bm{B}^T\bm{B})^{k}\hat{\bm{\phi}}_{t}
\end{align}
Summing everything together we have:
\begin{align}
    &(\hat{\bm{\theta}}^{(2k)}_{t})^T(\bm{I} - (\eta^2\bm{B}\bm{B}^T)^{k})(\hat{\bm{\theta}}^{(2k)}_{t}) +     (\hat{\bm{\phi}}^{(2k)}_{t})^T(\bm{I} - (\eta^2\bm{B}^T\bm{B})^{k})(\hat{\bm{\phi}}^{(2k)}_{t})\nonumber\\
    \leq& ((\bm{I}-(-\eta^2\bm{B}\bm{B}^T)^{k})\bm{P}_{\bm{\theta}}\hat{\bm{\theta}}_{t})^T(\bm{I} + \eta^2\bm{B}\bm{B}^T)^{-1}((\bm{I}-(-\eta^2\bm{B}\bm{B}^T)^{k})\bm{P}_{\bm{\theta}}\hat{\bm{\theta}}_{t})\nonumber\\
    +& (\hat{\bm{\theta}}_{t})^T(\eta^2\bm{B}\bm{B}^T)^{k}(\hat{\bm{\theta}}_{t}) -\lVert(-\eta^2 \bm{BB}^T)^{k} \hat{\bm{\theta}}_{t}\rVert^2\nonumber\\
    +&((\bm{I}-(-\eta^2\bm{B}^T\bm{B})^{k})\bm{P}_{\bm{\phi}}\hat{\bm{\phi}}_{t})^T(\bm{I} + \eta^2\bm{B}^T\bm{B})^{-1}((\bm{I}-(-\eta^2\bm{B}^T\bm{B})^{k})\bm{P}_{\bm{\phi}}\hat{\bm{\phi}}_{t})\nonumber\\
    +& (\hat{\bm{\phi}}_{t})^T (\eta^2\bm{B}^T\bm{B})^{k}(\hat{\bm{\phi}}_{t}) -\lVert(-\eta^2 \bm{B}^T\bm{B})^{k} \hat{\bm{\phi}}_{t}\rVert^2\nonumber
\end{align}
Note that, we assume that the trajectory $\{\hat{\bm{\theta}}_{t},\hat{\bm{\phi}}_{t}\}_{t\geq 0}$ is not in the kernel of $\bm{B}\bm{B}^T$ and $\bm{B}^T\bm{B}$, thus $\bm{B}\bm{B}^T\hat{\bm{\theta}}_{t}\neq 0$ and $\bm{B}^T\bm{B}\hat{\bm{\phi}}_{t}\neq 0$.
Now using the expression in (\ref{eq: sum of norm result}) and the fact that $\bm{P}_{\bm{\theta}} = \bm{P}_{\bm{\theta}}^T$, $\bm{P}_{\bm{\phi}} = \bm{P}_{\bm{\phi}}^T$ and $\bm{B}\bm{B}^T$ and $\bm{B}^T\bm{B}$ have the same set of non-zero eigenvalues, if we denote the minimum non-zero eigenvalues by $\lambda_{\min}(\bm{B}\bm{B}^T)$ and $\lambda_{\min}(\bm{B}^T\bm{B})$, we can write
\begin{align}
    &(1-(\eta^2\lambda_{max}(\bm{B}\bm{B}^T))^k)\lVert \hat{\bm{\theta}_{t}^{(2k)}}\rVert^2 + (1-(\eta^2\lambda_{max}(\bm{B}^T\bm{B}))^k)\lVert \hat{\bm{\phi}_{t}^{(2k)}}\rVert^2\nonumber\\
    \leq& \frac{(1-(-\eta^2\lambda(\bm{B}\bm{B}^T))^k)^2\rho^2(1-\eta\bm{A})\lVert\hat{\bm{\theta}}_{t}\rVert^2}{1+\eta^2\lambda_{min}(\bm{B}\bm{B}^T)}+(\eta^2\lambda(\bm{B}\bm{B}^T))^{k}\lVert\hat{\bm{\theta}}_{t}\rVert^2-(\eta^2\lambda(\bm{B}\bm{B}^T))^{2k}\lVert\hat{\bm{\theta}}_{t}\rVert^2\nonumber\\
    +& \frac{(1-(-\eta^2\lambda(\bm{B}^T\bm{B}))^k)^2\rho^2(1+\eta\bm{C})\lVert\hat{\bm{\phi}}_{t}\rVert^2}{1+\eta^2\lambda_{min}(\bm{B}^T\bm{B})}+(\eta^2\lambda(\bm{B}^T\bm{B}))^{k}\lVert\hat{\bm{\phi}}_{t}\rVert^2-(\eta^2\lambda(\bm{B}^T\bm{B}))^{2k}\lVert\hat{\bm{\phi}}_{t}\rVert^2\nonumber\\
    \leq& \frac{(1-(-\eta^2\lambda(\bm{B}\bm{B}^T))^k)^2\rho^2(1-\eta\bm{A})\lVert\hat{\bm{\theta}}_{t}\rVert^2}{1+\eta^2\lambda_{min}(\bm{B}\bm{B}^T)}+(\eta^2\lambda(\bm{B}\bm{B}^T))^{k}(1-(\eta^2\lambda(\bm{B}\bm{B}^T))^{k})\lVert\hat{\bm{\theta}}_{t}\rVert^2\nonumber\\
    +& \frac{(1-(-\eta^2\lambda(\bm{B}^T\bm{B}))^k)^2\rho^2(1+\eta\bm{C})\lVert\hat{\bm{\phi}}_{t}\rVert^2}{1+\eta^2\lambda_{min}(\bm{B}^T\bm{B})}+(\eta^2\lambda(\bm{B}^T\bm{B}))^{k}\lVert\hat{\bm{\phi}}_{t}\rVert^2-(\eta^2\lambda(\bm{B}^T\bm{B}))^{2k}\lVert\hat{\bm{\phi}}_{t}\rVert^2\nonumber\\
    \leq& \frac{(1-(-\eta^2\lambda(\bm{B}\bm{B}^T))^k)^2\rho^2(1-\eta\bm{A})\lVert\hat{\bm{\theta}}_{t}\rVert^2}{(1+\eta^2\lambda_{min}(\bm{B}\bm{B}^T))(1-(\eta^2\lambda_{max}(\bm{B}\bm{B}^T))^k)}+\frac{(\eta^2\lambda(\bm{B}\bm{B}^T))^{k}(1-(\eta^2\lambda(\bm{B}\bm{B}^T))^{k})\lVert\hat{\bm{\theta}}_{t}\rVert^2}{(1-(\eta^2\lambda_{max}(\bm{B}\bm{B}^T))^k)}\nonumber\\
    +& \frac{(1-(-\eta^2\lambda(\bm{B}^T\bm{B}))^k)^2\rho^2(1+\eta\bm{C})\lVert\hat{\bm{\phi}}_{t}\rVert^2}{(1+\eta^2\lambda_{min}(\bm{B}^T\bm{B}))(1-(\eta^2\lambda_{max}(\bm{B}^T\bm{B}))^k)}+\frac{(\eta^2\lambda(\bm{B}^T\bm{B}))^{k}(1-(\eta^2\lambda(\bm{B}^T\bm{B}))^{k})\lVert\hat{\bm{\phi}}_{t}\rVert^2}{(1-(\eta^2\lambda_{max}(\bm{B}^T\bm{B}))^k)}\nonumber\\
\end{align}
Let us define the distance as:
\begin{align}
\lVert r_{t}^{(k)}\rVert^2 &= \lVert\hat{\bm{\theta}}_{t}^{(k)} \rVert^2 + \lVert\hat{\bm{\phi}}_{t}^{(k)} \rVert^2\\
\lVert r_{t}\rVert^2 &= \lVert\hat{\bm{\theta}}_{t} \rVert^2 + \lVert\hat{\bm{\phi}}_{t} \rVert^2
\end{align}
Then we have
\begin{align}
    \lVert r_{t}^{(2k)}&\rVert\leq \frac{(1-(-\eta^2\lambda(\bm{B}\bm{B}^T))^k)^2(\rho^2(1-\eta\bm{A})\lVert\hat{\bm{\theta}}_{t}\rVert^2+\rho^2(1+\eta\bm{C})\lVert\hat{\bm{\phi}}_{t}\rVert^2)}{(1+\eta^2\lambda_{min}(\bm{B}\bm{B}^T))(1-(\eta^2\lambda_{max}(\bm{B}\bm{B}^T))^k)}\nonumber\\
    &+\frac{(\eta^2\lambda(\bm{B}\bm{B}^T))^{k}(1-(\eta^2\lambda(\bm{B}\bm{B}^T))^{k})}{(1-(\eta^2\lambda_{max}(\bm{B}\bm{B}^T))^k)}\lVert r_{t}\rVert^2\nonumber\\
    &= a(\frac{\rho^2(1-\eta\bm{A})\lVert\hat{\bm{\theta}}_{t}\rVert^2+\rho^2(1+\eta\bm{C})\lVert\hat{\bm{\phi}}_{t}\rVert^2}{1+\eta^2\lambda_{min}(\bm{B}\bm{B}^T)}) + b\lVert r_{t} \rVert^2
\end{align}
where 
\begin{align}
a =
\begin{cases}
    \frac{(1+(\eta^2\lambda_{\max}(\bm{B}\bm{B}^T))^{k})^2}{1-(\eta^2\lambda_{\max}(\bm{B}\bm{B}^T))^{k}} \text{ , odd $k$  }\\
    \frac{(1-(\eta^2\lambda_{\min}(\bm{B}\bm{B}^T))^{k})^2}{1-(\eta^2\lambda_{\max}(\bm{B}\bm{B}^T))^{k}} \text{  , even $k$}\nonumber
\end{cases}
\end{align}
and 
\begin{align}
    b = \frac{(\eta^2\lambda_{\max}(\bm{B}\bm{B}^T))^k(1-(\eta^2\lambda_{\min}(\bm{B}\bm{B}^T))^{k})}{1-(\eta^2\lambda_{\max}(\bm{B}\bm{B}^T))^{k}}
\end{align}
\end{proof}

\subsubsection{Proof of Lemma \ref{lemma: Q}}
Let $\bm{B} = \bm{U}\bm{\Sigma}\bm{V}^T$ be the singular value decomposition of $\bm{B}$. Here $\bm{U}$ and $\bm{V}$ are orthonormal matrices and $\bm{\Sigma}$ is a rectangular diagonal matrix. Then we have:
\begin{align}
    \bm{Q}_{\bm{\theta}}\bm{B} &= (\bm{I} + \eta^2\bm{U}\bm{\Sigma}\bm{V}^T\bm{V}\bm{\Sigma}^T\bm{U}^T)^{-1}\bm{U}\bm{\Sigma}\bm{V}^T \nonumber\\
    &= (\bm{U}(\eta^2\bm{\Sigma}\bm{\Sigma}^T + \bm{I})\bm{U}^T)^{-1}\bm{U}\bm{\Sigma}\bm{V}^T \nonumber\\
    &= \bm{U}(\eta^2\bm{\Sigma}\bm{\Sigma}^T + \bm{I})^{-1}\bm{U}^T\bm{U}\bm{\Sigma}\bm{V}^T \nonumber\\
    &= \bm{U}(\eta^2\bm{\Sigma}\bm{\Sigma}^T + \bm{I})^{-1}\bm{\Sigma}\bm{V}^T \nonumber
\end{align}
Here we used the fact that $\bm{U}$ and $\bm{V}$ are orthonormal matrices. Now, we simplify the other side to get:
\begin{align}
    \bm{B}\bm{Q}_{\bm{\phi}} &= \bm{U}\bm{\Sigma}\bm{V}^T(\bm{I} + \eta^2\bm{V}\bm{\Sigma}^T\bm{U}^T\bm{U}\bm{\Sigma}\bm{V}^T)^{-1} \nonumber\\
    &= \bm{U}\bm{\Sigma}\bm{V}^T(\bm{V}(\eta^2\bm{\Sigma}^T\bm{\Sigma} + \bm{I})\bm{V}^T)^{-1} \nonumber\\
    &= \bm{U}\bm{\Sigma}\bm{V}^T\bm{V}(\eta^2\bm{\Sigma}^T\bm{\Sigma} + \bm{I})^{-1}\bm{V}^T \nonumber\\
    &= \bm{U}\bm{\Sigma}(\eta^2\bm{\Sigma}^T\bm{\Sigma} + \bm{I})^{-1}\bm{V}^T \nonumber
\end{align}
Now we consider the following equation:
\begin{equation}
    \eta^2\bm{\Sigma}\bm{\Sigma}^T\bm{\Sigma} + \bm{\Sigma} = \bm{\Sigma}(\eta^2\bm{\Sigma}^T\bm{\Sigma} + \bm{I}) = (\eta^2\bm{\Sigma}\bm{\Sigma}^T + \bm{I})\bm{\Sigma}
\end{equation}
which indicates that $\bm{\Sigma}(\eta^2\bm{\Sigma}^T\bm{\Sigma} + \bm{I}) = (\eta^2\bm{\Sigma}\bm{\Sigma}^T + \bm{I})\bm{\Sigma}$. Multiplying both sides of this equation by $(\eta^2\bm{\Sigma}\bm{\Sigma}^T + \bm{I})^{-1}$ and $(\eta^2\bm{\Sigma}^T\bm{\Sigma} + \bm{I})^{-1}$ we have:
\begin{align}
    (\eta^2\bm{\Sigma}\bm{\Sigma}^T + \bm{I})^{-1}\bm{\Sigma}(\eta^2\bm{\Sigma}^T\bm{\Sigma} + \bm{I})(\eta^2\bm{\Sigma}^T\bm{\Sigma} + \bm{I})^{-1} &= (\eta^2\bm{\Sigma}\bm{\Sigma}^T + \bm{I})^{-1}(\eta^2\bm{\Sigma}\bm{\Sigma}^T + \bm{I})\bm{\Sigma}(\eta^2\bm{\Sigma}^T\bm{\Sigma} + \bm{I})^{-1}\nonumber\\
    (\eta^2\bm{\Sigma}\bm{\Sigma}^T + \bm{I})^{-1}\bm{\Sigma} &= \bm{\Sigma}(\eta^2\bm{\Sigma}^T\bm{\Sigma} + \bm{I})^{-1}\nonumber
\end{align}
Therefore, we have $\bm{Q}_{\bm{\theta}}\bm{B} = \bm{B}\bm{Q}_{\bm{\phi}}$. Using a similar argument, we can also prove the equality in Equation (\ref{eq:2}).
\subsection{Theorem \ref{thm: convergence in nonconvex gamae} without kernel assumption}\label{appendix: without kernel assumption}
\begin{theorem}
Consider the (\ref{eq: minimax problem}) problem under Assumption \ref{assumption: Lipschitz gradient assumption} and Lv.$k$ GP. Let $(\bm{\theta}^*,\bm{\phi}^*)$ be a stationary point. Suppose  $\eta<(L)^{-1}$. There exists a neighborhood $\mathcal{U}$ of $(\bm{\theta}^*,\bm{\phi}^*)$ such that if SPPM started at $(\bm{\theta}_{0},\bm{\phi}_{0})\in\mathcal{U}$, the iterates $\{\bm{\theta}_{t},\bm{\phi}_{t}\}_{t\geq0}$ generated by SPPM satisfy:
\begin{equation}
    \lVert\bm{\theta}_{t+1} - \bm{\theta}^*\rVert^2 + \lVert\bm{\phi}_{t+1} - \bm{\phi}^*\rVert^2\leq \frac{\rho^2(\bm{I}-\eta\bm{A})\lVert\bm{\theta}_{t}-\bm{\theta}^*\rVert^2}{\bm{I}+\eta^2\lambda_{\min}(\bm{B}\bm{B}^T)}
    +\frac{\rho^2(1+\eta\bm{C})\lVert\bm{\phi}_{t} - \bm{\phi}^*\rVert^2}{\bm{I}+\eta^2\lambda_{\min}(\bm{B}^T\bm{B})}.\nonumber
\end{equation}
where $f^* = f(\bm{\theta}^*,\bm{\phi}^*)$. Moreover, for any $\eta$ satisfying:
\begin{equation}
    \frac{\rho^2(\bm{I}-\eta\bm{A})\lVert\bm{\theta}_{t}-\bm{\theta}^*\rVert^2}{\bm{I}+\eta^2\lambda_{\min}(\bm{B}\bm{B}^T)}
    +\frac{\rho^2(1+\eta\bm{C})\lVert\bm{\phi}_{t} - \bm{\phi}^*\rVert^2}{\bm{I}+\eta^2\lambda_{\min}(\bm{B}^T\bm{B})}< \lVert \bm{\theta}_{t} - \bm{\theta}^*\rVert^2 + \lVert\bm{\bm{\phi}_{t} - \bm{\phi}^*}\rVert^2 
\end{equation}
SPPM converges asymptotically to $(\bm{\theta}^*,\bm{\phi}^*)$.
\end{theorem}
\begin{proof}
Following the same setting and procedure as in Appendix \ref{appendix: prove convergence in nonconvex games}, we have that
\begin{equation}
    \lVert \hat{\bm{\theta}}_{t+1}\rVert^2 + \lVert\hat{\bm{\phi}}_{t+1} \rVert^2 = \hat{\bm{\theta}}_{t}^T\bm{P}_{\bm{\theta}}^T(\bm{I} + \eta^2\bm{B}\bm{B}^T)^{-1}\bm{P}_{\bm{\theta}}\hat{\bm{\theta}}_{t} + \hat{\bm{\phi}}_{t}^T\bm{P}_{\bm{\phi}}^T(\bm{I} + \eta^2\bm{B}^T\bm{B})^{-1}\bm{P}_{\bm{\phi}}\hat{\bm{\phi}}_{t}
\end{equation}

Now using the fact that $\bm{P}_{\bm{\theta}} = \bm{P}_{\bm{\theta}}^T$, $\bm{P}_{\bm{\phi}} = \bm{P}_{\bm{\phi}}^T$, we can write
\begin{equation}
    \lVert\bm{\theta}_{t+1} - \bm{\theta}^*\rVert^2 + \lVert\bm{\phi}_{t+1} - \bm{\phi}^*\rVert^2\leq \frac{\rho^2(\bm{I}-\eta\bm{A})\lVert\bm{\theta}_{t}-\bm{\theta}^*\rVert^2}{\bm{I}+\eta^2\lambda_{\min}(\bm{B}\bm{B}^T)}
    +\frac{\rho^2(1+\eta\bm{C})\lVert\bm{\phi}_{t} - \bm{\phi}^*\rVert^2}{\bm{I}+\eta^2\lambda_{\min}(\bm{B}^T\bm{B})}.\nonumber
\end{equation}
For any $\eta$ that satisfying 
\begin{equation}
    \frac{\rho^2(\bm{I}-\eta\bm{A})\lVert\bm{\theta}_{t}-\bm{\theta}^*\rVert^2}{\bm{I}+\eta^2\lambda_{\min}(\bm{B}\bm{B}^T)}
    +\frac{\rho^2(1+\eta\bm{C})\lVert\bm{\phi}_{t} - \bm{\phi}^*\rVert^2}{\bm{I}+\eta^2\lambda_{\min}(\bm{B}^T\bm{B})}< \lVert \bm{\theta}_{t} - \bm{\theta}^*\rVert^2 + \lVert\bm{\bm{\phi}_{t} - \bm{\phi}^*}\rVert^2 
\end{equation}
we have that 
\begin{equation}
    \lVert \bm{\theta}_{t+1} - \bm{\theta}^*\rVert^2 + \lVert\bm{\bm{\phi}_{t+1} - \bm{\phi}^*}\rVert^2<\lVert \bm{\theta}_{t} - \bm{\theta}^*\rVert^2 + \lVert\bm{\bm{\phi}_{t} - \bm{\phi}^*}\rVert^2  
\end{equation}
i.e., SPPM converges asymptotically towards $(\bm{\theta}^*,\bm{\phi}^*)$.
\end{proof}
\subsection{Proof of Theorem \ref{thm: convergence in bilinear game}}
\begin{proof}
In order to proof the convergence of SPPM in bilinear games, we first show that the SPPM update rule is equivalent to that of the following Proximal Point Method:
\begin{equation}
    \begin{cases}
    \bm{\theta}_{t+1} = \bm{\theta}_{t} - \eta \nabla_{\theta}f(\bm{\theta}_{t+1},\bm{\phi}_{t+1})\\
    \bm{\phi}_{t+1} = \bm{\phi}_{t} + \eta \nabla_{\phi}f(\bm{\theta}_{t+1},\bm{\phi}_{t+1})
    \end{cases}
\end{equation}
In the \ref{eq: bilinear game}, the SPPM update is:
\begin{align}
    &\begin{cases}
    \bm{\theta}_{t+1} = \bm{\theta}_{t} - \eta \nabla_{\theta}f(\bm{\theta}_{t},\bm{\phi}_{t+1})\\
    \bm{\phi}_{t+1} = \bm{\phi}_{t} + \eta \nabla_{\phi}f(\bm{\theta}_{t},\bm{\phi}_{t+1})
    \end{cases}\nonumber\\
    &=\begin{cases}
    \bm{\theta}_{t+1} = \bm{\theta}_{t} - \eta \bm{M}\bm{\phi}_{t+1}\\
    \bm{\phi}_{t+1} = \bm{\phi}_{t} + \eta \bm{M}^T\bm{\theta}_{t+1}
    \end{cases}\nonumber
\end{align}
and the PPM update is:
\begin{align}
    &\begin{cases}
    \bm{\theta}_{t+1} = \bm{\theta}_{t} - \eta \nabla_{\theta}f(\bm{\theta}_{t+1},\bm{\phi}_{t+1})\\
    \bm{\phi}_{t+1} = \bm{\phi}_{t} + \eta \nabla_{\phi}f(\bm{\theta}_{t+1},\bm{\phi}_{t+1})
    \end{cases}\nonumber\\
    &=\begin{cases}
    \bm{\theta}_{t+1} = \bm{\theta}_{t} - \eta \bm{M}\bm{\phi}_{t+1}\\
    \bm{\phi}_{t+1} = \bm{\phi}_{t} + \eta \bm{M}^T\bm{\theta}_{t+1}
    \end{cases}\nonumber
\end{align}
Thus SPPM and PPM are equivalent in the \ref{eq: bilinear game}. The convergence result of PPM in bilinear games has been proved in Theorem 2 of  \cite{rockafellar1976monotone}:
\begin{theorem}
Consider the \ref{eq: bilinear game} and the PPM method. Further, we define $r_{t} = \lVert\bm{\theta}_{t} - \bm{\theta}^*\rVert^2 + \lVert\bm{\phi}_{t} - \bm{\phi}^*\rVert^2$. Then, for any $\eta>0$, the iterates $\{\bm{\theta}_{t},\bm{\phi}_{t}\}_{t\geq0}$ generated by SPPM satisfy
\begin{equation}
    r_{t+1}\leq \frac{1}{1+\eta^2\lambda_{\min}(\bm{M}^T\bm{M})}r_{t}.
\end{equation}
\end{theorem}
Therefore, SPPM and PPM have the same convergence property in bilinear games.
\end{proof}

\subsection{Proof of Theorem \ref{thm: convergence in quadratic game}}
\begin{proof}
Consider the learning dynamics:
\begin{align}
    \bm{\theta}_{t+1} &= \bm{\theta}_{t} - \eta\nabla_{\theta}f(\bm{\theta}_{t},\bm{\phi}_{t+1})\nonumber\\
    \bm{\phi}_{t+1} &= \bm{\phi}_{t} + \eta\nabla_{\theta}f(\bm{\theta}_{t+1},\bm{\phi}_{t})\nonumber
\end{align}
In the \ref{eq: quadratic game}, the SPPM update rule can be written as:
\begin{equation}
    \begin{cases}
    \bm{\theta}_{t+1} = \bm{\theta}_{t} - \eta \bm{A}\bm{\theta}_{t} - \bm{C}\bm{\phi}_{t+1}\\
    \bm{\phi}_{t+1} = \bm{\phi}_{t} + \eta \bm{B}\bm{\phi}_{t} + \bm{C}^T\bm{\theta}_{t+1}
    \end{cases}
\end{equation}
Then we can rewrite the learning dynamics:
\begin{align}
    \bm{\theta}_{t+1} &= \bm{\theta}_{t} - \eta \bm{A}\bm{\theta}_{t} - \eta \bm{C}\bm{\phi}_{t+1}\nonumber\\
    \bm{\theta}_{t+1} &= \bm{\theta}_{t} - \eta \bm{A}\bm{\theta}_{t} - \eta \bm{C}(\bm{\phi}_{t} + \eta \bm{C}^T\bm{\theta}_{t+1} + \eta \bm{B}\bm{\phi}_{t})\nonumber\\
    (\bm{I} + \eta^2\bm{CC}^T) \bm{\theta}_{t+1} &= (\bm{I} - \eta \bm{A})\bm{\theta}_{t} - \eta\bm{C}(\bm{I} + \eta\bm{B})\bm{\phi}_{t}\nonumber\\
    \bm{\theta}_{t+1} &= (\bm{I} + \eta^2\bm{CC}^T)^{-1}\left[(\bm{I} - \eta \bm{A})\bm{\theta}_{t} - \eta\bm{C}(\bm{I} + \eta\bm{B})\bm{\phi}_{t}\right]\label{eq: linearized xt+1}
\end{align}
Similarly, for the other player we have
\begin{align}
    \bm{\phi}_{t+1} &= \bm{\phi}_{t} + \eta \bm{C}^T\bm{\theta}_{t+1} + \eta \bm{B}\bm{\phi}_{t}\nonumber\\
    \bm{\phi}_{t+1} &= \bm{\phi}_{t} + \eta \bm{C}^T(\bm{\theta}_{t} - \eta \bm{A}\bm{\theta}_{t} - \eta \bm{C}\bm{\phi}_{t+1}) + \eta \bm{B}\bm{\phi}_{t}\nonumber\\
    (\bm{I} + \eta^2\bm{C}^T\bm{C}) \bm{\phi}_{t+1} &= \eta\bm{C}^T(\bm{I} - \eta \bm{A})\bm{\theta}_{t} + (\bm{I} + \eta\bm{B})\bm{\phi}_{t}\nonumber\\
    \bm{\phi}_{t+1} &= (\bm{I} + \eta^2\bm{C}^T\bm{C})^{-1}\left[ \eta\bm{C}^T(\bm{I} - \eta \bm{A})\bm{\theta}_{t} + (\bm{I} + \eta\bm{B})\bm{\phi}_{t}\right]\label{eq: linearized yt+1}
\end{align}
Let us define the symmetric matrices $\bm{Q}_{\bm{\theta}} = (\bm{I} + \eta^2\bm{CC}^T)^{-1}$, $\bm{Q}_{\bm{\phi}} = (\bm{I} + \eta^2\bm{C}^T\bm{C})^{-1}$ and $\bm{P}_{\bm{\theta}} = (\bm{I} - \eta\bm{A})$, $\bm{P}_{\bm{\phi}} = (\bm{I} + \eta\bm{B})$. Further we define $r_{t}= \lVert \bm{\theta}_{t+1}\rVert ^2 + \lVert \bm{\phi}_{t+1}\rVert ^2$. Based on these definitions, and the expressions in (\ref{eq: linearized xt+1}) and (\ref{eq: linearized yt+1}) we have
\begin{align}
    \lVert \bm{\theta}_{t+1}\rVert ^2 + \lVert \bm{\phi}_{t+1}\rVert ^2
    = \lVert \bm{Q}_{\bm{\theta}}\bm{P}_{\bm{\theta}}\bm{\theta}_{t}\rVert^2 &+ \eta^2\lVert \bm{Q}_{\bm{\theta}}\bm{C}\bm{P}_{\bm{\phi}}\bm{\phi}_{t}\rVert^2 + \lVert \bm{Q}_{\bm{\phi}}\bm{C}^T\bm{P}_{\bm{\theta}}\bm{\theta}_{t}\rVert^2 + \lVert \bm{Q}_{\bm{\phi}}\bm{P}_{\bm{\phi}}\bm{\phi}_{t}\rVert^2\nonumber\\
    &-2\eta\bm{\theta}_{t}^T\bm{P}_{\bm{\theta}}^T\bm{Q}_{\bm{\theta}}^T\bm{Q}_{\bm{\theta}}\bm{C}\bm{P}_{\bm{\phi}}\bm{\phi}_{t}+2\eta\bm{\phi}_{t}^T\bm{P}_{\bm{\phi}}^T\bm{Q}_{\bm{\phi}}^T\bm{Q}_{\bm{\phi}}\bm{C}^T\bm{P}_{\bm{\theta}}\bm{\theta}_{t}\label{eq: sum of norm}
\end{align}
To simplify the expression in (\ref{eq: sum of norm}) we use Lemma \ref{lemma: Q} to obtain the following equations:
\begin{align}
    \bm{Q}_{\bm{\theta}}\bm{C} &= \bm{C}\bm{Q}_{\bm{\phi}}\\
    \bm{Q}_{\bm{\phi}}\bm{C}^T &= \bm{C}^T\bm{Q}_{\bm{\theta}}
\end{align}
Using this lemma, we can show that
\begin{equation}
    \bm{\theta}_{t}^T\bm{P}_{\bm{\theta}}^T\bm{Q}_{\bm{\theta}}^T\bm{Q}_{\bm{\theta}}\bm{C}\bm{P}_{\bm{\phi}}\bm{\phi}_{t}=\bm{\theta}_{t}^T\bm{P}_{\bm{\theta}}^T\bm{Q}_{\bm{\theta}}^T\bm{C}\bm{Q}_{\bm{\phi}}\bm{P}_{\bm{\phi}}\bm{\phi}_{t}=\bm{\phi}_{t}^T\bm{P}_{\bm{\phi}}^T\bm{Q}_{\bm{\phi}}^T\bm{C}^T\bm{Q}_{\bm{\theta}}\bm{P}_{\bm{\theta}}\bm{\theta}_{t}=\bm{\phi}_{t}^T\bm{P}_{\bm{\phi}}^T\bm{Q}_{\bm{\phi}}^T\bm{Q}_{\bm{\phi}}\bm{C}^T\bm{P}_{\bm{\theta}}\bm{\theta}_{t}\nonumber
\end{equation}
where the intermediate equality holds as $\bm{a}^T\bm{C} = \bm{C}^T\bm{a}$. Hence, the expression in (\ref{eq: sum of norm}) can be simplified as
\begin{equation}
    \lVert \bm{\theta}_{t+1}\rVert ^2 + \lVert \bm{\phi}_{t+1}\rVert ^2
    = \lVert \bm{Q}_{\bm{\theta}}\bm{P}_{\bm{\theta}}\bm{\theta}_{t}\rVert^2 + \eta^2\lVert \bm{Q}_{\bm{\theta}}\bm{C}\bm{P}_{\bm{\phi}}\bm{\phi}_{t}\rVert^2 + \lVert \bm{Q}_{\bm{\phi}}\bm{C}^T\bm{P}_{\bm{\theta}}\bm{\theta}_{t}\rVert^2 + \lVert \bm{Q}_{\bm{\phi}}\bm{P}_{\bm{\phi}}\bm{\phi}_{t}\rVert^2\label{eq: norm of sum clean}
\end{equation}
We simplify equation (\ref{eq: norm of sum clean}) as follows. Consider the term involving $\bm{\theta}_{t}$. We have
\begin{align}
    \lVert \bm{Q}_{\bm{\theta}}\bm{P}_{\bm{\theta}}\bm{\theta}_{t}\rVert^2 + \eta^2\lVert \bm{Q}_{\bm{\phi}}\bm{C}^T\bm{P}_{\bm{\theta}}\bm{\theta}_{t}\rVert^2 &= \bm{\theta}_{t}^T\bm{P}_{\bm{\theta}}^T\bm{Q}_{\bm{\theta}}^2\bm{P}_{\bm{\theta}}\bm{\theta}_{t} + \eta^2\bm{\theta}_{t}^T\bm{P}_{\bm{\theta}}^T\bm{C}\bm{Q}_{\bm{\phi}}^2\bm{C}^T\bm{P}_{\bm{\theta}}\bm{\theta}_{t}\nonumber\\
    &= \bm{\theta}_{t}^T\bm{P}_{\bm{\theta}}^T(\bm{Q}_{\bm{\theta}}^2 + \eta^2\bm{C}\bm{Q}_{\bm{\phi}}^2\bm{C}^T)\bm{P}_{\bm{\theta}}\bm{\theta}_{t}\nonumber\\
    &= \bm{\theta}_{t}^T\bm{P}_{\bm{\theta}}^T(\bm{Q}_{\bm{\theta}}^2 + \eta^2\bm{C}\bm{Q}_{\bm{\phi}}\bm{C}^T\bm{Q}_{\bm{\theta}})\bm{P}_{\bm{\theta}}\bm{\theta}_{t}\nonumber\\
    &= \bm{\theta}_{t}^T\bm{P}_{\bm{\theta}}^T(\bm{Q}_{\bm{\theta}}^2 + \eta^2\bm{C}\bm{C}^T\bm{Q}_{\bm{\theta}}\bm{Q}_{\bm{\theta}})\bm{P}_{\bm{\theta}}\bm{\theta}_{t}\nonumber\\
    &= \bm{\theta}_{t}^T\bm{P}_{\bm{\theta}}^T(\bm{I} + \eta^2\bm{C}\bm{C}^T)\bm{Q}_{\bm{\theta}}^2\bm{P}_{\bm{\theta}}\bm{\theta}_{t}\nonumber\\
    &= \bm{\theta}_{t}^T\bm{P}_{\bm{\theta}}^T(\bm{I} + \eta^2\bm{C}\bm{C}^T)^{-1}\bm{P}_{\bm{\theta}}\bm{\theta}_{t}\label{eq: expression for xt2}
\end{align}
where the last equality follows by replacing $\bm{Q}_{\bm{\theta}}$ by its definition. The same procedure follows for the term involving $\bm{\phi}_{t}$ which leads to the expression
\begin{equation}
    \lVert \bm{Q}_{\bm{\phi}}\bm{P}_{\bm{\phi}}\bm{\phi}_{t}\rVert^2 + \eta^2\lVert \bm{Q}_{\bm{\theta}}\bm{C}\bm{P}_{\bm{\phi}}\bm{\phi}_{t}\rVert^2 = \bm{\phi}_{t}^T\bm{P}_{\bm{\phi}}^T(\bm{I} + \eta^2\bm{C}^T\bm{C})^{-1}\bm{P}_{\bm{\phi}}\bm{\phi}_{t}. \label{eq: expression for yt2}
\end{equation}
Substitute $\lVert \bm{Q}_{\bm{\theta}}\bm{P}_{\bm{\theta}}\bm{\theta}_{t}\rVert^2 + \eta^2\lVert \bm{Q}_{\bm{\phi}}\bm{C}^T\bm{P}_{\bm{\theta}}\bm{\theta}_{t}\rVert^2$ and $\lVert \bm{Q}_{\bm{\phi}}\bm{P}_{\bm{\phi}}\bm{\phi}_{t}\rVert^2 + \eta^2\lVert \bm{Q}_{\bm{\theta}}\bm{C}\bm{P}_{\bm{\phi}}\bm{\phi}_{t}\rVert^2 $ in (\ref{eq: norm of sum clean}) with the expressions in (\ref{eq: expression for xt2}) and (\ref{eq: expression for yt2}), respectively, to obtain
\begin{equation}
    \lVert \bm{\theta}_{t+1}\rVert^2 + \lVert\bm{\phi}_{t+1} \rVert^2 = \bm{\theta}_{t}^T\bm{P}_{\bm{\theta}}^T(\bm{I} + \eta^2\bm{C}\bm{C}^T)^{-1}\bm{P}_{\bm{\theta}}\bm{\theta}_{t} + \bm{\phi}_{t}^T\bm{P}_{\bm{\phi}}^T(\bm{I} + \eta^2\bm{C}^T\bm{C})^{-1}\bm{P}_{\bm{\phi}}\bm{\phi}_{t}.\label{eq: sum of norm result}
\end{equation}
Now using the expression in (\ref{eq: sum of norm result}) and the fact that $\bm{P}_{\bm{\theta}} = \bm{P}_{\bm{\theta}}^T$, $\bm{P}_{\bm{\phi}} = \bm{P}_{\bm{\phi}}^T$ and $\lambda_{\min}(\bm{C}^T\bm{C}) = \lambda_{\min}(\bm{C}\bm{C}^T)$, we can write
\begin{equation}
    \lVert\bm{\theta}_{t+1} - \bm{\theta}^*\rVert^2 + \lVert\bm{\phi}_{t+1} - \bm{\phi}^*\rVert^2\leq \frac{\rho^2(\bm{I}-\eta\bm{A})\lVert\bm{\theta}_{t} - \bm{\theta}^*\rVert^2+\rho^2(1+\eta\bm{B})\lVert\bm{\phi}_{t} - \bm{\phi}^*\rVert^2}{\bm{I}+\eta^2\lambda_{\min}(\bm{C}^T\bm{C})}.\nonumber
\end{equation}
\end{proof}
\subsection{Competitive Gradient Descent as an Approximation of SPPM}
In this section, we justify our results in Section 4.1 that Competitive Gradient Descent is a first order Taylor approximation of SPPM. 
Firstly, we consider the standard definition of CGD:
\begin{align}
\begin{cases}
    \bm{\theta}_{t+1} = \bm{\theta}_{t} - \eta(\bm{I} + \eta^2\nabla_{\theta\phi}f(\bm{\theta}_{t},\bm{\phi}_{t})\nabla_{\phi\theta}f(\bm{\theta}_{t},\bm{\phi}_{t}))^{-1}(\nabla_{\theta}f(\bm{\theta}_{t},\bm{\phi}_{t}) + \eta\nabla_{\theta\phi}f(\bm{\theta}_{t},\bm{\phi}_{t})\nabla_{\phi}f(\bm{\theta}_{t},\bm{\phi}_{t}))\nonumber\\
    \bm{\phi}_{t+1} = \bm{\phi}_{t} + \eta(\bm{I} + \eta^2\nabla_{\phi\theta}f(\bm{\theta}_{t},\bm{\phi}_{t})\nabla_{\theta\phi}f(\bm{\theta}_{t},\bm{\phi}_{t}))^{-1}(\nabla_{\phi}f(\bm{\theta}_{t},\bm{\phi}_{t}) - \eta\nabla_{\phi\theta}f(\bm{\theta}_{t},\bm{\phi}_{t})\nabla_{\theta}f(\bm{\theta}_{t},\bm{\phi}_{t}))\nonumber
\end{cases}
\end{align}
Rewriting the update rules we can get:
\begin{align}
    &(\bm{I} + \eta^2\nabla_{\theta\phi}f(\bm{\theta}_{t},\bm{\phi}_{t})\nabla_{\phi\theta}f(\bm{\theta}_{t},\bm{\phi}_{t}))(\bm{\theta}_{t+1}-\bm{\theta}_{t})=-\eta(\nabla_{\theta}f(\bm{\theta}_{t},\bm{\phi}_{t}) + \eta\nabla_{\theta\phi}f(\bm{\theta}_{t},\bm{\phi}_{t})\nabla_{\phi}f(\bm{\theta}_{t},\bm{\phi}_{t}))\nonumber\\
    &\bm{\theta}_{t+1} =\bm{\theta}_{t} - \eta \nabla_{\theta}f(\bm{\theta}_{t},\bm{\phi}_{t}) - \eta^2 \nabla_{\theta\phi}f(\bm{\theta}_{t},\bm{\phi}_{t}) \nabla_{\phi}f(\bm{\theta}_{t},\bm{\phi}_{t}) - \eta^2 \nabla_{\theta\phi}f(\bm{\theta}_{t},\bm{\phi}_{t})\nabla_{\phi\theta}f(\bm{\theta}_{t},\bm{\phi}_{t})(\bm{\theta}_{t+1} - \bm{\theta}_{t})\nonumber
\end{align}
Similarly, we have:
\begin{align}
    \bm{\phi}_{t+1} = \bm{\phi}_{t} + \eta \nabla_{\theta}f(\bm{\theta}_{t},\bm{\phi}_{t}) - \eta^2 \nabla_{\phi\theta}f(\bm{\theta}_{t},\bm{\phi}_{t}) \nabla_{\theta}f(\bm{\theta}_{t},\bm{\phi}_{t}) - \eta^2 \nabla_{\phi\theta}f(\bm{\theta}_{t},\bm{\phi}_{t})\nabla_{\theta\phi}f(\bm{\theta}_{t},\bm{\phi}_{t})(\bm{\phi}_{t+1} - \bm{\phi}_{t})\nonumber
\end{align}
Therefore, CGD is a first order approximation of SPPM.
Then we prove that the standard definition of CGD is equivalent to the update rule in Table \ref{tab:precisions of algorithms}. Consider the update rule in Table \ref{tab:precisions of algorithms} and its footnote, we have:
\begin{align}
    \begin{cases}
    \bm{\theta}_{t+1} = \bm{\theta}_{t} - \eta\nabla_{\theta}f(\bm{\theta}_{t},\bm{\phi}_{t}) - \eta\nabla_{\theta\phi}f(\bm{\theta}_{t},\bm{\phi}_{t})(\bm{\phi}_{t+1}-\bm{\phi}_{t})\\
    \bm{\phi}_{t+1} = \bm{\phi}_{t} +\eta\nabla_{\phi}f(\bm{\theta}_{t},\bm{\phi}_{t}) + \eta\nabla_{\phi\theta}f(\bm{\theta}_{t},\phi_{t})(\bm{\theta}_{t+1}-\bm{\theta}_{t})
    \end{cases}\label{cases: CGD update}
\end{align}
Substituting $(\bm{\phi}_{t+1} - \bm{\phi}_{t})$ into the first equation of (\ref{cases: CGD update}) we get:
\begin{align}
    &\bm{\theta}_{t+1} = \bm{\theta}_{t} - \eta\nabla_{\theta}f(\bm{\theta}_{t},\bm{\phi}_{t}) - \eta\nabla_{\theta\phi}f(\bm{\theta}_{t},\bm{\phi}_{t})(\eta\nabla_{\phi}f(\bm{\theta}_{t},\bm{\phi}_{t}) + \eta\nabla_{\phi\theta}f(\bm{\theta}_{t},\phi_{t})(\bm{\theta}_{t+1}-\bm{\theta}_{t}))\nonumber\\
    &\bm{\theta}_{t+1} = \bm{\theta}_{t} - \eta\nabla_{\theta}f(\bm{\theta}_{t},\bm{\phi}_{t}) - \eta^2\nabla_{\theta\phi}f(\bm{\theta}_{t},\bm{\phi}_{t})\nabla_{\phi}f(\bm{\theta}_{t},\bm{\phi}_{t}) - \eta^2\nabla_{\theta\phi}f(\bm{\theta}_{t},\bm{\phi}_{t})\nabla_{\phi\theta}f(\bm{\theta}_{t},\phi_{t})(\bm{\theta}_{t+1}-\bm{\theta}_{t})\nonumber\\
    &(\bm{I} + \eta^2\nabla_{\theta\phi}f(\bm{\theta}_{t},\bm{\phi}_{t})\nabla_{\phi}f(\bm{\theta}_{t},\bm{\phi}_{t}) )(\bm{\theta}_{t+1}-\bm{\theta}_{t}) = - \eta\nabla_{\theta}f(\bm{\theta}_{t},\bm{\phi}_{t}) - \eta^2\nabla_{\theta\phi}f(\bm{\theta}_{t},\bm{\phi}_{t})\nabla_{\phi}f(\bm{\theta}_{t},\bm{\phi}_{t})\nonumber\\
    &\bm{\theta}_{t+1} = \bm{\theta}_{t} - \eta(\bm{I} + \eta^2\nabla_{\theta\phi}f(\bm{\theta}_{t},\bm{\phi}_{t})\nabla_{\phi\theta}f(\bm{\theta}_{t},\bm{\phi}_{t}))^{-1}(\nabla_{\theta}f(\bm{\theta}_{t},\bm{\phi}_{t}) + \eta\nabla_{\theta\phi}f(\bm{\theta}_{t},\bm{\phi}_{t})\nabla_{\phi}f(\bm{\theta}_{t},\bm{\phi}_{t}))\nonumber.
\end{align}
Substituting $(\bm{\theta}_{t+1} - \bm{\theta}_{t})$ into the second equation of (\ref{cases: CGD update}) and applying similar arguments we get:
\begin{equation}
    \bm{\phi}_{t+1} = \bm{\phi}_{t} + \eta(\bm{I} + \eta^2\nabla_{\phi\theta}f(\bm{\theta}_{t},\bm{\phi}_{t})\nabla_{\theta\phi}f(\bm{\theta}_{t},\bm{\phi}_{t}))^{-1}(\nabla_{\phi}f(\bm{\theta}_{t},\bm{\phi}_{t}) - \eta\nabla_{\phi\theta}f(\bm{\theta}_{t},\bm{\phi}_{t})\nabla_{\theta}f(\bm{\theta}_{t},\bm{\phi}_{t}))\nonumber
\end{equation}
Thus the update rule in Table \ref{tab:precisions of algorithms} is equivalent to the standard definition of CGD and it is equivalent to the first order Taylor approximation of SPPM.
\subsection{Experiments on Bilinear and Quadratic Games}
\begin{figure}[h]
\centering
\makebox[\textwidth][c]{\includegraphics[width=1.25\textwidth]{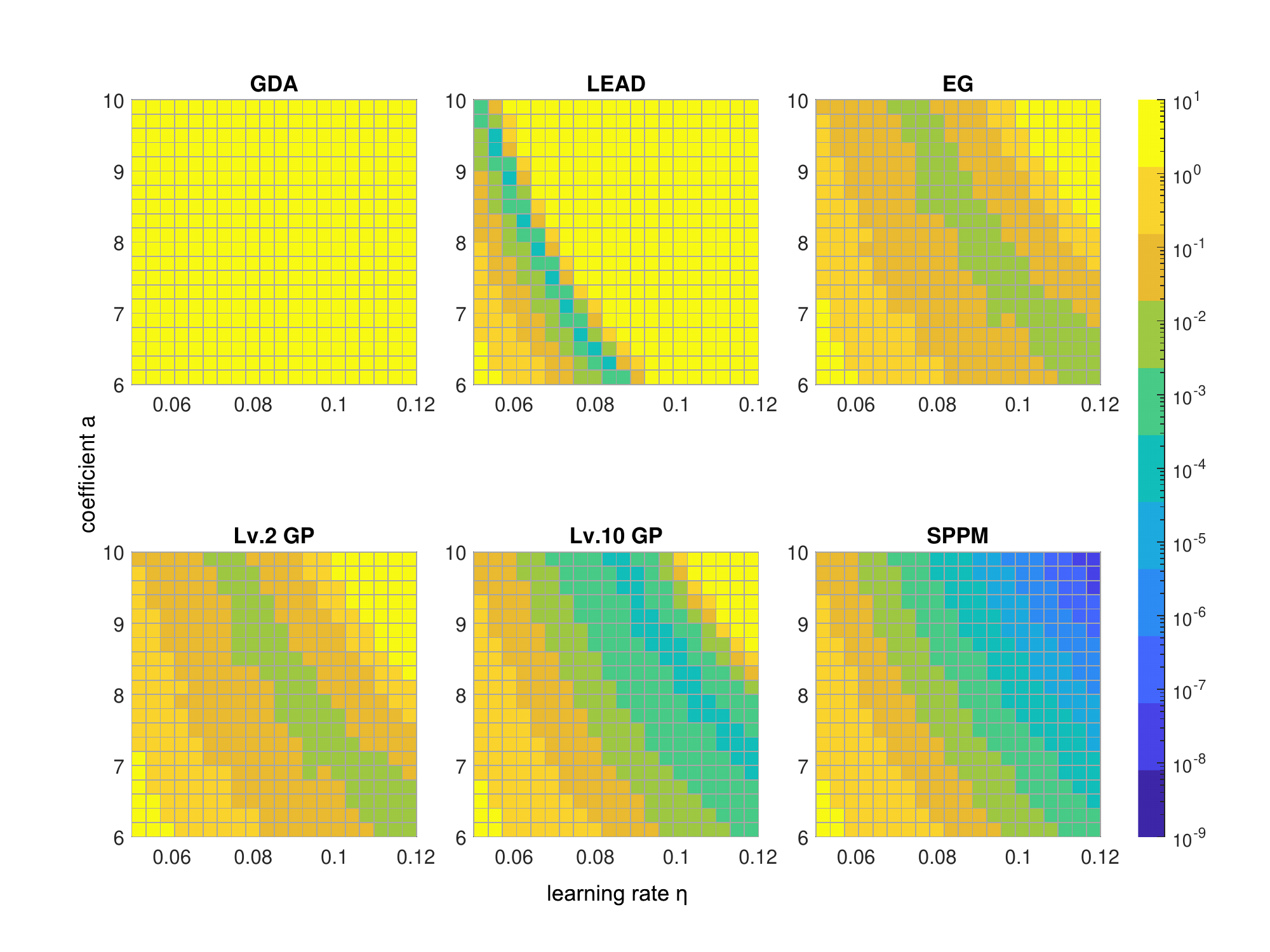}}
\vspace*{-1mm}\caption{A grid of experiments on the bilinear game for different algorithms with different values of coefficient $a$ and learning rates $\eta$. The color in each cell indicates the distance to the equilibrium after 50 iterations.} 
\label{fig: distance in bilinear games}
\end{figure}

\paragraph{Bilinear Game}
Consider the following bilinear game:
\begin{equation}
    \min_{\theta\in\mathbb{R}}\max_{\phi\in\mathbb{R}} a\theta\phi
\end{equation}
the example presented in Figure \ref{fig:convergence of LVk} is an example of using different algorithms to solve the bilinear game above with coefficient $a = 10$. For sake of completeness, we also provide a grid of experiment results for different algorithms with different coefficients $a$ and learning rates $\eta$, starting from the same point $(\theta_{0},\phi_{0}) = (-12,10)$. The result is presented in Figure \ref{fig: distance in bilinear games}.
The experiment demonstrates that, in a bilinear game, Lv.$2$ GP is equivalent to the extra-gradient method, and higher level Lv.$k$ GP performs better with increased coefficient $a$ and learning rate $\eta$ as long as it remains a contraction (i.e., $\eta<a^{-1}$).
\paragraph{Quadratic Game}
For the quadratic game presented in Figure \ref{fig:quadratic game example}, we randomly initialize the matrices $\bm{A}$ and $\bm{B}$:
\begin{align}
    \bm{A} = \begin{bmatrix}
    1.8398 & 0.5195 & 1.2537 & 1.7470 & 1.2769\\
    0.5195 & 0.6586 & 0.4476 & 0.8898 & 1.1309\\
    1.2537 & 0.4476 & 1.4440 & 1.3923 & 0.8877\\
    1.7470 & 0.8898 & 1.3923 & 2.1249 & 1.7664\\
    1.2769 & 1.1309 & 0.8877 & 1.7664 & 2.1553
    \end{bmatrix}
    \bm{B} = -\begin{bmatrix}
    1.0821 & 1.2427 & 1.0093 & 1.3335 & 0.6761\\
    1.2427 & 2.2031 & 1.3236 & 1.8566 & 0.9394\\
    1.0093 & 1.3236 & 1.2393 & 1.3675 & 0.9065\\
    1.3335 & 1.8566 & 1.3675 & 1.9081 & 0.9693\\
    0.6761 & 0.9394 & 0.9065 & 0.9693 & 0.7141
    \end{bmatrix}\nonumber
\end{align}
where $\bm{A}$ is symmetric and positive definite and $\bm{B}$ is symmetric and negative definite. The interaction matrix is defined as:
\begin{align}
    \bm{C} = \begin{bmatrix}
    c & 0 & 0 & 0 & 0\\
    0 & c & 0 & 0 & 0\\
    0 & 0 & c & 0 & 0\\
    0 & 0 & 0 & c & 0\\
    0 & 0 & 0 & 0 & c
    \end{bmatrix}\nonumber
\end{align}
where $c$ represents the strength of the interaction between the two players. The starting point $\bm{\theta}_{0}$ and $\bm{\phi}_{0}$ are $[0.1270,0.9667,0.2605,0.8972,0.3767]^T$ and $[0.3362,0.4514,0.8403,0.1231,0.5430]^T$ respectively.
\begin{figure}[h]
\centering
\makebox[\textwidth][c]{\includegraphics[width=1.25\textwidth]{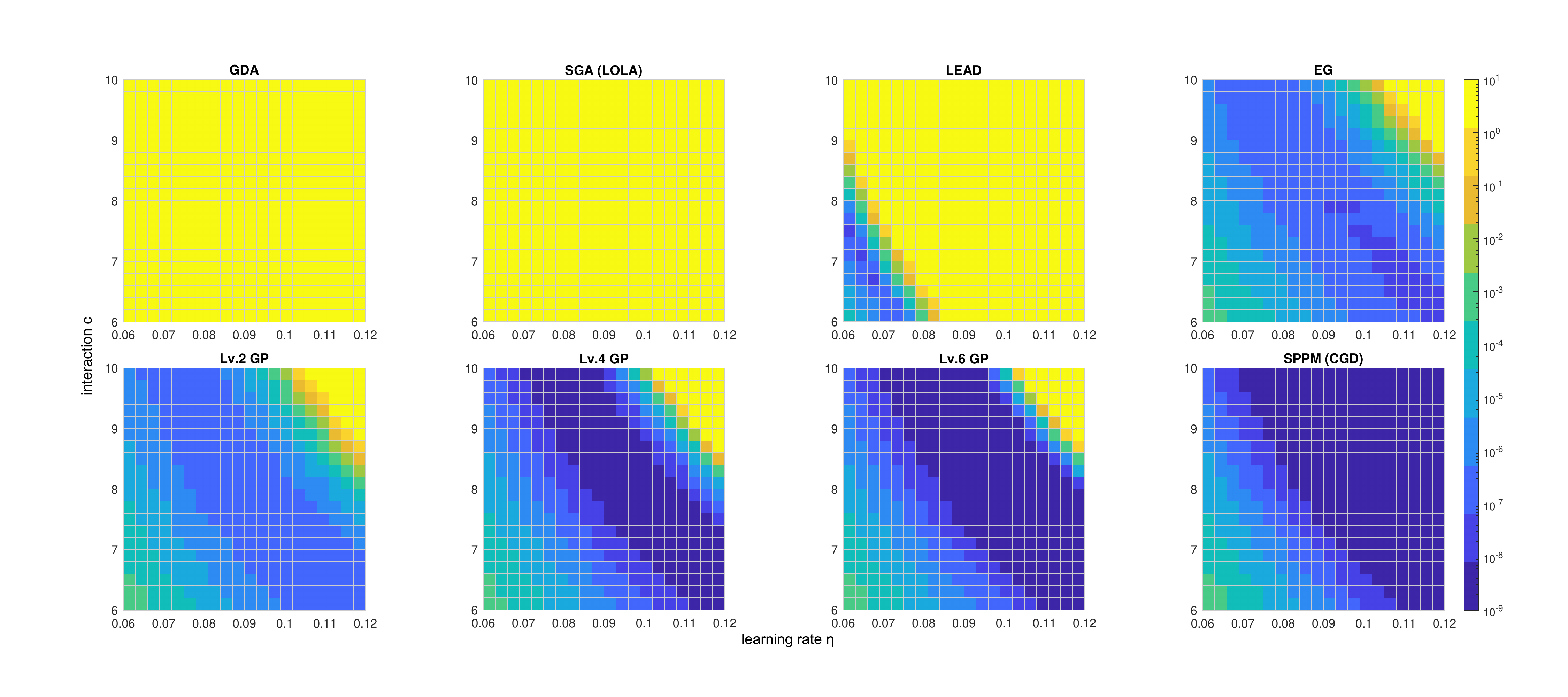}}
\vspace*{-1mm}\caption{A grid of experiments on the quadratic game for different algorithms with different values of coefficient $a$ and learning rates $\eta$. The color in each cell indicates the distance to the equilibrium after 50 iterations.} 
\label{fig: new quadratic game 8}
\end{figure}
\subsection{Experiments on 8-Gaussians}
\paragraph{Dataset} The target distribution is a mixture of 8-Gaussians with standard deviation equal to $0.05$ and modes uniformly distributed around a unit circle.
  
\paragraph{Experiment}
 For our experiments, we used the PyTorch framework. Furthermore, the batch size we used is $128$.
\subsection{Experiments on CIFAR-10 and STL-10}\label{appendix: cifar10 and stl10}
For our experiments, we used the PyTorch\footnote{\url{https://pytorch.org/}} framework. For experiments on CIFAR-10 and STL-10, we compute the FID and IS metrics using the provided implementations in Tensorflow\footnote{\url{https://tensorflow.org/}} for consistency with related works.
\paragraph{Lv.$k$ GP vs Lv.$k$ Adam}
\begin{figure}[h]
    \centering
    \includegraphics[width=\textwidth]{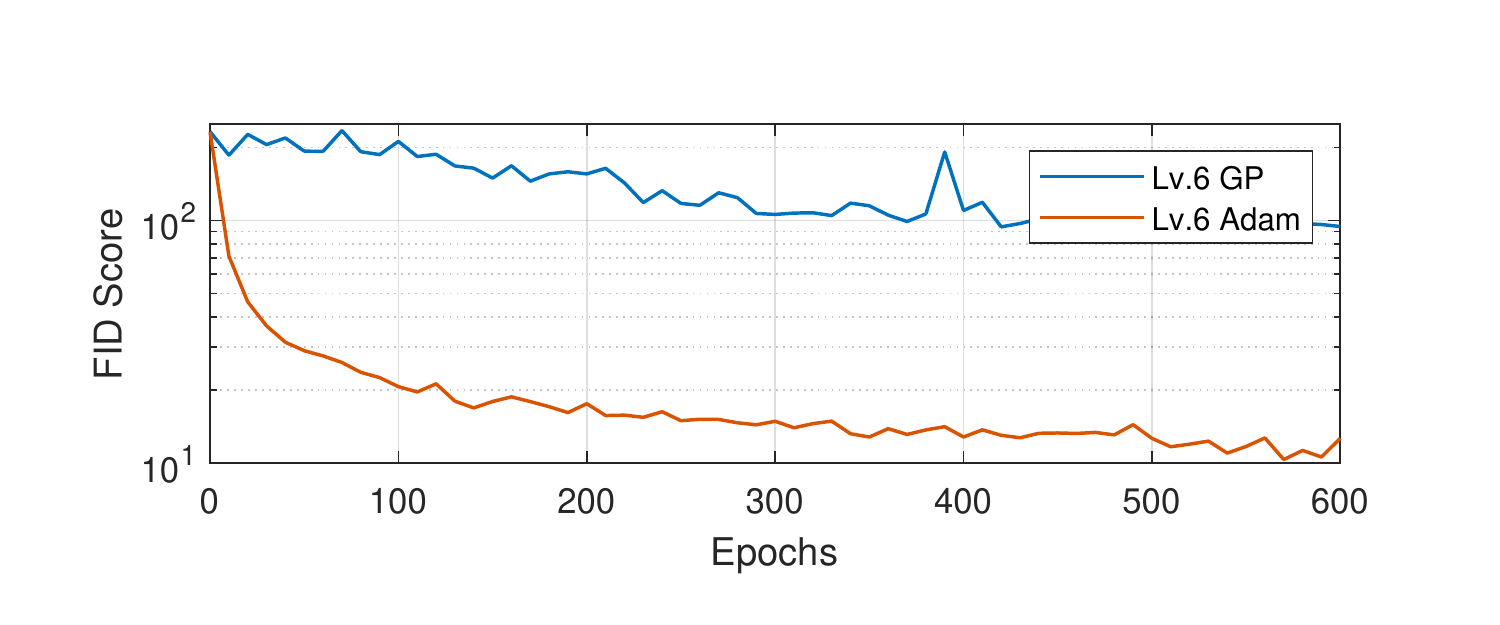}
    \caption{Comparison between Lv.$k$ GP and Lv.$k$ Adam on generating CIFAR-10 images. We can see significant improvements in FID when using the Lv.$k$ Adam algorithm we proposed.}
    \label{fig: LvkGP vs LvkAdam}
\end{figure}
In experiments, we compare the performance of Lv.$k$ GP and Lv.$k$ Adam on the task of CIFAR-10 image generation. The experiment results is presented in Figure \ref{fig: LvkGP vs LvkAdam}. The experiments on Lv.$k$ GP and Lv.$k$ Adam use the same initialization and hyperparameters. According to our experiments, Lv.$k$ Adam converges much faster than Lv.$k$ GP for the same choice of $k$ and learning rates.
\paragraph{Adam vs Lv.$k$ Adam}
\begin{figure}[h]
    \centering
    \includegraphics[width=\textwidth]{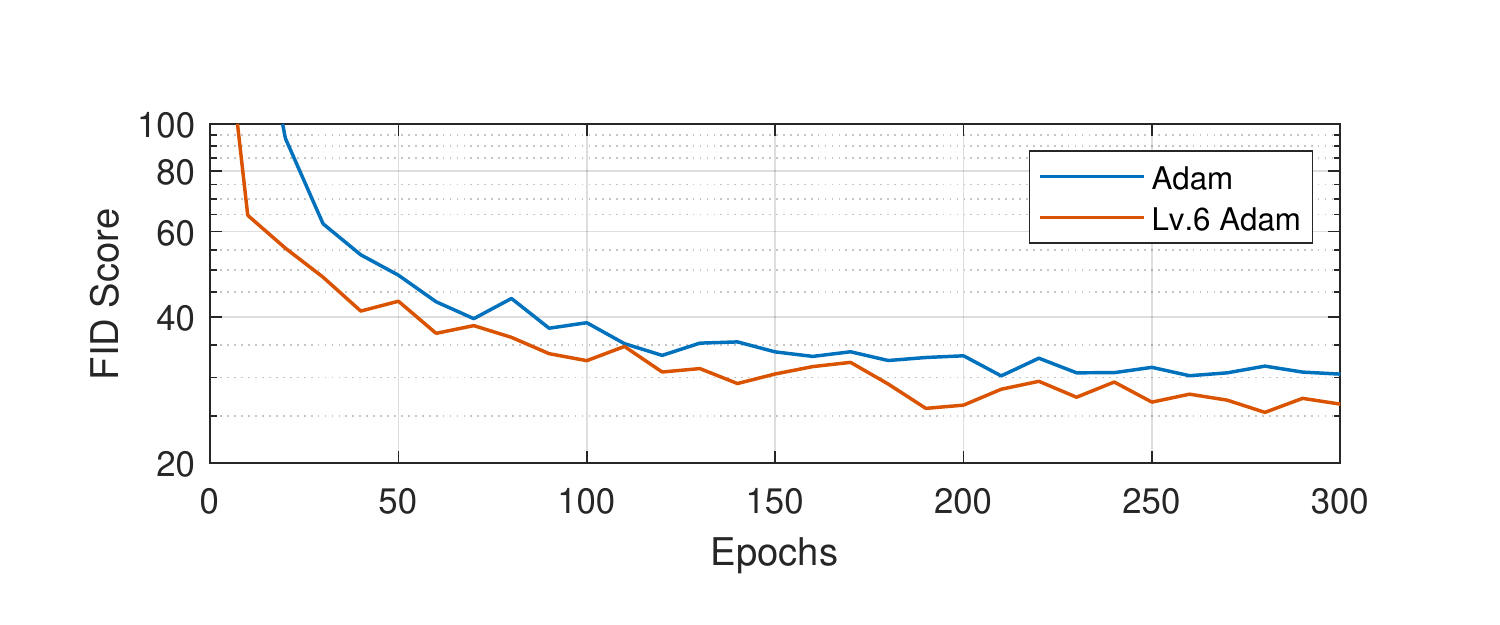}
    \caption{Comparison between Adam and Lv.$k$ Adam on generating STL-10 images. We can see that Lv.$k$ Adam consistently outperforms the Adam optimizer in terms of FID score.}
    \label{fig:STL-10 exp comparison}
\end{figure}
We also present a comparison between the performance of Adam and Lv.$k$ Adam optimizers on the task of STL-10 image generation. The experiment results is presented in Figure \ref{fig:STL-10 exp comparison}. Under the same choice of hyperparameters and identical model parameter initialization, Lv.$k$ Adam consistently outperforms the Adam optimizer in terms of FID score.
\paragraph{Accelerated Lv.$k$ Adam}
In this section, we propose an accelerated version of Lv.$k$ Adam. The intuition is that we update the min player $\bm{\theta}$ and the max player $\bm{\phi}$ in an alternating order. The corresponding Lv.$k$ GP algorithm can be writen as:
\begin{equation}
\mathmakebox[0.8\textwidth]{
\text{Reasoning:}
\begin{cases}
        \bm{\theta}^{(k)}_{t} = \bm{\theta}_{t} - \eta\nabla_{\bm{\theta}}f(\bm{\theta}_{t},\bm{\phi}_{t}^{(k-1)})\nonumber\\
        \bm{\phi}^{(k)}_{t} = \bm{\phi}_{t} - \eta\nabla_{\bm{\phi}}g(\bm{\theta}_{t}^{(k)},\bm{\phi}_{t})
\end{cases}
\hspace{-0.4cm}\text{Update:}
\begin{cases}
    \bm{\theta}_{t+1} = \bm{\theta}_{t} - \eta\nabla_{\bm{\theta}}f(\bm{\theta}_{t},\bm{\phi}_{t}^{(k)})\nonumber\\
    \bm{\phi}_{t+1} = \bm{\phi}_{t} - \eta\nabla_{\bm{\phi}}g(\bm{\theta}_{t}^{(k)},\bm{\phi}_{t})
\end{cases}}\label{eq: alternating level k gradient play}\tag{Alt-Lv.$k$ GP}
\end{equation}
Instead of responding to $\bm{\theta}^{(k-1)}_{t}$, in Alt-Lv.$k$ GP, the max player $\bm{\phi}_{t}^{(k)}$ acts in response to the min player's current action, $\bm{\theta}^{(k)}_{t}$. A Lv.$k$ min player in Alt-Lv.$k$ GP is equivalent to a Lv.$2k-1$ player in the Lv.$k$ GP, and a Lv.$k$ max player in Alt-Lv.$k$ GP is equivalent to a Lv.$2k$ player in the Lv.$k$ GP, respectively. Therefore, it is easy to verify that Alt-Lv.$k$ GP converges two times faster than Lv.$k$ GP and the corresponding Alt-Lv.$k$ Adam algorithm is provided in Algorithm \ref{algo: alt-level k adam}.
\begin{algorithm}[h]
\caption{Accelerated Level $k$ Adam: proposed Adam with recursive reasoning steps}\label{algo: alt-level k adam}
\KwIn{Stopping time $T$, reasoning steps $k$, learning rate $\eta_{\bm{\theta}},\eta_{\bm{\phi}}$, decay rates for momentum estimates $\beta_1,\beta_2$, initial weight $(\bm{\theta}_{0},\bm{\phi}_{0})$, $\bm{P}_{\vx}$ and $\bm{P}_{\vz}$ real and noise-data distributions, losses $\mathcal{L}_{G}(\bm{\theta},\bm{\phi},\vx,\vz)$ and $\mathcal{L}_{D}(\bm{\theta},\bm{\phi},\vx,\vz)$, $\epsilon=1e-8$.}
\Parameter{Initial parameters: $\bm{\theta}_{0},\bm{\phi}_{0}$\\
Initialize first moments:$\bm{m}_{\theta,0} \xleftarrow{} 0,\bm{m}_{\phi,0} \xleftarrow{} 0$\\
Initialize second moments:$\bm{v}_{\theta,0} \xleftarrow{} 0,\bm{v}_{\phi,0} \xleftarrow{} 0$}
\For{t=0,\dots,T-1}{
    \text{\textbf{Sample} new mini-batch:} $\vx,\vz\sim\bm{P}_{\vx},\bm{P}_{\vz}$,\\
    $\bm{\theta}_{t}^{(0)} \xleftarrow{} \bm{\theta}_{t},\bm{\phi}_{t}^{(0)} \xleftarrow{} \bm{\phi}_{t}$,\\
    \For{n=1,\dots,k}{
        Compute stochastic gradient: $\bm{g}_{\bm{\theta},t}^{(n)} = \nabla_{\theta}\mathcal{L}_{G}(\bm{\theta}_{t},\bm{\phi}^{(n-1)}_{t},\vx,\vz)$;\\
        Update estimate of first moment:
        $\bm{m}_{\theta,t}^{(n)}=\beta_1\bm{m}_{\theta,t-1} + (1-\beta_1)\bm{g}^{(n)}_{\theta,t}$;\\
        Update estimate of second moment:
        $\bm{v}_{\theta,t}^{(n)}=\beta_2\bm{v}_{\theta,t-1} + (1-\beta_2)(\bm{g}^{(n)}_{\theta,t})^2$;\\
        Correct the bias for the moments: $\bm{\hat{m}}_{\theta,t}^{(n)} = \frac{\bm{m}^{(n)}_{\theta,t}}{(1-\beta_1^{t})},\bm{\hat{v}}_{\theta,t}^{(n)} = \frac{\bm{v}^{(n)}_{\theta,t}}{(1-\beta_2^{t})}$;\\
        Perform Adam update:   $\bm{\theta}^{(n)}_{t} = \bm{\theta}_{t} - \eta_{\theta}\frac{\bm{\hat{m}}_{\theta,t}^{(n)}}{\sqrt{\bm{\hat{v}}_{\theta,t}^{(n)}}+\epsilon}$;\\
        Compute stochastic gradient: $\bm{g}_{\bm{\phi},t}^{(n)} =\nabla_{\phi}\mathcal{L}_{D}(\bm{\theta}^{(n)}_{t},\bm{\phi}_{t},\vx,\vz)$;\\
        Update estimate of first moment:$\bm{m}_{\phi,t}^{(n)}=\beta_1\bm{m}_{\phi,t-1} + (1-\beta_1)\bm{g}^{(n)}_{\phi,t}$;\\
        Update estimate of second moment:$\bm{v}_{\phi,t}^{(n)}=\beta_2\bm{v}_{\phi,t-1} + (1-\beta_2)(\bm{g}^{(n)}_{\phi,t})^2$;\\
        Correct the bias for the moments:$\bm{\hat{m}}_{\phi,t}^{(n)} = \frac{\bm{m}^{(n)}_{\phi,t}}{(1-\beta_1^{t})},\bm{\hat{v}}_{\phi,t}^{(n)} = \frac{\bm{v}^{(n)}_{\phi,t}}{(1-\beta_2^{t})}$;\\
        Perform Adam update:$\bm{\phi}^{(n)}_{t} = \bm{\phi}_{t} - \eta_{\phi}\frac{\bm{\hat{m}}_{\phi,t}^{(n)}}{\sqrt{\bm{\hat{v}}_{\phi,t}^{(n)}}+\epsilon}$;\\
    }
    $\bm{\theta}_{t+1} \xleftarrow{} \bm{\theta}_{t}^{(k)},\bm{\phi}_{t+1} \xleftarrow{} \bm{\phi}_{t}^{(k)}$;\\
    $\bm{m}_{\theta,t} \xleftarrow{} \bm{m}_{\theta,t}^{(k)},\bm{m}_{\phi,t} \xleftarrow{} \bm{m}_{\phi,t}^{(k)}$;\\
    $\bm{v}_{\theta,t} \xleftarrow{} \bm{v}_{\theta,t}^{(k)},\bm{v}_{\phi,t} \xleftarrow{} \bm{v}_{\phi,t}^{(k)}$\\
}
\end{algorithm}
\paragraph{Architecture}
In this section, we describe the model we used to evaluate the performance of Lv.$k$ Adam for generating CIFAR-10\footnote{CIFAR10 is released under the MIT license.} and STL-10 datasets. With 'conv' we denote a convolutional layer and 'transposed conv' a transposed convolution layer. The models use Batch Normalization and Spectral Normalization. The model's parameters are initialized with Xavier initialization.
\begin{table}[]
    \caption{ResNet blocks used for the SN-GAN architectures on CIFAR-10 image generation, for the generator (left) and the discriminator (right).}
    \label{tab:my_label}
    \begin{subtable}[c]{0.5\textwidth}
    \centering
    \begin{tabular}{c}\toprule
    \textbf{G-ResBlock}\\\midrule
    \multicolumn{1}{l}{\textit{Shortcut:}}\\
    Upsample($\times2$)\\
    \multicolumn{1}{l}{\textit{Residual:}}\\
    Batch Normalization\\
    ReLU\\
    Upsample($\times2$)\\
    conv (ker:$3\times3$, $256\to256$; stride: $1$; pad: $1$)\\
    Batch Normalization\\
    ReLU\\
    conv (ker:$3\times3$, $256\to256$; stride: $1$; pad: $1$)\\\bottomrule
    \end{tabular}
    \end{subtable}
    \begin{subtable}[c]{0.5\textwidth}
    \centering
    \begin{tabular}{c}\toprule
    \textbf{D-ResBlock ($l-$th block)}\\\midrule
    \multicolumn{1}{l}{\textit{Shortcut:}}\\
    $[$AvgPool (ker: $2\times2$)$]$, if $l=1$\\
    conv (ker: $1\times1$, $3_{l=1}/128_{l\neq1}\to128$; stride: $1$)\\
    Spectral Normalization\\
    $[$AvgPool (ker: $2\times2$, stride: $2$)$]$, if $l=1$\\
    \multicolumn{1}{l}{\textit{Residual:}}\\
    $[$ReLU$]$, if $l\neq 1$\\
    conv (ker: $3\times3$, $3_{l=1}/128_{l\neq1}\to128$; stride: $1$; pad: $1$)\\
    Spectral Normalization\\
    ReLU\\
    conv (ker: $1\times1$, $128\to128$; stride: $1$)\\
    Spectral Normalization\\
    AvgPool (ker: $2\times 2$)\\\bottomrule
    \end{tabular}
    \end{subtable}
\end{table}

\begin{table}[h]
    \centering
    \caption{SN-GAN architectures for experiments on CIFAR-10}
    \label{tab:8-Gaussians}
    \begin{tabular}{c c}\toprule
    \textbf{Generator}     &  \textbf{Discriminator}\\\midrule
    \textit{Input:} $\vz\in\mathbb{R}^{128}\sim\mathcal{N}(0,\bm{I})$     & \textit{Input:} $\vx\in\mathbb{R}^{3\times32\times32}$\\
    Linear($128\to4096$) & D-ResBlock\\
    G-ResBlock & D-ResBlock\\
    G-ResBlock & D-ResBlock\\
    G-ResBlock & D-ResBlock\\
    Batch Normalization & ReLU\\
    ReLU & AvgPool(ker:$8\times8$)\\
    conv (ker:$3\times3$, $256\to3$; stride: $1$; pad: $1$) & Linear($128\to1$)\\
    Tanh & Spectral Normalization\\
    \bottomrule
    \end{tabular}
\end{table}

\begin{table}[h]
    \centering
    \caption{SN-GAN architectures for experiments on STL-10}
    \label{tab:8-Gaussians}
    \begin{tabular}{c c}\toprule
    \textbf{Generator}     &  \textbf{Discriminator}\\\midrule
    \textit{Input:} $\vz\in\mathbb{R}^{128}\sim\mathcal{N}(0,\bm{I})$     & \textit{Input:} $\vx\in\mathbb{R}^{3\times48\times48}$\\
    Linear($128\to6\times6\times512$) & D-ResBlock down $64\to128$\\
    G-ResBlock up $512\to256$ & D-ResBlock down $3\to128$\\
    G-ResBlock up $256\to128$ & D-ResBlock down $128\to256$\\
    G-ResBlock up $128\to64$& D-ResBlock down $256\to512$\\
    Batch Normalization & D-ResBlock $512\to1024$\\
    ReLU & ReLU, AvgPool (ker: $8\times8$)\\
    conv (ker: $3\times3$, $64\to3$; stride: $1$; pad: $1$) & Linear($128\to1$)\\
    Tanh & Spectral Normalization\\
    \bottomrule
    \end{tabular}
\end{table}

\paragraph{Images generated on CIFAR-10 and STL-10}
In this section, we present sample images generated by the best performing trained generators on CIFAR-10 and STL-10.
\begin{figure}[h]
    \centering
    \includegraphics[width = \textwidth]{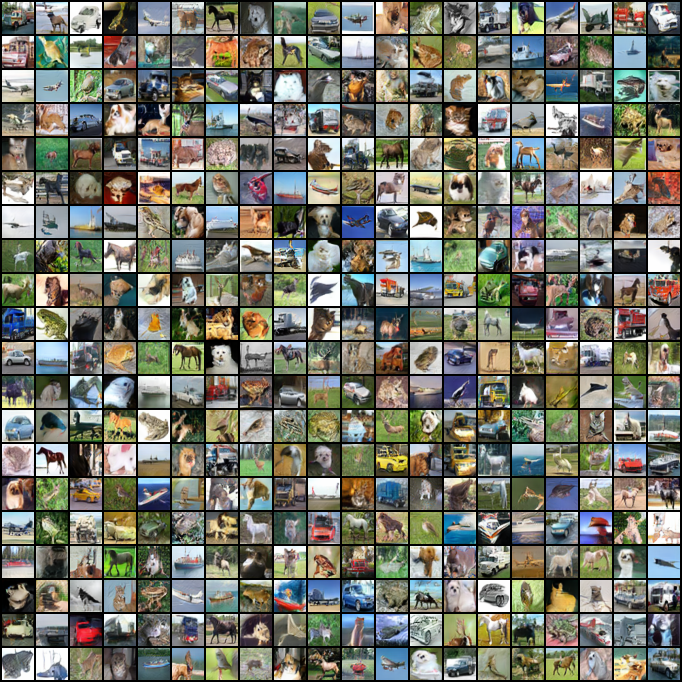}
    \caption{The presented samples are generated by the best performing trained generator on CIFAR-10, using Lv.$6$ Adam. This gives a FID score of $10.12$.}
    \label{fig: images generated on CIFAR-10 dataset, with $10.12$ FID score.}
\end{figure}
\begin{figure}[h]
    \centering
    \includegraphics[width=\textwidth]{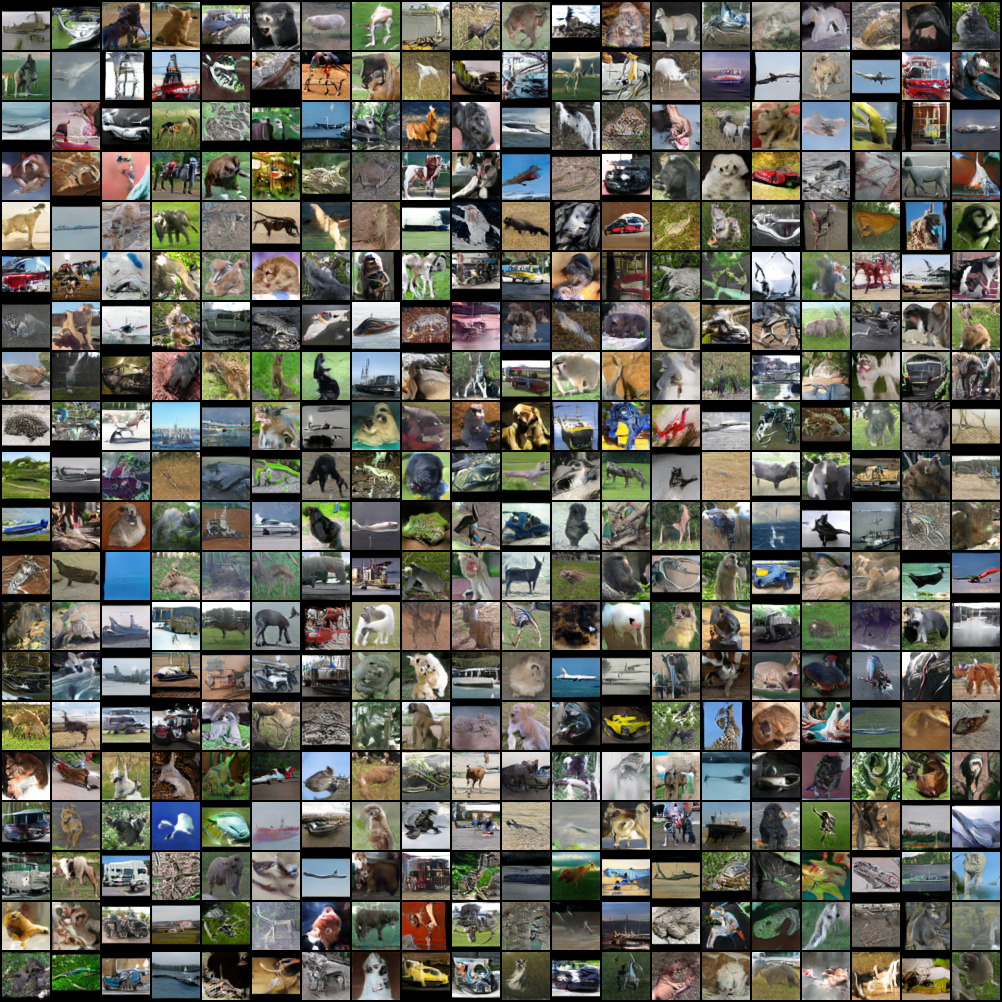}
    \caption{The presented samples are generated by the best performing trained generator on STL-10, using Lv.$6$ Adam. This gives a FID score of $25.43$.}
    \label{fig: images generated on STL-10 dataset, with 25.43 FID score}
\end{figure}

\end{document}